\documentclass{article}

\PassOptionsToPackage{numbers, compress}{natbib}

\usepackage[preprint]{neurips_2026}

\usepackage[utf8]{inputenc} 
\usepackage[T1]{fontenc}    
\usepackage{hyperref}       
\usepackage{url}            
\usepackage{booktabs}       
\usepackage{amsfonts}       
\usepackage{nicefrac}       
\usepackage{microtype}      
\usepackage{xcolor}         

\definecolor{cornflowerblue}{rgb}{0.39, 0.58, 0.93}
\hypersetup{
    colorlinks=true,
    linkcolor=cornflowerblue,
    filecolor=magenta,
    urlcolor=teal,
    citecolor=cornflowerblue
}

\usepackage{amsmath}
\usepackage{amssymb}
\usepackage{amsthm}
\usepackage{multirow}

\usepackage{algorithm}
\usepackage{algorithmic}
\usepackage{sidecap}
\usepackage{subcaption}
\usepackage{wrapfig}
\usepackage{cleveref}
\usepackage{array}

\usepackage{enumitem}
\setlist[itemize]{leftmargin=*}

\definecolor{metafg}{HTML}{1C2B33}
\definecolor{metabg}{HTML}{F1F4F7}

\theoremstyle{plain}

\theoremstyle{definition}
\newtheorem{definition}{Definition}[section]

\newtheorem{assumption}{Assumption}[section]

\newtheorem{desideratum}{Desideratum}

\crefname{definition}{definition}{definitions}
\Crefname{definition}{Definition}{Definitions}

\crefname{property}{property}{properties}
\Crefname{property}{Property}{Properties}

\crefname{assumption}{assumption}{assumptions}
\Crefname{assumption}{Assumption}{Assumptions}
\Crefname{proposition}{Proposition}{Propositions}

\Crefname{result}{Result}{Results}
\crefname{result}{Result}{Results}

\crefname{desideratum}{desideratum}{desiderata}
\Crefname{desideratum}{Desideratum}{Desiderata}

\usepackage{tcolorbox}
\tcbuselibrary{breakable, skins}

\newtcolorbox{observation}[1][Observation]{
  enhanced,
  breakable,
  colback=metabg,
  colframe=metafg,
  coltext=metafg,
  fonttitle=\sffamily\bfseries,
  title=#1,
  arc=4pt,
  left=0.4cm, right=0.4cm, top=0.3cm, bottom=0.3cm,
  boxrule=0.5pt,
}

\title{Quantifying Hyperparameter Transfer and the Importance of Embedding Layer Learning Rate}

\makeatletter
\newcommand{\myfnsymbol}[1]{%
  \expandafter\@myfnsymbol\csname c@#1\endcsname
}

\newcommand{\@myfnsymbol}[1]{%
  \ifcase #1
  \or 1
  \or 2
  \or 3
  \or 4
  \fi
}

\newcommand{\affiliationUMD}{\@myfnsymbol{1}}
\newcommand{\affiliationCS}{\@myfnsymbol{2}}
\newcommand{\affiliationJQI}{\@myfnsymbol{3}}
\newcommand{\affiliationMeta}{\@myfnsymbol{4}}

\author{Dayal Singh Kalra \textsuperscript{\normalfont{\affiliationCS}} 
\And
Maissam Barkeshli \textsuperscript{\normalfont{\affiliationUMD, \affiliationJQI,\affiliationMeta}} \hspace{1em}
}

\begin{document}

\maketitle

\footnotetext[1]{Department of Physics, University of Maryland, College Park}
\footnotetext[2]{Department of Computer Science, University of Maryland, College Park}
\footnotetext[3]{Joint Quantum Institute, University of Maryland, College Park}
\footnotetext[4]{Meta Superintelligence Labs, Fundamental AI Research}

\begin{abstract}
  Hyperparameter transfer allows extrapolating optimal optimization hyperparameters from small to large scales, making it critical for training large language models (LLMs). This is done either by fitting a scaling law to the hyperparameters or by a judicious choice of parameterization, such as Maximal Update ($\mu$P), that renders optimal hyperparameters approximately scale invariant. In this paper, we first develop a framework to quantify hyperparameter transfer through three metrics: (1) the quality of the scaling law fit, (2) the robustness to extrapolation errors, and (3) the asymptotic loss penalty due to choice of parameterization. Next, we investigate through a comprehensive series of ablations why $\mu$P appears to offer high-quality learning rate transfer relative to standard parameterization (SP), as existing theory is inadequate. We find that the overwhelming benefit of $\mu$P relative to SP when training with AdamW arises simply from maximizing the learning rate of the embedding layer. In SP, the embedding layer learning rate acts as a bottleneck that induces training instabilities; increasing it by a factor of width to match $\mu$P dramatically smooths out training while improving hyperparameter transfer. We also find that weight decay improves the scaling law fits, while, in the fixed token-per-parameter setting, it hurts the robustness of the extrapolation. 
\end{abstract}

\section{Introduction}
\label{section:introduction}

Training large neural networks requires carefully tuning numerous hyperparameters, including the learning rate, weight decay, and batch size, among others~\citep{li2025predictablescalei,deepseekai2024,qwen2025qwen25technicalreport,grattafiori2024llama3herdmodels,olmo20242,bergsma2025power}. 
As models scale into trillions of parameters, the cost of hyperparameter tuning at scale becomes prohibitive. The solution to this is \it hyperparameter transfer\rm: one finds the optimal hyperparameters at small scales, along with a scaling law, which can then be used to extrapolate optimal hyperparameters at large scales. There are two main approaches in practice. The first approach fits functional forms to predict how the optimal learning rate $\eta^*$ scales as the model and/or data is scaled~\citep{deepseekai2024,porian2024resolving,bjorck2025scaling}. 
While effective, these fitted relationships may not generalize beyond their fitted domain and themselves require expensive hyperparameter searches to obtain a robust functional form. 

\looseness -1
The second approach is more structural\textemdash it parameterizes the model such that the activations and their updates remain independent of width~\citep{yang21feature}, keeping the training dynamics invariant across scales~\citep{bordelon2022selfconsistent,yaida2022metaprincipledfamilyhyperparameterscaling}.~\citet{yang21feature} derived Maximal Update Parameterization ($\mu$P) under this desideratum, originally to study feature learning in the infinite width limit. Empirically, $\mu$P has a desirable practical property\textemdash it maintains \emph{fairly consistent} optimal learning rates across width. This property, termed \emph{learning rate transfer}, allows practitioners to utilize optimal learning rates from a smaller proxy model to train larger ones~\citep{yang2021tuning}, reducing the need for expensive hyperparameter searches at scale. 
Several works have since extended this framework to depth scaling~\citep{bordelon2024depthwise,yang2024tensor,dey2025dontlazycompletepenables}, alternative architectures~\citep{bordelon2024infinite_multihead,dey2024sparse,vankadara2024on,małaśnicki2025muparametrizationmixtureexperts,jiang2026hyperparametertransfermixtureofexpertlayers}, and optimizers~\citep{littwin2023adaptive,haas2024boldsymbolmumathbfp,qiu2025hyperparametertransferenablesconsistent}, suggesting the underlying principle generalizes to the design space of modern networks.

The current practice raises several questions. First, after fitting a scaling law to the hyperparameters, they can always be reparameterized to be (approximately) scale-invariant. In this case, is there any distinction with $\mu$P and its variants? This suggests an urgent need to \it quantify \rm the quality of hyperparameter transfer in order to compare across different schemes. 

\looseness -1
Secondly, the theoretical derivation of $\mu$P and its variants make several assumptions that do not hold in practice. It assumes a finite number of training steps in the infinite width limit, whereas practical training runs are in the opposite limit, of a large number of training steps compared to width. It also assumes full alignment between weight updates and the activations~\citep{yang21feature}, yet this alignment is low early in training and never reaches the full alignment even late in training~\citep{everett2024scaling}. It does not take into account aspects of the learning rate schedule, like learning rate warmup, which are crucial to observe transfer, as we show in~\Cref{appendix:learning_rate_warmup}. Finally, it is derived in the setting of fixed dataset size, whereas typical training runs scale data and number of parameters in tandem, often in the setting of a fixed number of tokens per parameter (TPP)~\cite{hoffmann2022an}. Surprisingly, despite these assumptions not holding, $\mu$P still exhibits high-quality learning rate transfer, raising the question of why. 

\looseness -1
In this paper, we begin by developing a framework to quantify hyperparameter transfer. We develop three key metrics. The first is the quality of the scaling law fit itself. The second is a robustness metric: whether the choice of optimal hyperparameter is more robust to perturbations as the model scales. The third regards the asymptotic performance of the loss: it is possible that some parameterizations exhibit higher quality transfer at the cost of degradation in the absolute value of the loss, and this also needs to be taken into account. The three metrics can trade off against each other, as we will see. 

\looseness -1
Armed with a quantitative framework to measure the quality of hyperparameter transfer, we then investigate in more detail why $\mu$P appears to exhibit high-quality transfer. We show that in standard decoder-only Transformers~\cite{attention} trained with AdamW~\cite{adamwloshchilov2018}, $\mu$P has four key changes compared to SP. By examining all $16$ ablations, we isolate the embedding layer learning rate as the key factor. $\mu$P has a much larger embedding layer learning rate compared to SP. By making this one alteration to SP, the hyperparameter transfer quality of SP essentially matches that of $\mu$P. We further study training dynamics, finding that the embedding layer learning rate is important throughout training, especially early in training, and if it is not sufficiently large, there are training instabilities. Therefore, one interesting source of training instabilities is that the learning can be bottlenecked by the embedding layer learning rate. 

These results highlight the fundamental importance of the embedding layer learning rate. It is known that the embedding layer learns a significant amount of structure from the data~\cite{nanda2023progress,he2024learning,tao2025how,karkada2026symmetrylanguagestatisticsshapes}, and as such, perhaps it is not surprising that keeping it too low can severely throttle training, while maximizing it helps stabilize the learning process. 

{\bf Our contributions}. Our paper contains the following contributions:
\begin{itemize}[topsep=1pt, itemsep=1pt, parsep=0pt, leftmargin=*]
    \item A quantitative framework for measuring the quality of hyperparameter transfer. This involves three metrics: (1) the quality of the scaling law fits in terms of an error $\mathcal{E}$, (2) a transfer robustness exponent $\kappa$ which measures sensitivity to errors in extrapolation of a hyperparameter from small to large scales, and (3) the asymptotic loss degradation $\mathcal{R}(\infty)$, which measures performance at scale relative to the optimal parameterization.  
    \item Focusing on GPT pre-training with AdamW, we express SP and $\mu$P in a common form (\Cref{tab:parameterizations}). The two parameterizations differ in four key ways (SP $\to \mu$P): (1) embedding learning rate $\Theta(\nicefrac{1}{n}) \to \Theta(1)$, (2) last layer initialization variance $\Theta(\nicefrac{1}{n}) \to \Theta(\nicefrac{1}{n^2})$, (3) LayerNorm learning rate ($\Theta(\nicefrac{1}{n}) \to \Theta(1)$), (4) attention scaling ($\nicefrac{1}{\sqrt{d}} \to \nicefrac{1}{d}$). 
    \item 
    We examine all 16 ablations that distinguish $\mu$P from SP and isolate the embedding layer learning rate as the primary factor: SP with an appropriately scaled embedding layer learning rate essentially matches $\mu$P in quality of hyperparameter transfer. 
    \item We find that the embedding layer learning rate, if not large enough, can cause training instabilities.
    \item 
    \looseness -1
    We study the effect of weight decay to show that (1) it can improve the scaling law fit quality (smaller $\mathcal{E}$), (2) in the fixed token per parameter setting, it hurts the robustness of the extrapolation. 
\end{itemize}

\section{Preliminaries: Neural Network Parameterizations}
\label{section:prelims}

\begin{table}[t]
\centering
\caption{Comparison of Standard (SP) and Maximal Update Parameterization ($\mu$P) for AdamW. Weight decay is scaled with the learning rate so that its contribution to the weight update, $\eta \cdot \lambda \cdot \theta$, remains $\Theta(1)$ across widths. }
\vspace{0.05 in}
\label{tab:parameterizations}
\renewcommand{\arraystretch}{1.2}
\setlength{\tabcolsep}{4pt}
\begin{tabular}{|c|l|cccc|}
\hline
Parameterization & Layer & Multiplier $(n^{-a})$ & Variance $(n^{-2b})$ & LR $(n^{-c})$ & WD $(n^{-d})$ \\
\hline
\multirow{5}{*}{\shortstack{Standard\\(SP)}}
& Embedding & $1$ & $1$ & $\nicefrac{1}{n}$ & $n$ \\
& Hidden    & $1$ & $\nicefrac{1}{n}$ & $\nicefrac{1}{n}$ & $n$ \\
& Last      & $1$ & $\nicefrac{1}{n}$ & $\nicefrac{1}{n}$ & $n$ \\
& LayerNorm & $-$ & $-$ & $\nicefrac{1}{n}$ & $-$ \\
\cline{2-6}
& Attention scale & $\nicefrac{1}{\sqrt{d}}$ & $-$ & $-$ & $-$ \\
\hline
\multirow{5}{*}{$\mu$P}
& Embedding & $1$ & $1$ & $1$ & $1$ \\
& Hidden    & $1$ & $\nicefrac{1}{n}$ & $\nicefrac{1}{n}$ & $n$ \\
& Last      & $1$ & $\nicefrac{1}{n^2}$ & $\nicefrac{1}{n}$ & $n$ \\
& LayerNorm & $-$ & $-$ & $1$ & $-$ \\
\cline{2-6}
& Attention scale & $\nicefrac{1}{d}$ & $-$ & $-$ & $-$ \\
\hline
\end{tabular}
\end{table}

We consider a neural network $f_{\mathbf{\theta}}$ with parameters $\mathbf{\theta}$ trained by minimizing a loss $L(\mathbf{\theta})$ using AdamW optimizer~\citep{adamwloshchilov2018} with learning rate $\eta$ and weight decay $\lambda$.
The performance of a model can be improved by scaling along different dimensions, such as width, depth, context length and data.
A \emph{parameterization} is then a set of rules that specify how key network and optimizer hyperparameters must be adjusted as the model is scaled to maintain stable training dynamics. 
In this work, we focus on the width $n$ as the scaling dimension and the learning rate $\eta$ as the hyperparameter to be scaled. 

Following~\citet{yang21feature}, we parameterize the network and optimizer using four scalar exponents $\{a_l, b_l, c_l, d_l\}$ per layer $l$, which control different aspects of width scaling. The exponent $a_l$ controls the forward pass scaling 
$\mathbf{h}^{(l+1)} = n^{-a_l} {W}^{(l+1)}\phi(\mathbf{h}^{(l)}),$
where $\mathbf{h}^{(l)}$ and ${W}^{(l)}$ denote the pre-activations and weights in layer $l$, and $\phi(\cdot)$ is the activation function. 
The exponent $b_l$ scales the initialization variance $W^{(l)} \sim \mathcal{N}(0, n^{-2b_l})$. 
The exponent $c_l$ scales the layer-wise learning rate $\eta^{(l)} = \eta \cdot n^{-c_l}$, where $\eta$ is the global learning rate. 
Finally, the exponent $d_l$ scales the weight decay strength $\lambda^l = \lambda \cdot n^{-d_l}$ such that the product $\eta^l \cdot \lambda^l$ is width-independent. 
For Transformers, the usual scaled dot-product attention comes with a per-head scaling by the 
head dimension $d$, with the standard choice being $\nicefrac{1}{\sqrt{d}}$~\cite{NIPS2017_3f5ee243}. 

\looseness -1
Given this general setup, specific parameterizations are obtained by imposing stability conditions on the training dynamics.
Two canonical examples of $abcd$ parameterization are SP~\citep{sohl2020infinite} and $\mu$P~\citep{yang21feature}.
SP requires that activations at initialization do not blow up or vanish as the width grows, imposing one constraint per layer; any additional freedom  is used to make the simplest choices, such as a uniform learning rate for all layers. 
$\mu$P imposes a stronger condition: both the activations and their updates must be width-independent, resulting in two constraints per layer. 
For analytical tractability, these additional constraints are derived under three additional assumptions: (1) a finite number of training steps as the width $n \rightarrow \infty$, (2) full alignment between weight updates and the activations they act on~\citep{yang21feature}, and (3) a fixed dataset as $n$ scales. 
We provide a self-contained derivation of $\mu$P for both SGD and Adam optimizers in~\Cref{appendix:mup_derivation}.

\looseness -1
Since one of our goals is to understand which aspects of $\mu$P are essential for transfer, we need to express SP and $\mu$P in a common form.
To this end, SP is defined with a weight initialization variance of $\Theta(\nicefrac{1}{\text{fan-in}})$, and a global learning rate $\eta_l = \eta \cdot n^{-1}$ for all layers. Note that we choose a convention where we peel off a $1/n$ factor in the learning rate, which differs from other conventions.
To make the comparison between SP and $\mu$P more transparent, we use the symmetry of abcd parameterization~\citep{yang21feature} to set the multipliers $a = 0$ in our description of $\mu$P. As shown in \Cref{tab:parameterizations}, the two differ in four key ways: (1) embedding layer learning rate, (2) last layer initialization variance, (3) LayerNorm learning rate, and (4) attention scaling. 
Weight decay strength $\lambda$ is scaled with learning rate so that its contribution to the weight update, $\eta \lambda {\theta}$, is $\Theta(1)$ in both parameterizations.

\looseness -1
The intuitive reason for these changes is as follows. 
SP applies a uniform $\Theta(\nicefrac{1}{n})$ learning rate to all layers.
This scaling is needed for the hidden layers, which compute their output as a sum of $n$ input terms $h^{(l)}_i = \sum_{j=1}^n W^{l+1}_{ij} h^{(l)}_j$; without a $\Theta(\nicefrac{1}{n})$ learning rate the activations updates $\Delta {h}^{(l+1)}_i$ would scale as $\Theta(n)$.
By comparison, the embedding (a per-token lookup) and LayerNorm (an elementwise operation) do not involve a sum over the width dimension, so a $\Theta(1)$ learning rate is a natural choice for these layers.
Next, SP's larger last layer initialization variance leads to higher Hessian sharpness at initialization, making training unstable at large learning rates~\cite{kalra2025universal}. However, in practice, learning rate warmup mitigates this by gradually reducing sharpness to a level determined by the peak learning rate~\citep{kalra2024why}, making the initialization difference between SP and $\mu$P negligible.
Finally, $\mu$P replaces SP's $\nicefrac{1}{\sqrt{d}}$ attention scaling with $\nicefrac{1}{d}$, motivated by the observation that key and query vectors are projections of the same input and therefore would be more aligned than independent random vectors during training.
In this work, we show that the embedding layer learning rate is the primary driver of $\mu$P's advantage, with the remaining modifications contributing little.

\section{Quantifying Hyperparameter Transfer} 
\label{section:transfer-framework}

\looseness -1
In this section, we focus on learning rate $\eta$ as the hyperparameter of interest and consider scaling the width $n$. In principle, the discussion can be extended to include additional hyperparameters (e.g., batch size, weight decay) and scaling other quantities such as depth, context length, training steps. 

A parameterization is effective for learning rate transfer if the optimal learning rate can be reliably extrapolated from small to large widths without sacrificing performance at scale. Reliability requires two conditions. First, the loss at the end of training must be predictable, for example, by satisfying a simple functional scaling law form. 
Second, even if the loss is predictable, the transferred learning rate will inevitably carry some residual error (e.g., from finite sampling at the small scale).
The loss at scale must be robust to such errors\textemdash if a small error in predicting the learning rate induces an increasingly large loss gap at larger widths, transfer becomes sensitive despite predictability in the loss. 
Accordingly, we introduce three metrics:
\emph{Loss Predictability Error} $\mathcal{E}$, \emph{Transfer Robustness Exponent} $\kappa$, and \emph{Asymptotic Loss Degradation} $\mathcal{R}(\infty)$ that capture loss predictability, robustness to prediction errors, and performance at scale, respectively.

\looseness -1
To formalize these metrics, we first need to model how the loss landscape changes with width and learning rate.
Following standard practice in neural scaling laws~\citep{kaplan2020scalinglawsneurallanguage,hoffmann2022an,barkeshli2026originneuralscalinglaws}, we model the optimal loss $L^*(n)$ as a power law in width:
\begin{align}
    L^*(n) = L^*(\infty) + A n^{-\alpha},
\end{align}
where $L^*({\infty})$ is the irreducible loss, $A$ is the scaling coefficient, and $\alpha$ is the scaling exponent. Here we can consider the dataset size to be held fixed or increasing along with $n$ as in compute-optimal training. 
We model the optimal learning rate using a scaling law with an irreducible term as well:
\begin{align}
    \eta^*(n) = \eta^*(\infty) + B' n^{-\beta}, 
    \label{eq:irr_lr_scaling_law}
\end{align}
where $0 < \eta^*(\infty) < \infty$ is the asymptotic optimal learning rate, $B'$ is the scaling coefficient and $\beta > 0$ controls the rate of convergence. Note that this differs from the form $\eta^*(n) \sim n^{-\beta'}$ used elsewhere ~\citep{everett2024scaling,li2025predictablescalei}, which implicitly assumes that the optimal learning rate vanishes ($\beta' > 0$) or diverges ($\beta' < 0$) at large widths. In \Cref{eq:irr_lr_scaling_law}, we assume the dominant scaling law has been peeled off (as in our convention for SP), leaving the residual correction $B' n^{-\beta}$.

\begin{figure}[t]
    \vspace{-0.2 in}
    \centering
    \includegraphics[width=\textwidth]{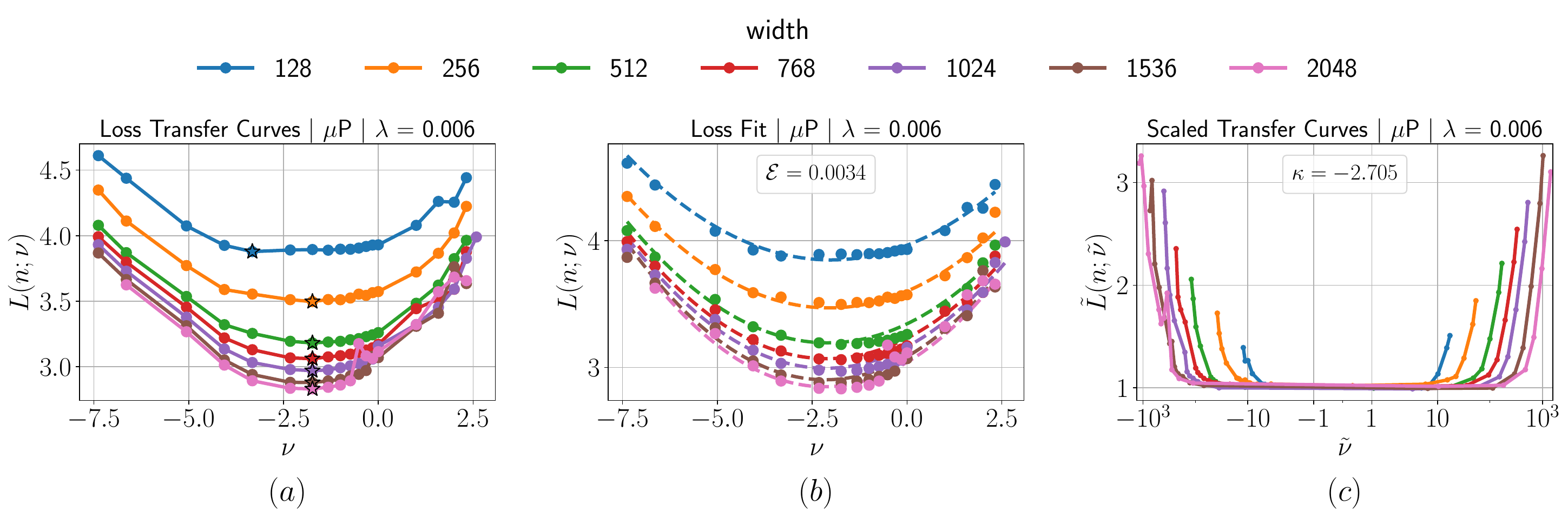}        \includegraphics[width=\textwidth]{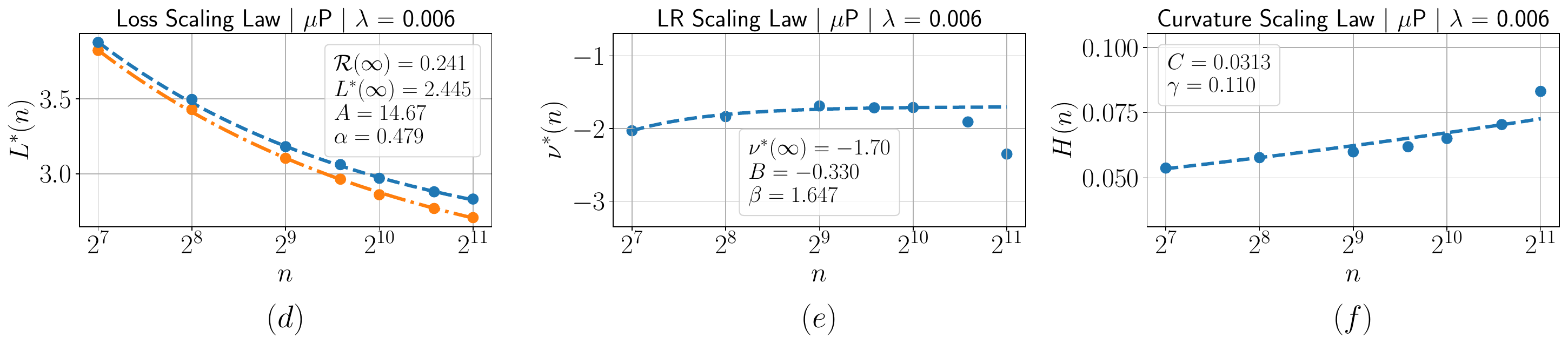}
    \caption{
    \looseness -1
    \emph{Computing the three transfer metrics for $\mu$P}.
    (a) Loss vs. log learning rate $\nu$, with star marking the optimum $\nu^*(n)$,
    (b) Joint fit of the loss model (\Cref{eq:full_model}, dashed lines), with a low predictability error $\mathcal{E} = 0.0034$ , 
    (c) Loss curves in the normalized coordinates (\Cref{equation:normalized_coordinates}), with $\kappa = -2.640$ indicating robust transfer.
    (d-f)
    Scaling laws for optimal loss $L^*(n)$, optimal log-learning-rate $\nu^*(n)$, and curvature $H(n)$. In (d), the orange curve shows the best loss across parameterizations at each width, used for estimating the asymptotic loss gap $\mathcal{R}(\infty)$.}
    \label{fig:transfer_curves_mup}
    \vspace{-0.2 in}
\end{figure}

\looseness -1
The scaling law above is expressed in terms of the learning rate $\eta$. However, the loss as a function of $\eta$ is typically asymmetric around its optimum $\eta^*$\textemdash beyond it training becomes unstable, and the loss increases sharply, while below it the loss degrades gradually. We therefore work in log-learning-rate space $\nu := \log_2 \eta$ for the rest of the paper, where the loss landscape is more symmetric around its optimum (see \Cref{fig:transfer_curves_mup}(a)). \Cref{eq:irr_lr_scaling_law} then takes the form:
\begin{align}
    \nu^*(n) 
    = \nu^*(\infty) + \log\left( 1+ \frac{B' n^{-\beta}}{\eta^*(\infty)}  \right) \approx \nu^*(\infty) + B n^{-\beta}, \nonumber
\end{align}
\looseness -1
where $B = B' / \eta^*(\infty)$. 
Since Eq. \ref{eq:irr_lr_scaling_law} captures the residual width dependence after the dominant scaling has been removed, the term $B'n^{-\beta}/\eta^*(\infty)$ is small, justifying the first-order Taylor expansion above. 

The above scaling laws hold for each parameterization separately; different parameterizations may, for example, exhibit different irreducible loss. As such, a parameterization with high-quality transfer might sacrifice loss performance. This motivates the following definition. 
\begin{definition}[Loss Degradation $\mathcal{R}(\infty)$]
\label{def:loss-degradation}
For a parameterization, let $L^*(\infty)$ denote its irreducible loss, and let $L^*_{\text{best}}(\infty)$ be the best possible irreducible loss across hyperparameters and parameterizations. We define loss degradation as the asymptotic loss gap:
\begin{align}
\mathcal{R}(\infty) = L^*(\infty) - L^*_{\text{best}}(\infty) \geq 0.
\end{align}
A parameterization with $\mathcal{R}(\infty) = 0$ achieves best-in-class loss at scale, while $\mathcal{R}(\infty) > 0$ indicates a performance gap at scale. In practice, $L^*_{\text{best}}$ is computed using a finite set of parameterizations under consideration. 
\end{definition}
Next, to capture the reliability of transfer, we model the loss around $\nu^*(n)$ as a local quadratic form:
\begin{align}
    L(\nu; n)  = L^*(n) + \frac{1}{2} H(n) \cdot (\nu - \nu^*(n))^2 + \mathcal{O}(\nu^3),
    \label{equation:loss_quadratic}
\end{align}
where $H(n) = \nabla_\nu^2 L(\nu^*(n); n)$ is the loss Hessian evaluated at $\nu^*(n)$. 
In addition to the loss and log-learning rate, we model the Hessian scaling as a power law\footnote{Empirically, we find this form well approximates the Hessian scaling for many parameterizations we considered.}:
\begin{align}
    H(n) = C n^{\gamma},
\end{align}
where $C$ is the scaling coefficient and $\gamma$ is the scaling exponent. 
Substituting the scaling laws into \Cref{equation:loss_quadratic}, we obtain:
\begin{align}
    L(\nu; n) = L^*(\infty) + A n^{-\alpha} + \frac{1}{2}C n^{\gamma} \cdot \left( \nu - \nu^*(\infty) - B n^{-\beta} \right)^2.
    \label{eq:full_model}
\end{align}
This functional form serves as our scaling ansatz for learning rate transfer.
\begin{definition}[Loss Predictability Error $\mathcal{E}$]
\label{def:transfer-quality}
Given loss observations $\{L(\nu_i; n_j)\}$ at $N_\nu$ log-learning-rates $\{\nu_i\}_{i=1}^{N_\nu}$ and $N_n$ widths $\{n_j\}_{j = 1}^{N_n}$, we define loss predictability as the normalized mean squared error between observed and predicted loss $\{\hat{L}(\nu_i; n_j)\}$ from \Cref{eq:full_model} with fitted parameters:
\begin{align}
\mathcal{E} = \frac{1}{N_\nu N_n}  \sum_{i,j}^{N_\nu, N_n}\left[L(\nu_i; n_j) - \hat{L}(\nu_i; n_j) \right]^2.
\end{align}
When $\mathcal{E} \approx 0$, the loss is well captured by \Cref{eq:full_model}, suggesting reliable extrapolation across widths is possible. High $\mathcal{E}$ in contrast suggests a complex landscape arising from various sources, such as training instabilities, finite-size effects, or phase transitions, making an extrapolation unreliable.
\end{definition}

While $\mathcal{E}$ captures the predictability of the landscape, it does not reveal how errors in extrapolating the optimal learning rate impact performance at large scales. 
To analyze this sensitivity, we normalize both loss and log-learning-rate, focusing on the scale-invariant landscape defined by the normalized coordinates:
\begin{align}
    \tilde{L}(n) = \frac{L(\nu; n) - L^*(\infty)}{A n^{-\alpha}}, \qquad \tilde{\nu}(n) = \frac{\nu - \nu^*(\infty)}{B n^{-\beta}}.
    \label{equation:normalized_coordinates}
\end{align}
In these normalized coordinates, \Cref{eq:full_model} becomes:
\begin{align}
    \tilde{L}(n) = 1 + \frac{C}{2A B^2} n^{\alpha - 2\beta + \gamma} \left(\tilde{\nu}(n) - 1 \right)^2.
\end{align}
The normalized loss curvature scales as $n^{\kappa}$, where $\kappa = \alpha - 2\beta + \gamma$, determines whether the landscape flattens ($\kappa < 0$) or sharpens ($\kappa > 0$) as width increases. The sign of $\kappa$ thus determines the sensitivity of the loss to learning rate prediction errors at scale.
\begin{definition}[Transfer Robustness Exponent $\kappa$]
\label{def:robust-transfer}
We say that a parameterization under our ansatz (\Cref{eq:full_model}) exhibits \emph{robust transfer} if:
\begin{align}
    \kappa = \alpha - 2\beta + \gamma \leq 0.
\end{align}
\end{definition}
The sign of $\kappa$ controls how prediction errors propagate to loss at scale.
Negative $\kappa$ means the landscape flattens, so errors in extrapolating 
$\nu^*(n)$ from small widths results in diminishing loss penalties at scale. By comparison, positive $\kappa$ amplifies these errors and degrades transfer reliability, as the landscape sharpens with width.
When $\gamma = 0$, the condition $\alpha \leq 2 \beta$ coincides with the \emph{fast transfer} criterion of~\citet{ghosh2025understandingmechanismsfasthyperparameter}.

Taken together, these metrics capture three complementary axes of transfer. $\mathcal{R}(\infty)$ captures performance at scale, $\mathcal{E}$ quantifies if the loss is predictable, and $\kappa$ measures whether prediction errors amplify or dampen at large widths.
A parameterization can excel on one while failing on others, and as we show in later sections, the metrics can even be at odds with each other.

\section{Examining $\mu$P and SP Through the Lens of the Three Transfer Metrics}
\label{section:mup_sp}

\begin{figure}[t]
    \vspace{-0.3 in}
    \centering
    \includegraphics[width=\textwidth]{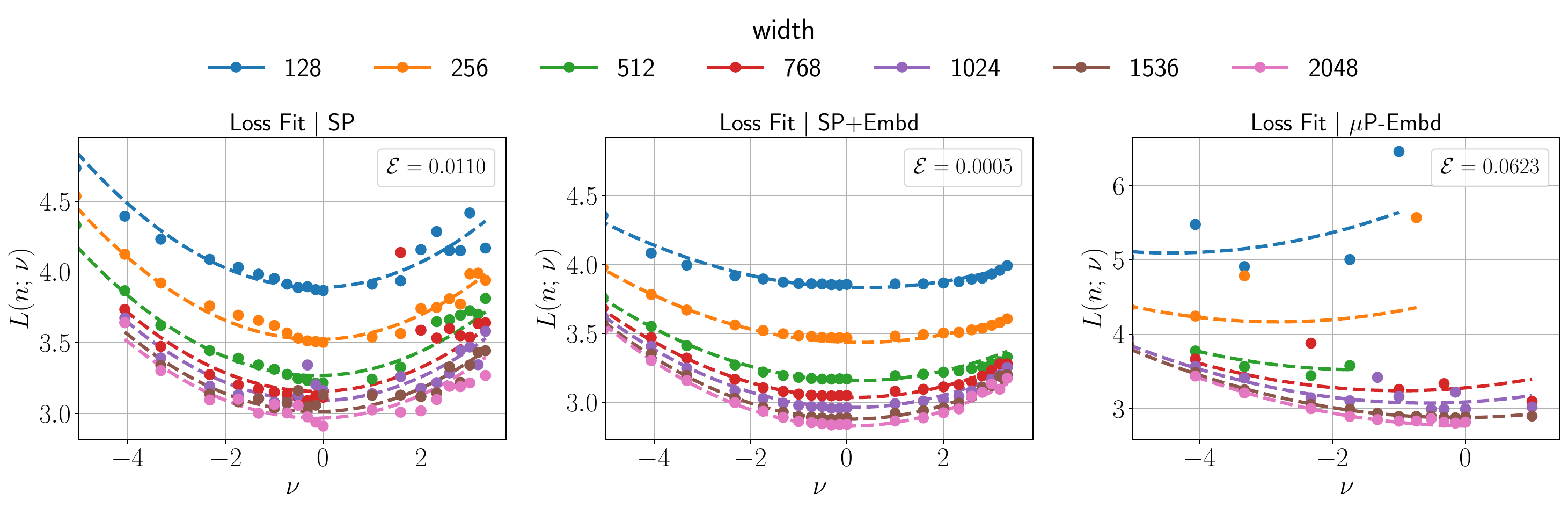}        
    \caption{
    \looseness -1
    \emph{Embedding layer learning rate is the critical difference between SP and $\mu$P.}
    Loss vs. log learning rate $\nu$ across widths for:
    (a) SP with $\Theta(\nicefrac{1}{n})$ embedding learning rate;
    (b) SP modified to use $\Theta(1)$ embedding learning rate (SP+Embd);
    (c) $\mu$P modified to use $\Theta(\nicefrac{1}{n})$ embedding learning rate ($\mu$P-Embd).
    Speeding up the embedding in SP eliminates training instabilities and yields smooth, $\mu$P-like curves, while slowing it down in $\mu$P reintroduces SP-like instabilities. These interventions isolate the embedding learning rate as the primary driver of $\mu$P's advantage, and show that training it fast enough is critical for stable training.
    }
    \label{fig:transfer_curves_embd}
    \vspace{-0.2 in}
\end{figure}

\looseness -1
We pre-train GPT-style Transformers on FineWeb-Edu~\citep{penedo2024the} using AdamW for a fixed number of steps ($T = 10,000$) using  Warmup-Stable-Decay (WSD) schedule~\citep{hu2024minicpm} ($20\%$ warmup, $60\%$ stable, $20\%$ decay) with batch size $1024$ ($\approx 1$M tokens per step).
We scale the embedding dimension (width) $n \in [128, 2048]$ by increasing the number of heads, while keeping the head dimension fixed at $d = 64$. 
For each width, we sweep the peak learning rate and weight decay strength $\lambda$.
For full experimental details, see~\Cref{appendix:experimental_details}.
We consider compute-optimal (fixed TPP) training in~\Cref{section:wd_and_tpp}.
\subsection{Methodology}
\label{subsection:methodology}
\looseness -1
\textbf{Filtering and Interpolation.}
For each parameterization and width, we retain runs with loss within $1.35\times$ the per-width optimal loss to focus on the landscape around the minimum (\Cref{fig:transfer_curves_mup}(a)).
Since discrete sampling introduces noise in the estimated $\nu^*(n)$, we interpolate the resulting per-width curves using a cubic spline to obtain smoother, denser curves (see \Cref{fig:interpolated_curves_mup}, \Cref{appendix:transfer_framework_fitting}), enabling reliable estimation of $\nu^*(n)$ and $H(n)$. For the optimal loss, we use the raw per-width minimum directly, as it's already reliable.

\looseness -1
\textbf{Estimating the Three Transfer Metrics.}
By fitting scaling laws to $L^*(n)$ and $\nu^*(n)$ (\Cref{fig:transfer_curves_mup}(d, e)), we obtain the exponents $\alpha$, $\beta$ and the asymptotic loss $L^*(\infty)$, from which we compute $\mathcal{R}(\infty)$. 
Since $\mathcal{R}(\infty)$ is non-negative by definition (\Cref{def:loss-degradation}), we clamp fitted values near zero when they are negative due to finite-size fitting artifacts.
For the curvature $H(n)$, we first fit a quadratic centered at $\nu^*(n)$ to the per-width loss curves (see \Cref{fig:curvature_fits_mup} in \Cref{appendix:transfer_framework_fitting}), and then fit the scaling law $H(n) = Cn^\gamma$ (\Cref{fig:transfer_curves_mup}(f)) to obtain the exponent $\gamma$. 
Combined with $\alpha$ and $\beta$ from above, this gives $\kappa = \alpha - 2\beta + \gamma$, whose sign can be directly read off from the normalized transfer curves (\Cref{fig:transfer_curves_mup}(c)), with a flattening landscape indicating robust transfer.
Finally, to compute the loss predictability error $\mathcal{E}$, we jointly fit all the  parameters using the interpolated curves, then measure $\mathcal{E}$ using the fit evaluated on the raw filtered data points. Fitting on the interpolated data gives a smoother fit, while evaluating on the raw data measures the fit on the observed data. 
Throughout, we cap all scaling exponents at $2.0$ to prevent spuriously large values, which would otherwise dominate comparisons across parameterization.

\looseness -1
When $\nu^*(n)$ is nearly constant across widths, its scaling law admits two degenerate solutions: $\beta \to 0$ (constant) and $\beta \to \infty$ (rapid convergence). 
We develop a procedure to distinguish between these two cases, and prefer the $\beta \to \infty$ solution as it is consistent with our framework's convergence-based interpretation.
We provide full details on the fitting procedures and degeneracy resolution in \Cref{appendix:transfer_framework_fitting}.

\subsection{What Exactly is $\mu$P's Advantage over SP?}

To find out the essential elements required for reliable transfer, we compare $\mu$P (\Cref{fig:transfer_curves_mup}) and SP (\Cref{fig:transfer_curves_embd}(a) and \Cref{fig:transfer_curves_sp} in \Cref{appendix:transfer_framework_add_on}) using the three metrics. 
SP exhibits visibly noisier loss curves than $\mu$P due to training instabilities.
Despite this, the two parameterizations are surprisingly similar on most metrics.
The asymptotic loss gap $\mathcal{R}(\infty)$ of SP is slightly worse but still comparable to $\mu$P. 
For both parameterizations, $\nu^*(n)$ empirically converges to a finite asymptotic value.
Finally, the normalized loss flattens for both parameterizations, with large negative robustness exponents, indicating transfer is robust in both cases.
Where SP falls short is in the loss predictability: the loss vs. $\nu$ curves are visibly noisier due to training instabilities. As a result, the predictability error $\mathcal{E}$ is roughly $3 \times$ larger than for $\mu$P, suggesting that the loss is poorly described by our ansatz. 
While SP can exhibit transfer in principle, training instabilities make it unreliable in practice.

\subsection{A Step by Step Journey from SP to $\mu$P}

\begin{figure}[t]
    \vspace{-0.3 in}
    \centering
    \includegraphics[width=\linewidth]{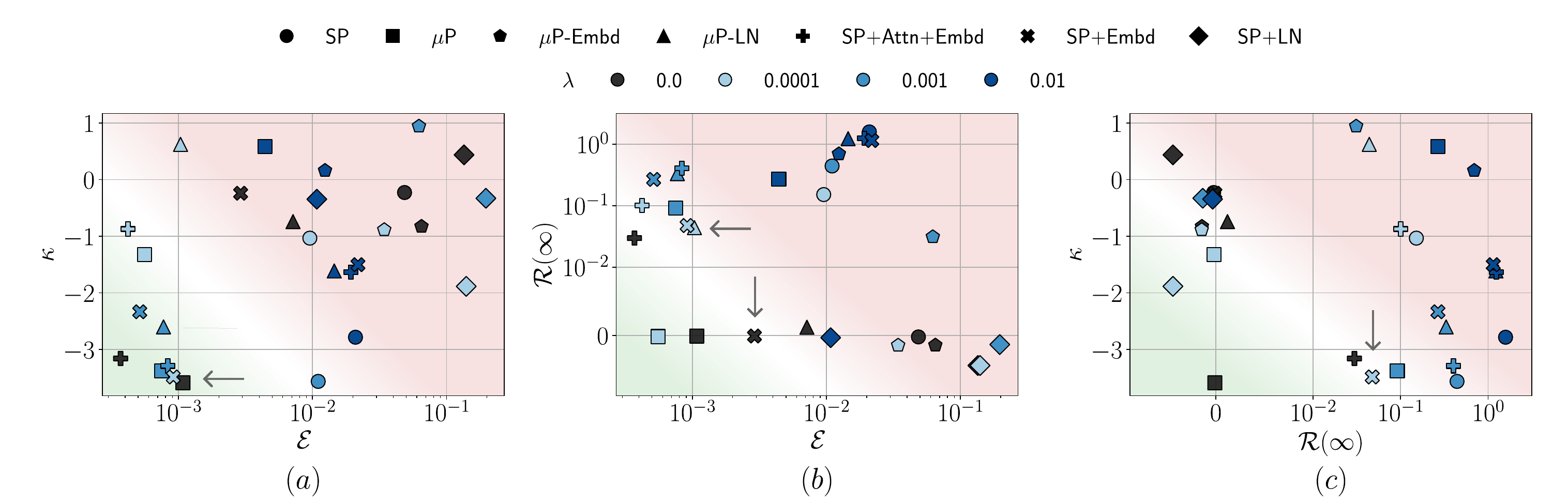}
    \caption{
    \looseness -1
    \emph{Transfer metrics for parameterizations interpolating between SP and $\mu$P.}
    Parameterizations with `$+$' denote incremental changes from SP towards $\mu$P, while `$-$' denotes changes from $\mu$P towards SP. Green and red regions indicate desirable and undesirable regimes, respectively. The orange arrow highlights SP+Embd (SP with $\Theta(1)$ embedding learning rate), which matches $\mu$P across all three metrics, suggesting the embedding layer learning rate is the primary driver of $\mu$P's advantage.}
    \label{fig:transfer-framework-phase-diagram}
    \vspace{-0.2 in}
\end{figure}

\looseness -1
The results above suggest that SP has the right ingredients for reliable learning rate transfer, but is held back by training instability. 
Since SP and $\mu$P differ in only four ways (\Cref{section:prelims}), a natural question is which change, if any, is most significant.
To answer this, we perform systematic modifications, starting from SP and making one change at a time, sweeping the peak learning rate and weight decay across widths.
We caution that these modifications interact non-linearly, so the effect of one change may depend on the scaling of other layers.
\Cref{fig:transfer-framework-phase-diagram} shows the three transfer metrics for selected ablations and weight decay values. We defer the full results to \Cref{fig:transfer-framework-phase-diagram-full} in \Cref{appendix:transfer_framework_add_on} and summarize the key findings here.

The embedding layer learning rate emerges as the most critical modification (\Cref{fig:transfer_curves_embd}): training SP with an $\Theta(1)$ embedding layer learning rate (SP+Embd) matches $\mu$P across metrics, while training $\mu$P with $\Theta(\nicefrac{1}{n})$ embedding layer learning rate ($\mu$P-Embd) degrades it. 
Attention scale has a more subtle effect: both adding $\nicefrac{1}{d}$ scaling to SP (SP+Attn) and removing it from $\mu$P ($\mu$P-Attn) result in large positive $\kappa$, making transfer brittle. 
Increasing the LayerNorm learning rate to $\Theta(1)$ in SP worsens instability, while decreasing it to $\Theta(\nicefrac{1}{n})$ in $\mu$P has a negligible effect, suggesting that LayerNorm parameters can be trained slowly without hurting performance.
Finally, the last layer initialization variance has a negligible effect, though $\mu$P with a $\nicefrac{1}{n}$ initialization ($\mu$P-Last) exhibits instabilities at small widths.
We leave a detailed understanding of these observations to future work.

\section{The Importance of Embedding Layer Learning Rate}
\label{section:embedding_layer_switch}

\looseness -1
The critical role of the embedding layer learning rate in stabilizing SP's training is surprising. 
Since the embedding layer performs a width-independent lookup, one would expect its learning rate to scale as $\Theta(1)$. 
But what is unexpected is that a smaller learning rate results in noisy loss curves in SP\footnote{These instabilities appear near the optimal learning rate; at smaller learning rates, SP trains stably but converges to worse final loss.}(\Cref{fig:transfer_curves_embd}), 
as one might expect later layers to compensate for a poorly trained embedding. 
In this section, we dig deeper into why training the embedding layer fast enough is critical.

\begin{figure}[!t]
\vspace{-0.3in}
  \centering
  \begin{minipage}[t]{0.03\textwidth}
    \vskip -0.4in
    \caption*{(a)}
  \end{minipage}
  \begin{minipage}[c]{0.43\textwidth}
    \centering
    \includegraphics[width=0.85\linewidth]{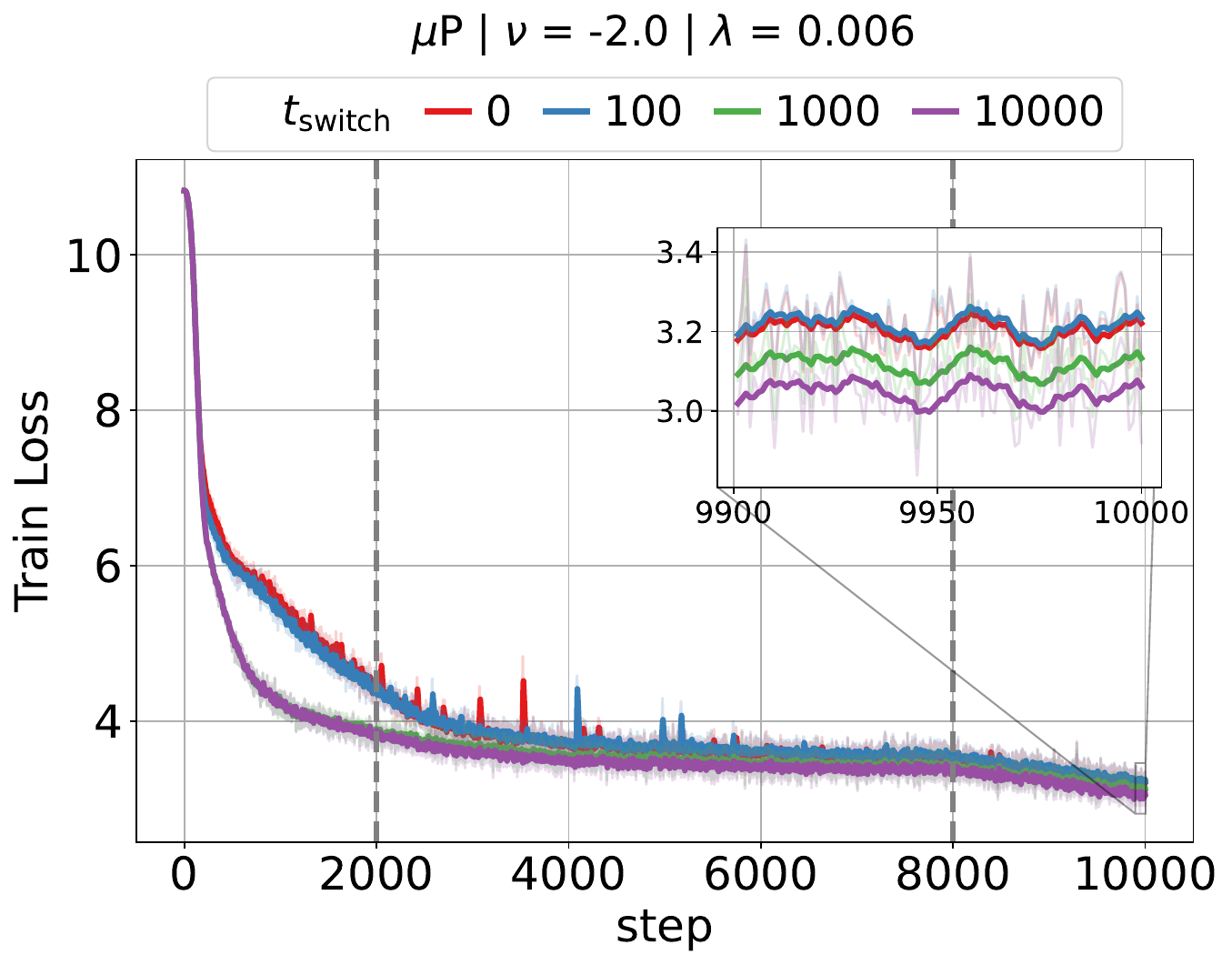}
  \end{minipage}
  \hskip 0.2in
  \begin{minipage}[t]{0.03\textwidth}
    \vskip -0.4in
    \caption*{(b)}
  \end{minipage}
  \begin{minipage}[c]{0.43\textwidth}
    \centering
    \includegraphics[width=0.85\linewidth]{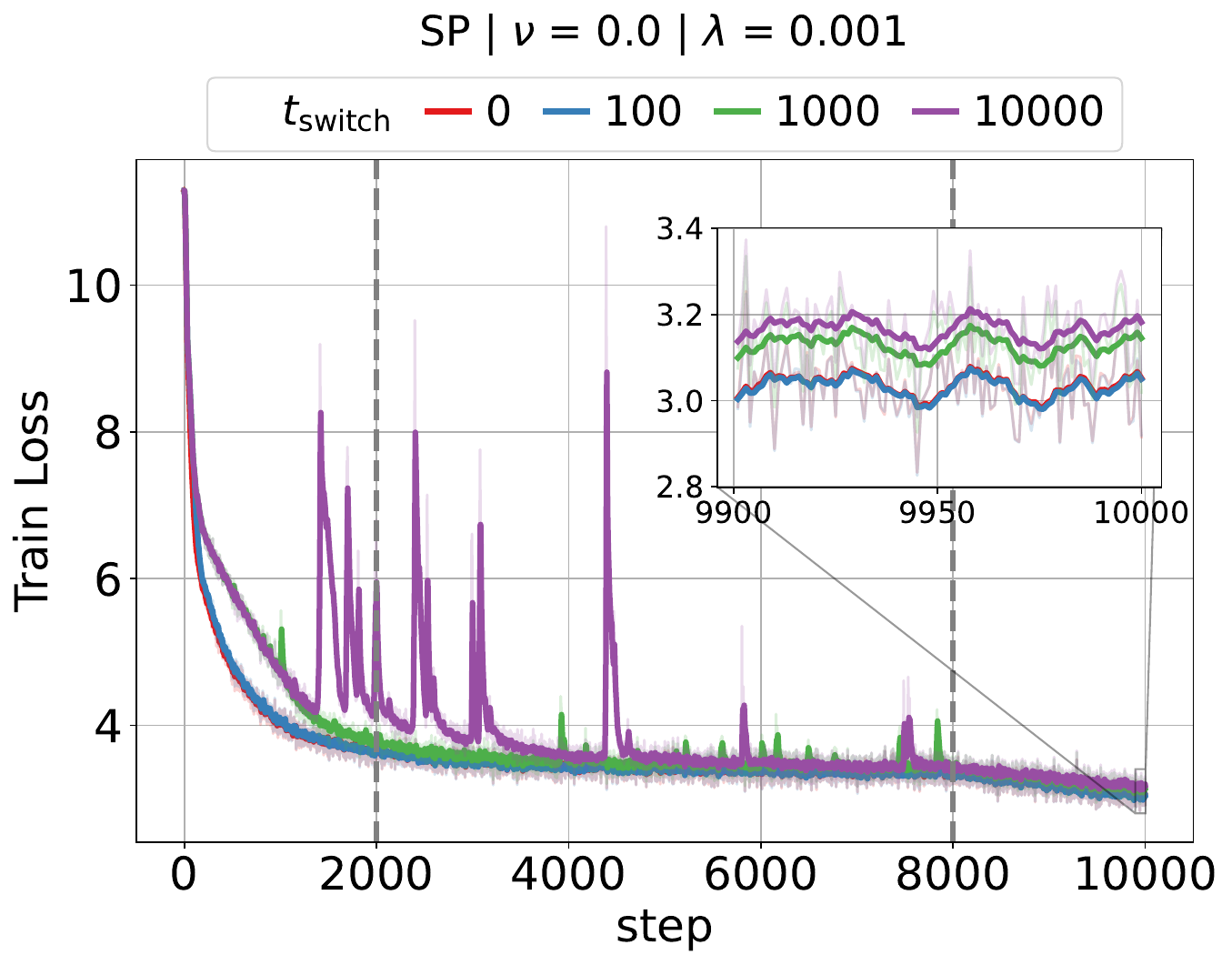}
    \label{fig:sub2}
  \end{minipage}
  \caption{(a) Reducing the embedding learning rate from $\Theta(1)$ to $\Theta(\nicefrac{1}{n})$ in $\mu$P at step $t_{\text{switch}}$ causes a persistent loss gap that grows larger for earlier switches (inset). (b) Increasing it from $\Theta(\nicefrac{1}{n})$ to $\Theta(1)$ in SP at step $t_{\text{switch}}$ eliminates training instabilities and improves performance over SP ($t_{\text{switch}}=10^4$).}
  \label{fig:switch_step_experiments}
  \vspace{-0.2in}
\end{figure}

\looseness -1
To better understand the role of training the embedding layer, we examine when it matters most by switching its learning rate at various points during training.
We perform two experiments: in $\mu$P, we slow down the embedding learning rate from $\Theta(1)$ to $\Theta(\nicefrac{1}{n})$ at step $t_{\text{switch}}$, and in SP, we speed it up from $\Theta(\nicefrac{1}{n})$ to $\Theta(1)$.
Together, these experiments reveal that a small embedding layer learning rate not only slows down training but also causes instabilities, with early training being the most critical.
In the $\mu$P case (\Cref{fig:switch_step_experiments}a), switching to $\Theta(\nicefrac{1}{n})$ embedding learning rate early in training drastically slows down training.
While further training eventually closes much of this gap, a residual loss difference of $\sim 0.1$-$0.2$ persists at the end of training, with a larger gap for earlier switches. 
By comparison, in the SP case (\Cref{fig:switch_step_experiments}b), switching to $\Theta(1)$ embedding learning rate at an early stage simultaneously eliminates the training instabilities and improves performance.
\looseness -1
We additionally perform two complementary experiments to better understand the role of different layers. First, in \Cref{appendix:frozen_embd}, we completely freeze the embedding layer at initialization, finding that this hurts both parameterizations, but $\mu$P much more than SP. This is surprising, as one might expect later layers to learn useful representations even with a randomly initialized embedding; however, the persistent gap indicates that they do not compensate for an untrained embedding.
Second, we test whether the importance of training the first layer fast is specific to Transformers. In \Cref{appendix:cnn_cifar_adam}, we show that for CNNs trained on CIFAR-100~\citep{cifar100}, SP with a $\Theta(1)$ input-layer learning rate matches $\mu$P's transfer quality, while changing only the last-layer initialization has little effect.
Together, these results suggest that the first and last layers play a special role across architectures, likely because they sit at the network boundary with no upstream or downstream processing to compensate for poor training. Therefore, extra care should be taken in setting their hyperparameters.

A natural question is: why does the embedding layer learning rate matter so much while the last-layer initialization does not?
First, the two are not on equal footing in $\mu$P's derivation.
The $\Theta(1)$ embedding learning rate is required for the embedding's activation updates to be $\Theta(1)$. This first-layer update has only a single term contributing to it, so its scaling has to be correct.
By contrast, the last-layer initialization contributes to only one of the three terms of the function update $\Delta f_1$, and its specific value depends on whether the last-layer weights are \textit{fully aligned} with their activation updates during training (see \Cref{assumption:mup_full_alignment} in \Cref{appendix:mup_derivation}).
The constraint is therefore weaker and depends on an empirical alignment value that may not hold in practice\textemdash indeed,~\citet{everett2024scaling} showed that relaxing this assumption still yields transfer.
Second, learning rate warmup steers training away from the higher sharpness induced by a larger last-layer initialization, into flatter regions of the landscape~\citep{kalra2024why}, compensating for the difference in initialization scale.

\section{The Effect of Weight Decay and Compute Optimal Training}
\label{section:wd_and_tpp}

Most studies examining transfer focus on the fixed-step setting, often with little or no weight decay~\citep{yang2021tuning,everett2024scaling}. 
While recent work has begun examining the effect of weight decay~\citep{kosson2026weight} and compute-optimal scaling regime~\citep{bergsma2025power}, their effect on transfer quality remains unclear. 
We find that weight decay improves loss predictability error $\mathcal{E}$ but consistently hurts asymptotic performance in the fixed-step setting. 
By comparison, this performance penalty disappears in the compute optimal scaling regime (20 tokens per parameter, following~\citet{hoffmann2022an}), but transfer robustness degrades with weight decay, likely because the appropriate weight decay scaling in this regime is not well understood.

\begin{figure}[t]
    \vspace{-0.3 in}
    \centering
    \includegraphics[width=\linewidth]{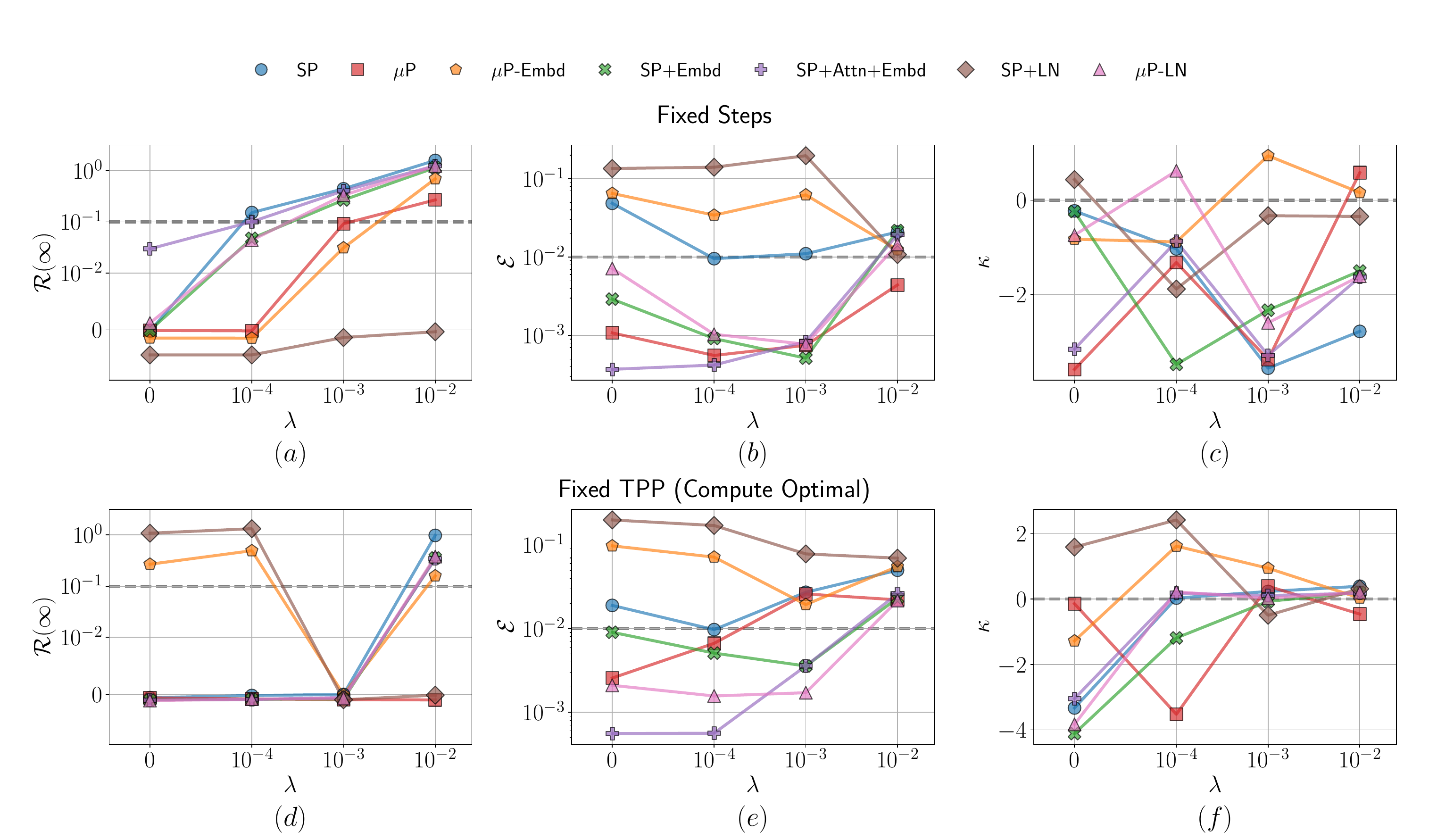}
    \caption{\looseness -1
    \emph{Effect of weight decay on the three transfer metrics in the fixed-step (top) and fixed TPP (bottom) settings.} In the fixed-step setting, weight decay improves predictability error $\mathcal{E}$ at the cost of asymptotic performance $\mathcal{R}(\infty)$. In the TPP setting, stable parameterizations achieve near-zero $\mathcal{R}(\infty)$, and landscape predictability trends are similar, but robustness $\kappa$ degrades as weight decay increases.}
    \label{fig:transfer-framework-phase-diagram-weight-decay}
    \vspace{-0.15 in}
\end{figure}

\looseness -1
\textbf{Weight Decay.}
In the fixed-step setting, weight decay consistently hurts asymptotic performance: $\mathcal{R}(\infty)$ monotonically increases with $\lambda$ across parameterizations, from $\mathcal{R}(\infty) \approx 0.01$ to $\mathcal{R}(\infty) \approx 1$ at large $\lambda$ (\Cref{fig:transfer-framework-phase-diagram-weight-decay}a). 
The effect on loss predictability error $\mathcal{E}$ is more nuanced and depends on the baseline stability of the parameterization without weight decay (\Cref{fig:transfer-framework-phase-diagram-weight-decay}b). 
For $\mu$P and SP+Embd, which already exhibit low $\mathcal{E}$ at $\lambda = 0$, small weight decay further improves predictability before worsening at large $\lambda$.
For SP and SP+LN, which suffer from training instabilities at $\lambda = 0$, weight decay steadily reduces $\mathcal{E}$ but is never sufficient to match the stable parameterizations.
Interestingly, at large $\lambda$, $\mathcal{E}$ converges to $\approx 0.01$ across parameterizations, suggesting that strong weight decay regularizes the landscape to a similar level regardless of parameterization.
The improvement in loss predictability error $\mathcal{E}$ with weight decay has a natural interpretation: weight decay regularizes the loss landscape, reducing its complexity and making it better captured by our loss ansatz (\Cref{eq:full_model}).
The transfer robustness exponent $\kappa$ remains largely negative and shows no clear dependence on $\lambda$ (\Cref{fig:transfer-framework-phase-diagram-weight-decay}c).

\looseness -1
\textbf{Compute Optimal Training.}
In the compute-optimal setting, most parameterizations achieve near-zero $\mathcal{R}(\infty)$ (except for SP+LN and $\mu$P-Embd due to exhibiting training instabilities), suggesting that the loss is not sensitive to the choice of parameterization. 
Predictability error $\mathcal{E}$ follows similar trends to the fixed-step setting, with minor trend differences. 
A notable difference from the fixed-step setting lies in $\kappa$: at $\lambda = 0$, most parameterizations exhibit negative $\kappa$, but this robustness degrades sharply with increasing weight decay, with most parameterizations converging to $\kappa \approx 0$.
A natural explanation is that $\mu$P assumes $\Theta(1)$ training steps relative to width, but in the compute optimal regime, steps $T$ scale as $n^2$, violating this assumption and making the current convention $\eta \cdot \lambda = \Theta(1)$ inadequate. 
In~\Cref{appendix:wd_step_scaling_tpp}, we find that scaling weight decay as $\eta \cdot \lambda = \Theta(\nicefrac{1}{n^2})$ reduces the shift in optimal learning rate, but the shape of the loss curves around the minimum changes across widths.
We leave a detailed analysis of the appropriate weight decay scaling in this regime to future work.

\section{Related Works}
\label{section:related_works}
\looseness -1
Our work is closely related to several recent works on hyperparameter transfer~\cite{everett2024scaling,ghosh2025understandingmechanismsfasthyperparameter,kosson2026weight,bergsma2025power}. 
~\citet{kosson2026weight} argue that weight decay stabilizes feature learning and is central to learning rate transfer. 
Our three transfer metrics provide complementary insights into the role of weight decay: its primary effect to improve the loss predictability error $\mathcal{E}$, but it comes at the cost of increasing the asymptotic loss gap $\mathcal{R}(\infty)$ in the fixed step setting. 
Furthermore, very stable parameterizations such as SP+Embd+Attn achieve reliable transfer even without weight decay, and weight decay alone is insufficient to stabilize unstable parameterizations such as SP+LN.
The analysis of~\citet{bergsma2025power} implies that $\eta \cdot \lambda$ should scale as $\Theta(1/n^2)$ in the compute-optimal setting. 
Our preliminary experiments in this regime suggest that additional scaling considerations may be needed to fully resolve the transfer robustness degradation we observe, leaving the appropriate weight decay scaling as an open question.

\section{Discussion and Conclusion}

In this work, we introduced a quantitative framework for evaluating hyperparameter transfer, with three metrics $\mathcal{R}(\infty)$, $\mathcal{E}$, and $\kappa$ that together serve as a diagnostic lens for identifying what a given transfer setup is lacking.
Our framework generalizes beyond learning rate transfer with width and can be applied to any hyperparameter and scaling dimension. 
For instance, it can help determine which depth scaling strategy yields more reliable transfer, whether learning rate transfer across tokens is as robust as across width, and whether current batch size and expert scaling conventions are brittle.

\looseness -1
Using this framework, we find that under AdamW, the primary advantage of $\mu$P over SP comes from training the embedding layer at a sufficiently fast learning rate. 
This suggests that the full set of $\mu$P conditions is excessive, and practitioners using SP can recover comparable transfer performance by simply correcting the embedding layer learning rate. 
Interestingly, training the embedding layer too slowly not only slows down learning but can also cause training instabilities, which is counterintuitive, as one does not expect a layer trained too slowly to destabilize training. 
Training the embedding layer slowly may therefore be an overlooked source of training instabilities observed in practice.
While our analysis is specific to AdamW, it would be interesting to extend it to other optimizers, such as SGD and Muon~\citep{jordan2024muon}, whose different update geometries may yield different minimal variants analogous to SP+Embd. 
In~\Cref{appendix:mup_derivation} we also examine the weight-tied embedding case, where the embedding and last layer share parameters, and show that a naive SP would require a $\nicefrac{1}{n}$ output multiplier in addition to $\Theta(1)$ embedding layer learning rate for reliable transfer. 
Our results also show that the weight decay scaling convention $\eta \cdot \lambda = \Theta(1)$, derived under $\mu$P's fixed step assumption, is inadequate in the compute optimal regime where the training horizon itself scales as $\Theta({n^2})$.
We leave finding the correct weight decay scaling in this regime to future work.

\looseness -1
\textbf{Limitations.} Our experiments are limited to decoder-only Transformers with fixed depth, scaled to $\sim 1$B parameters, trained with AdamW on a single dataset (FineWeb-Edu). Due to the large hyperparameter sweep, each configuration is run with a single random seed. We leave the analysis of other architectures, optimizers, depth scaling, and datasets to future work.

\section*{Acknowledgments}
We thank Tianyu He, Darshil Doshi, Sean McLeish, John Kirchenbauer, and Tom Goldstein for helpful discussions.
MB and DSK thank the Simons Collaboration on Physics of Learning and Neural Computation (SFI-MPS-POL-00012574-09).
The authors acknowledge the University of Maryland supercomputing resources (\url{http://hpcc.umd.edu}) made available for conducting the research reported in this paper.

{
\small
\bibliography{ref}
\bibliographystyle{plainnat}

}


\newpage
\appendix

\section{Experimental Details}
\label{appendix:experimental_details}

\looseness -1
We pre-trained GPT-style Transformers on FineWeb-Edu~\citep{penedo2024the}, building on the nanoGPT codebase~\citep{karpathy2022nanogpt}.
All experiments use a depth of $12$ Transformer blocks without biases, a context length of $1024$, with the data tokenized using the GPT-2 tokenizer~\cite{Radford2019LanguageMA}, resulting in a vocabulary size of $50, 304$. 
We scale the embedding dimension (width) $n \in [128, 2048]$ by increasing the number of heads, while keeping the head dimension fixed at $d = 64$. 
We train the models using AdamW ($\beta_1 = 0.9, \beta_2 = 0.95, \epsilon = 10^{-8}$) with a Warmup-Stable-Decay (WSD) schedule~\citep{hu2024minicpm}  ($20\%$ warmup, $60\%$ stable, $20\%$ decay). 
For each width, we sweep the peak learning rate and weight decay strength $\lambda$.

In the fixed-step setting, we train the models for $10,000$ steps with a batch size of $1024$ ($\approx 1$M tokens per step), corresponding to a total of approximately $10$B tokens. By comparison, in the compute-optimal setting (fixed token-per-parameter), we scale training tokens proportional to the number of parameters, with a ratio of $20$ tokens per parameter following~\cite{hoffmann2022an}. To ensure that small models are trained for a sufficient number of steps, we reduce the batch size to $256$ ($\approx 0.25$M tokens).

\paragraph{Compute Usage.} All experiments were run on H100 GPUs. The total sweep covers 
$20$ learning rates $\times$ $8$ weight decay values $\times$ $8$ widths $\times$ 
$16$ parameterizations $\times$ $2$ training regimes, with each run taking 
approximately $2$ hours on average, for an estimated total of $\sim$160,000 H100 
GPU hours.

\section{Estimating the Three Transfer Metrics}
\label{appendix:transfer_framework_fitting}

This appendix describes the filtering, interpolation, and fitting procedures used to compute the transfer framework metrics $\mathcal{R}(\infty), \mathcal{E}$, and $ \kappa.$

\begin{figure}[h]
    \centering
    \begin{subfigure}[t]{0.48\linewidth}
        \centering
        \includegraphics[width=\linewidth]{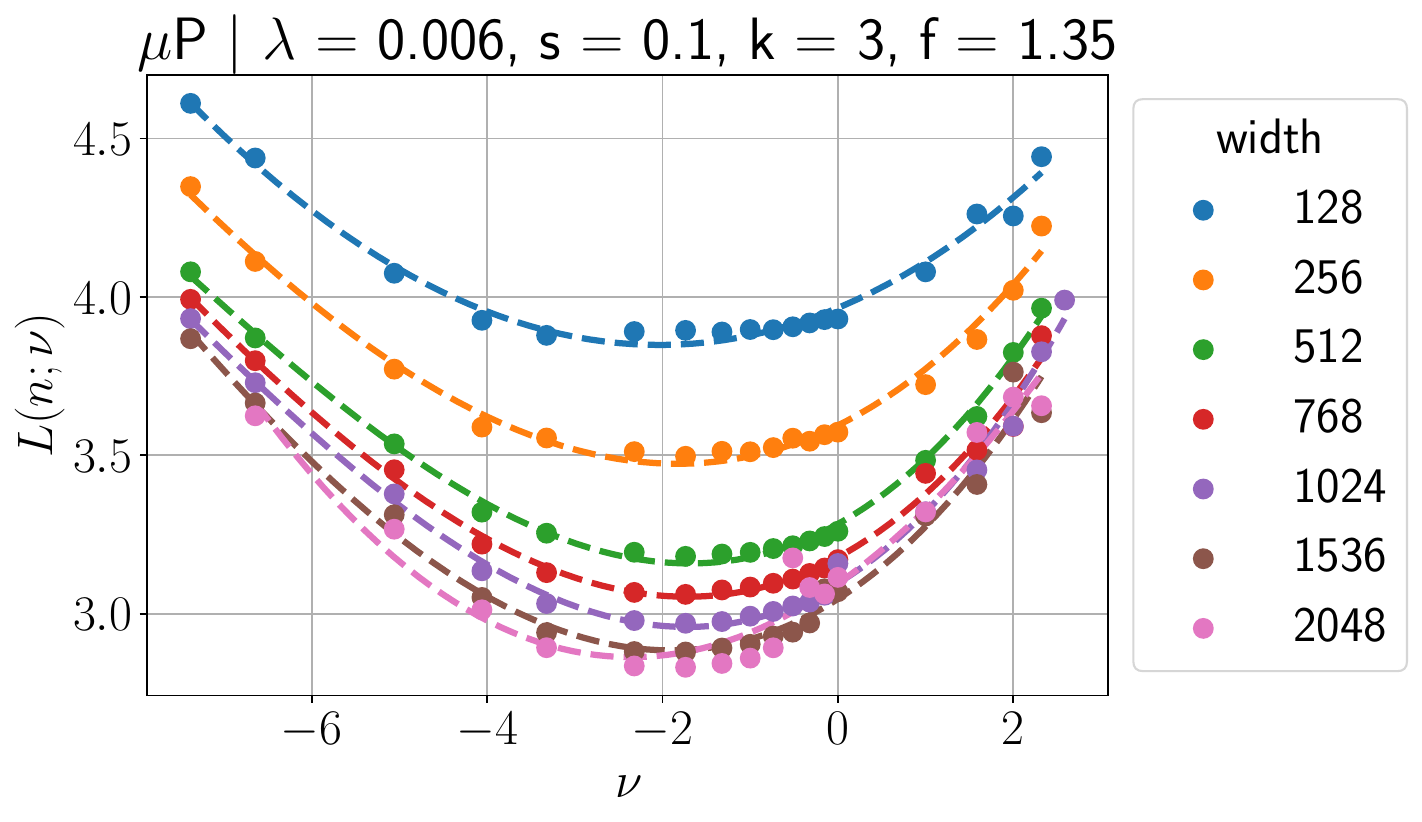}
        \caption{}
        \label{fig:interpolated_curves_mup}
    \end{subfigure}
    \hfill
    \begin{subfigure}[t]{0.48\linewidth}
        \centering
        \includegraphics[width=\linewidth]{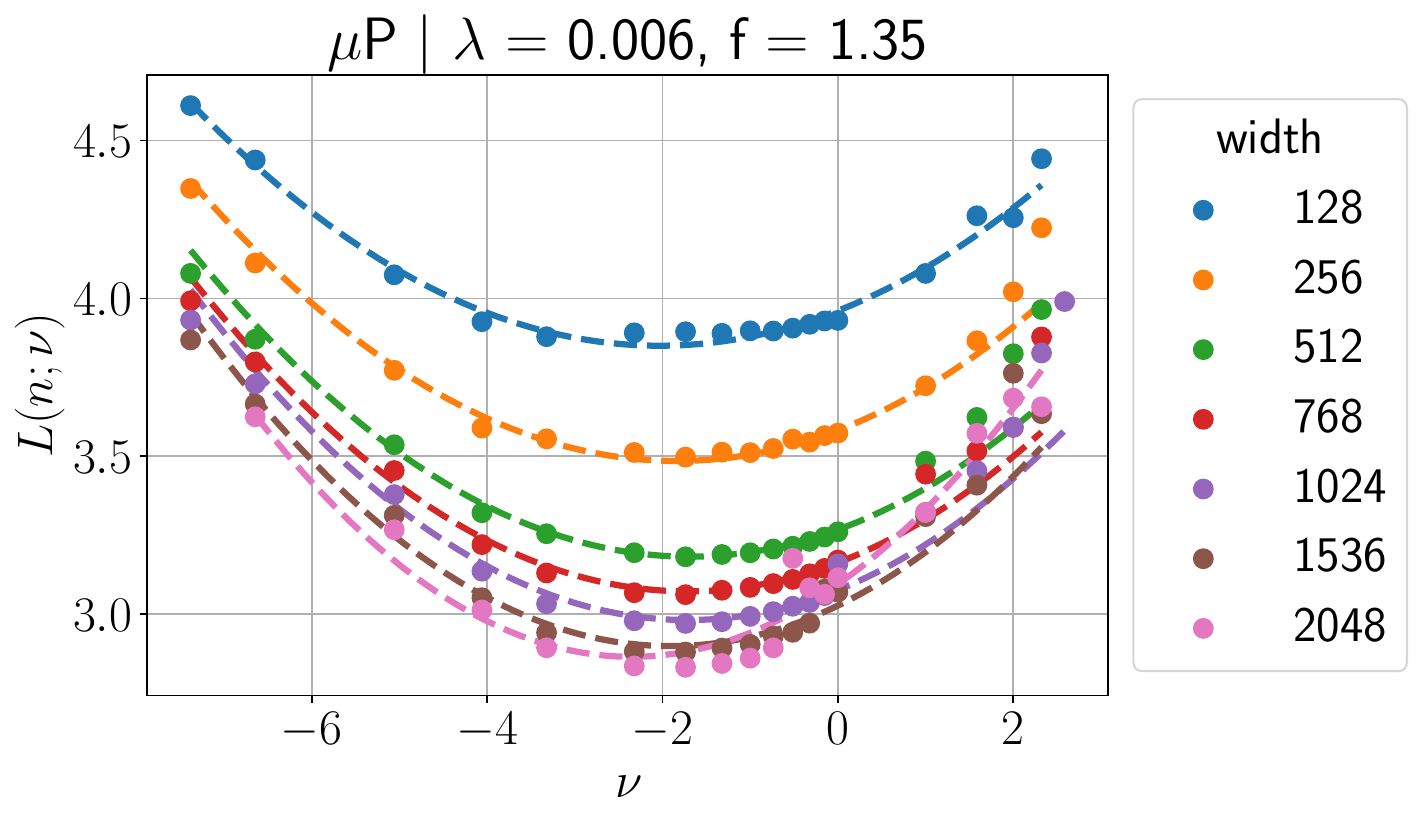}
        \caption{}
        \label{fig:curvature_fits_mup}
    \end{subfigure}
    \caption{(a) Interpolated loss curves for $\mu$P across widths with loss filtering threshold $f = 1.35$. Raw observed points (circles) and the fitted smoothing spline (solid lines) are shown for each width. (b) Per-width curvature fits for $\mu$P. For each width, we show the interpolated loss curve (solid line) and the fitted centered quadratic $L(\nu) = L_{\min} + \frac{1}{2}H(n)(\nu - \nu^*)^2$ (dashed line).}
    \label{fig:interpolation_and_curvature_mup}
\end{figure}

\paragraph{\textbf{Filtering, Smoothing and Interpolation.}}
For each width, we retain runs with loss within $f = 1.35\times$ the per-width optimal loss $L^*(n)$ to focus the analysis on the landscape around the minimum. The threshold $f$ is chosen to avoid unstable and divergent runs while retaining as many data points as possible.
We then fit a cubic spline (\texttt{UnivariateSpline}, degree $k=3$) to the filtered per-width curves with smoothing parameter $S = s \cdot N \cdot \mathrm{Var}(L)$, where $s = 0.1$ is the base smoothing coefficient, $N$ is the number of points, and $\mathrm{Var}(L)$ is the variance of the loss values. Each spline is evaluated on a uniform grid of $400$ points spanning the observed $\nu$ range, resulting in smooth, dense curves for the subsequent analysis (see \Cref{fig:interpolated_curves_mup} for an example). Out of $128$ total combinations ($16$ parameterizations $\times$ $8$ weight decay values), $4$ failed this procedure either due to too few points after the filtering step or the interpolated loss curve exhibiting excessive noise resulting in negative loss values, and were therefore excluded from the analysis.

\paragraph{\textbf{Estimating the Three Transfer Metrics.}}
By fitting scaling laws to $L^*(n)$ and $\nu^*(n)$, we obtain the exponents $\alpha$, $\beta$ and the asymptotic loss $L^*(\infty)$, from which we compute $\mathcal{R}(\infty)$. 
The loss scaling law is fit in log space with constraints $L^*(\infty), A, \alpha \geq 0$, ensuring the fitted scaling law strictly decreases with width and converges to a finite irreducible value. 
The log-learning-rate scaling law is fit in linear space, since $\nu$ is already in log space, with $\beta \geq 0$ enforced to ensure convergence to the asymptotic value. 
Since $\mathcal{R}(\infty)$ is non-negative by definition (\Cref{def:loss-degradation}), we clamp fitted values to zero when they are negative due to finite-size fitting artifacts.
For the curvature $H(n)$, we first fit a centered quadratic $L(\nu) = L^* + \frac{1}{2} H(n) (\nu - \nu^*(n))^2$ to the interpolated per-width curves (see \Cref{fig:curvature_fits_mup}), with the center fixed at $\nu^*(n)$ to prevent noise in the loss curve from shifting the fitted center away from the true optimum.
We then fit the scaling law $H(n) = Cn^\gamma$ (\Cref{fig:transfer_curves_mup}(f)) to obtain the exponent $\gamma$. 
Combined with $\alpha$ and $\beta$ from above, this gives $\kappa = \alpha - 2\beta + \gamma$.
Finally, we jointly fit all the parameters of \Cref{eq:full_model} using the interpolated curves, and evaluate the fit on the raw filtered data to obtain the loss predictability error $\mathcal{E}$.

\looseness -1
We fit all scaling laws using a Huber loss objective with $\delta = 10^{-3}$ to improve robustness to outliers and $200$ random initializations to avoid local minima in the non-convex scaling law fits.
We additionally cap all scaling exponents at $2.0$, which serves as a practical proxy for exponents running to infinity and prevents spuriously large values that can make comparisons across parameterization unreliable.

\begin{figure}[t]
    \centering
    \begin{subfigure}[b]{0.35\textwidth}
        \centering
        \includegraphics[width=\textwidth]{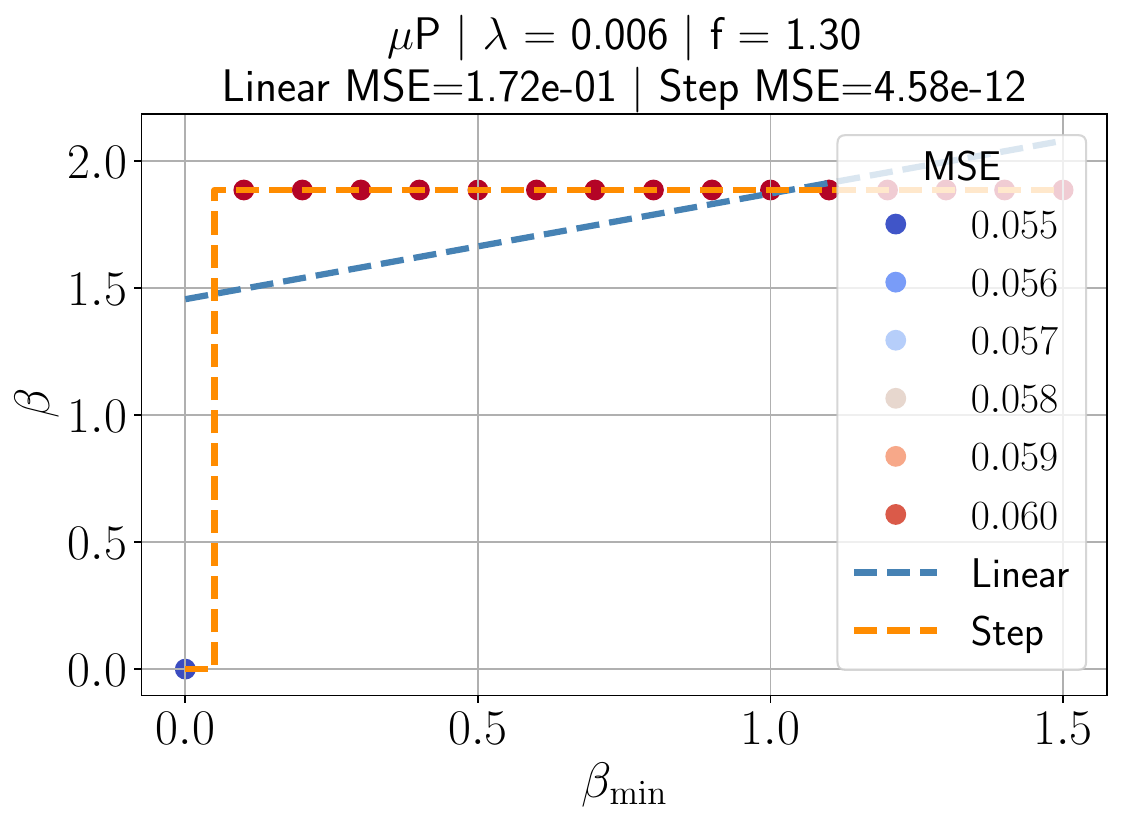}
        \caption{}
    \end{subfigure}
    \begin{subfigure}[b]{0.35\textwidth}
        \centering
        \includegraphics[width=\textwidth]{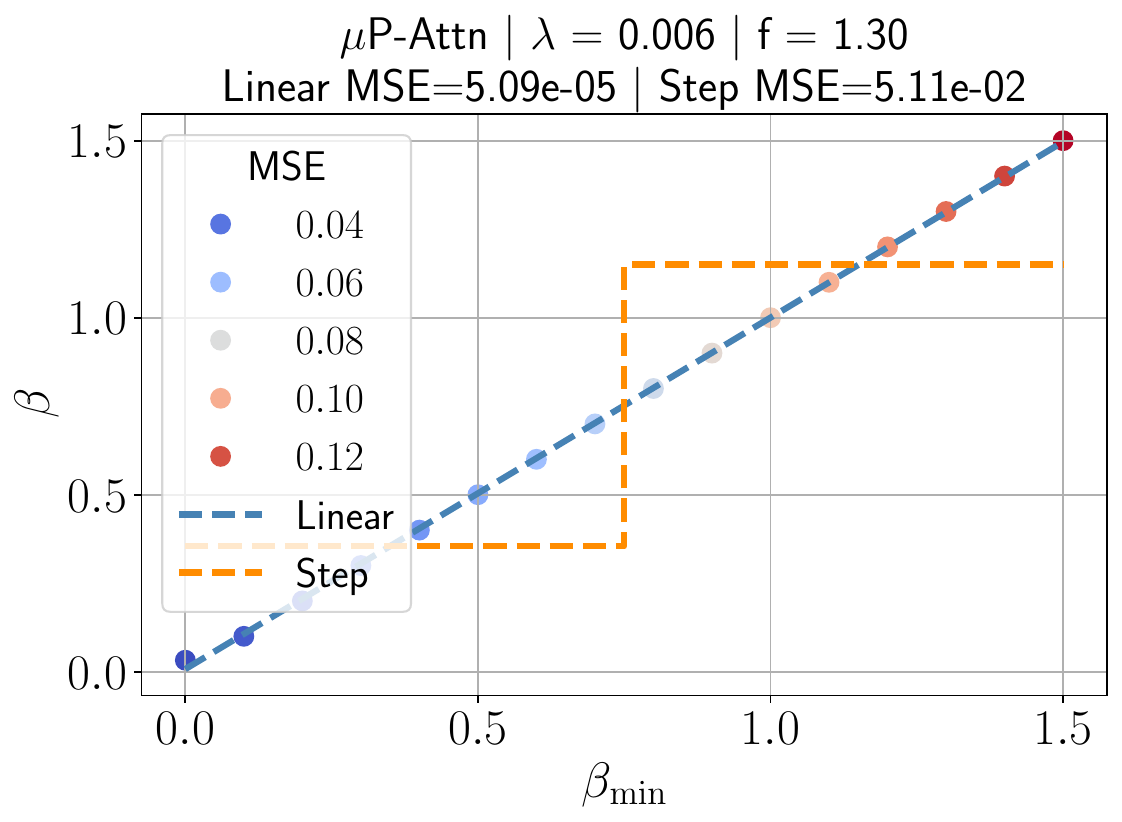}
        \caption{}
        \label{fig:degeneracy_genuine}
    \end{subfigure}
    \caption{Fitted $\beta$ as a function of the lower bound $\beta_{\min}$ for two cases. Left: a degenerate case ($\mu$P) where the step function better fits the observed trend, and we adopt the optimal solution with $\beta > \beta_{\min}^*$. Right: a genuine case ($\mu$P-Attn) where $\beta$ monotonically increases with $\beta_{\min}$, suggesting that the fitted small $\beta$ can be trusted.}
    \label{fig:degeneracy_examples}
\end{figure}

\paragraph{\textbf{Degeneracy in $\nu^*$ Scaling.}}
A subtle issue arises when $\nu^*(n)$ is nearly constant across widths. 
In this regime, the scaling law $\nu^*(n) = \nu^*(\infty) + Bn^{-\beta}$ admits two degenerate solutions: (1) $\beta \to 0$, reducing the model to a constant $\nu^*(\infty) + B$ and (2) $\beta \to \infty$, indicating rapid convergence to $\nu^*(\infty)$. 
We prefer the $\beta \to \infty$ solution, as $\beta \to 0$ reduces the scaling law to a constant and is inconsistent with the convergence-based interpretation of our framework.
Since $\beta$ may not be exactly zero numerically, distinguishing a genuinely small $\beta$ from the degenerate $\beta \to 0$ solution is non-trivial. 
To resolve this, we repeatedly fit the scaling law with an increasing lower bound $\beta_{\min}$ on $\beta$ and track how the fitted $\beta$ changes. 
If the true solution is degenerate, the fitted $\beta$ will exhibit a sudden jump at some $\beta^*_{\min}$, beyond which random initializations begin to prefer the $\beta \to \infty$ solution. 
In contrast, if the solution is not degenerate, the fitted $\beta$ will linearly increase from $\beta_{\min}$ (see \Cref{fig:degeneracy_examples}).
We then fit both a step function and a linear function to the resulting $\beta$ vs. $\beta_{\min}$ trend. If the step function fits better, we have identified the small $\beta$ as a degenerate solution and select the best solution with $\beta > \beta^*_{\min}$. Otherwise, we treat the small observed $\beta$ as genuine.
We apply this procedure only to the $\nu^*(n)$ scaling law fits to estimate $\beta$ and hence $\kappa$. 

\paragraph{\textbf{Degeneracy in the Full Model.}}
The full loss model (\Cref{eq:full_model}) introduces additional degeneracies beyond those in the individual $\nu^*$ scaling law. Specifically, the term $\frac{1}{2}Cn^\gamma(\nu - \nu^*(\infty) - Bn^{-\beta})^2$ couples the curvature scaling parameters $C$ and $\gamma$ with the $\nu$ scaling parameters $B$ and $\beta$ through the product $C n^\gamma \cdot B^2n^{-2\beta}$, making individual parameter estimates unreliable even when the overall fit is good. We therefore do not use the joint fit for estimating $\beta$, $\kappa$, and $\mathcal{R}(\infty)$, and instead rely on the individually fitted scaling laws. The parameters of the full model are constrained in the same way as the individual scaling laws, with $L^*(\infty), A, \alpha, C, \beta \geq 0$ and $B$ unconstrained. Nevertheless, the joint fit consistently yields smaller fit errors than substituting the individually fitted parameters directly into \Cref{eq:full_model}, and we therefore use it for reporting $\mathcal{E}$.

\clearpage
\newpage

\section{Transfer Metrics for Additional Parameterizations}
\label{appendix:transfer_framework_add_on}

\begin{figure}[!htb]  
\vspace{-0.15in}
    \centering
    \includegraphics[width=0.9\textwidth]{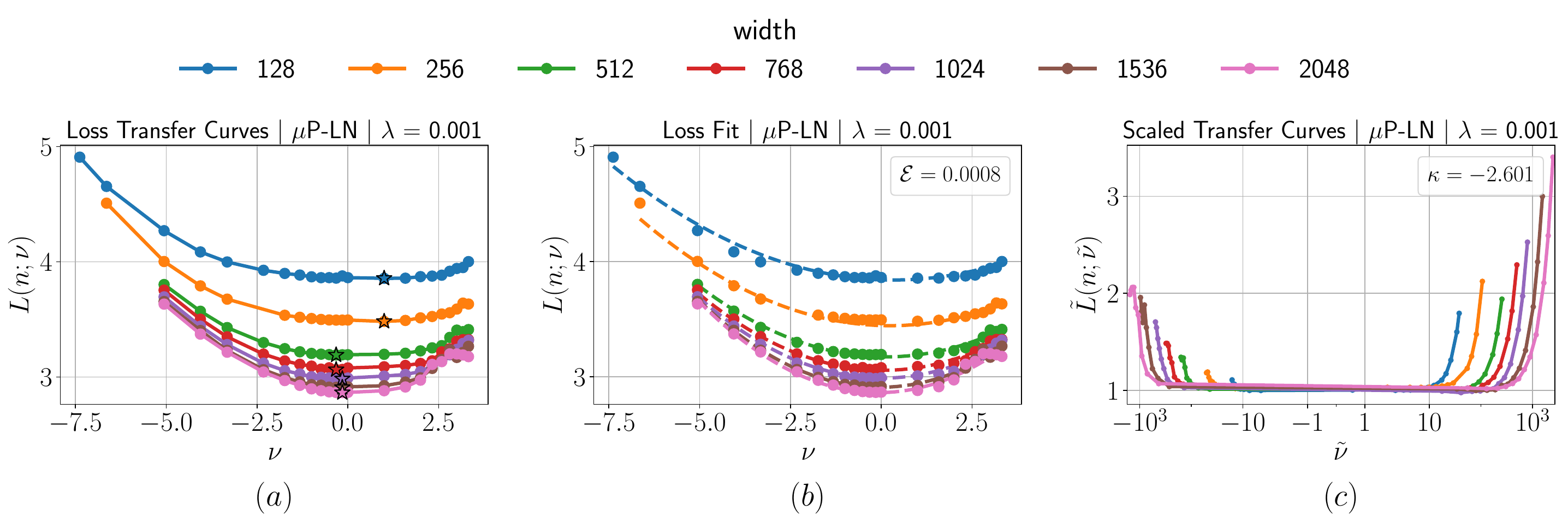}        \includegraphics[width=0.9\textwidth]{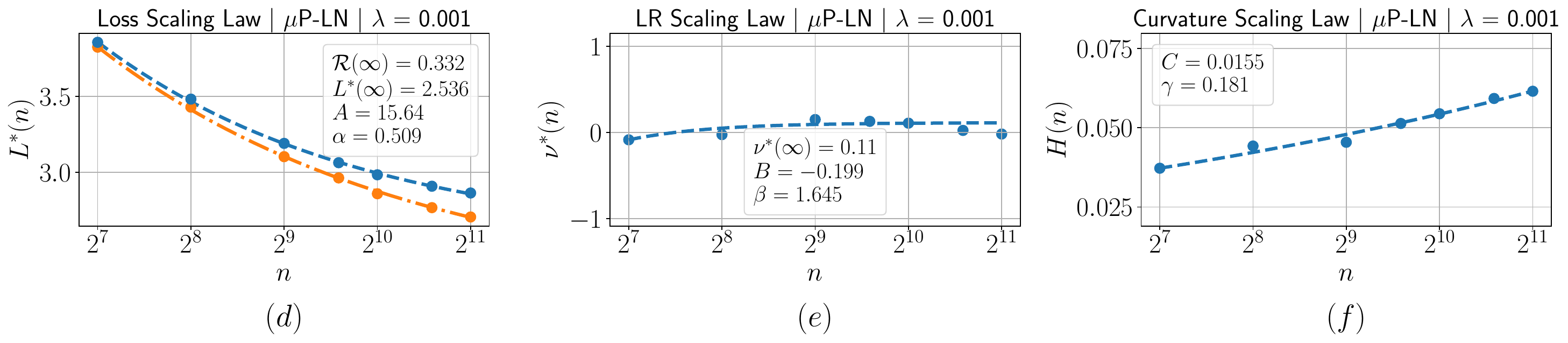}
    \caption{Transfer metrics for $\mu$P-LN with weight decay $\lambda = 0.001$. }
    \label{fig:transfer_curves_mup_sp_ln}
\end{figure}

\begin{figure}[!htb]  
\vspace{-0.15in}
    \centering
    \includegraphics[width=0.9\textwidth]{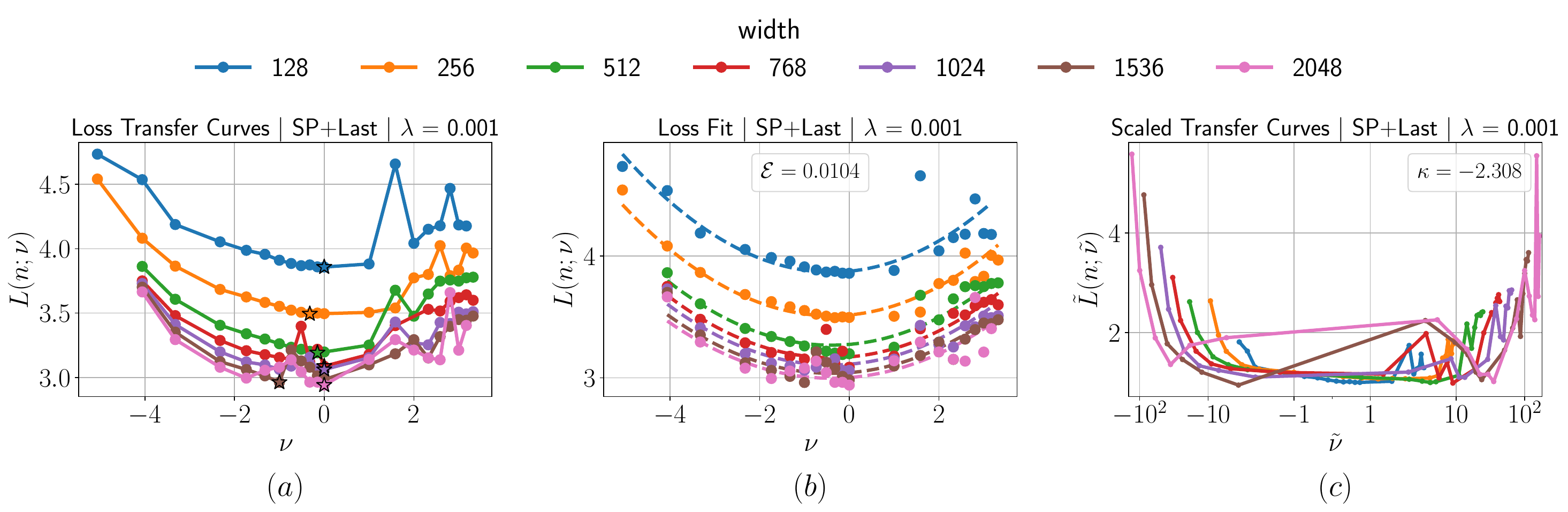}        \includegraphics[width=0.9\textwidth]{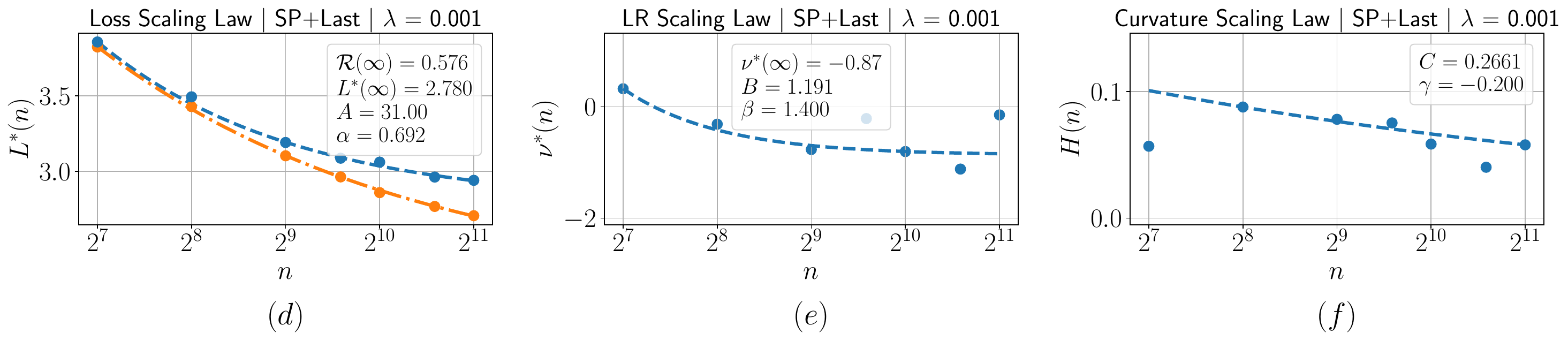}
    \caption{Transfer metrics for SP+Last with weight decay $\lambda = 0.001$. }
    \label{fig:transfer_curves_sp_last}
\end{figure}

\begin{figure}[!htb]  
\vspace{-0.15in}
    \centering
    \includegraphics[width=0.9\textwidth]{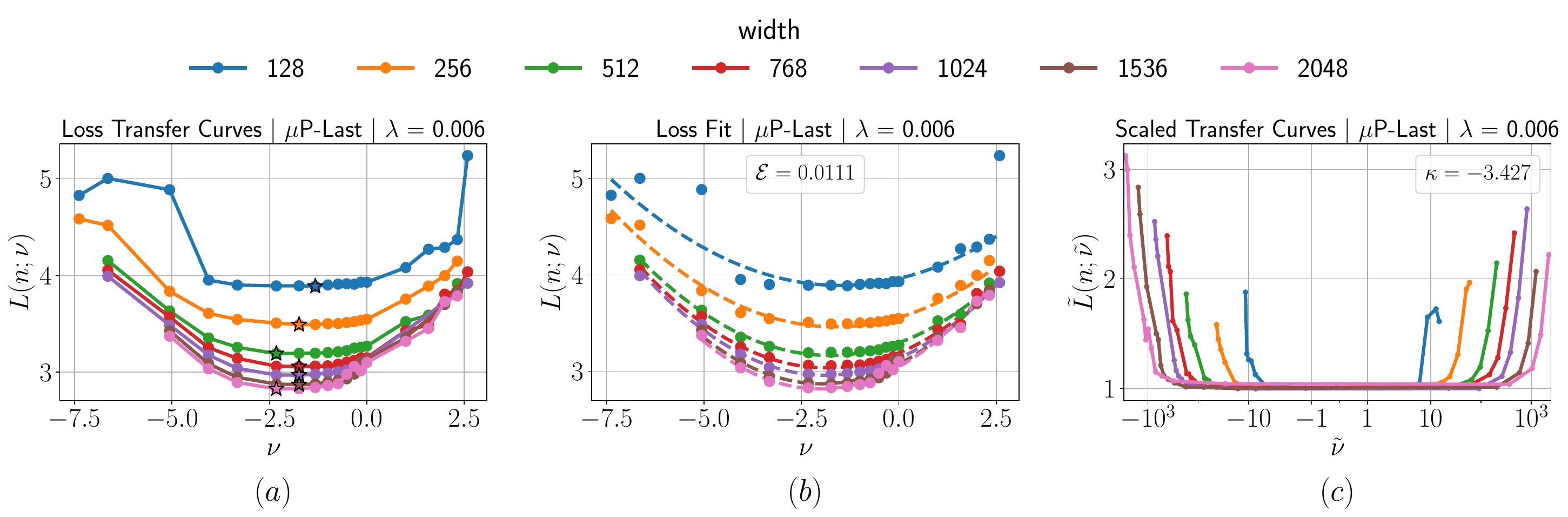}        \includegraphics[width=0.9\textwidth]{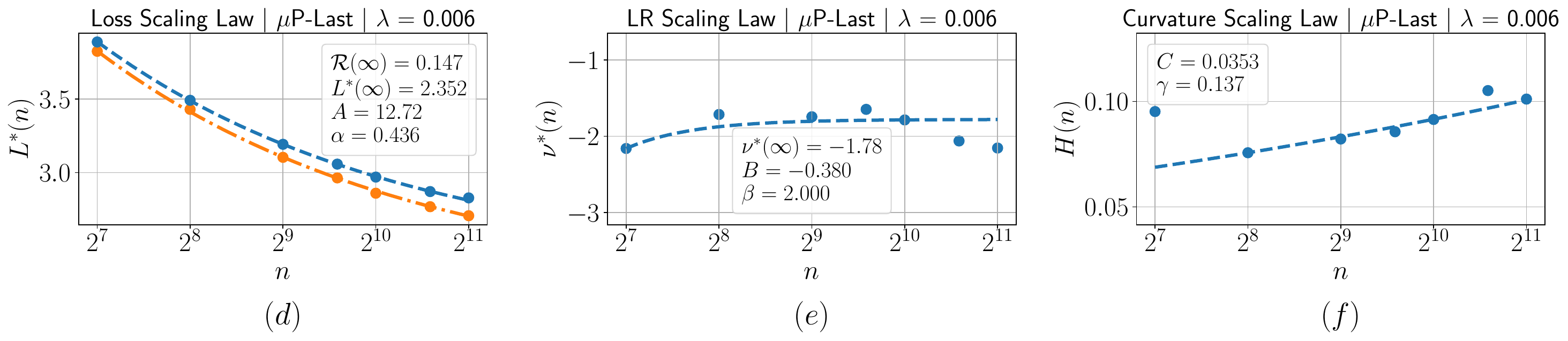}
    \caption{Transfer metrics for $\mu$P-Last with weight decay $\lambda = 0.006$. }
    \label{fig:transfer_curves_mup_last}
\end{figure}

In this section, we analyze the three transfer metrics for all parameterizations interpolating between SP and $\mu$P in further detail. ~\Cref{fig:transfer_curves_mup_appendix}--\ref{fig:transfer_curves_mup_last} show the loss and scaling law curves for each parameterization. 
We use `$+$' to denote incremental changes from SP towards $\mu$P and `$-$' to denote changes from $\mu$P towards SP.
For each parameterization, we select the weight decay value that best represents its typical behavior.

\subsection{Effect of Different Layer Types on Transfer Metrics}

\begin{figure}[!htb]
    \vspace{-0.15 in}
    \centering
    \includegraphics[width=0.9\textwidth]{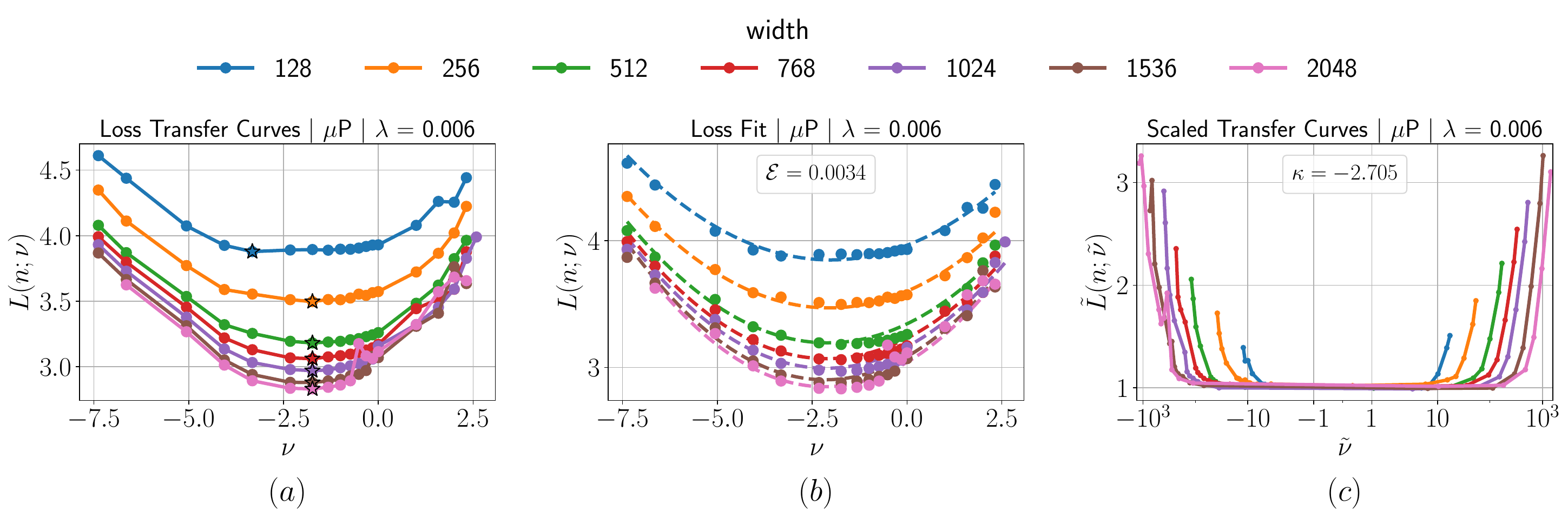}        \includegraphics[width=0.9\textwidth]{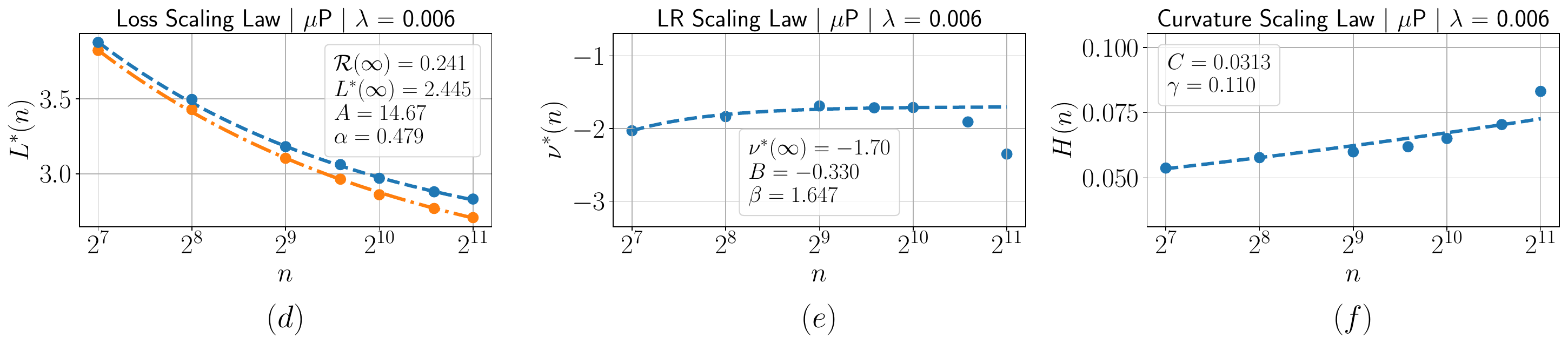}
    \caption{
    \looseness -1
    Transfer metrics for $\mu$P with weight decay $\lambda = 0.006$. Repeated \Cref{fig:transfer_curves_mup} for completeness.}
    \label{fig:transfer_curves_mup_appendix}
\end{figure}

\begin{figure}[!htb]  
    \vspace{-0.15in}
    \centering
    \includegraphics[width=0.9\textwidth]{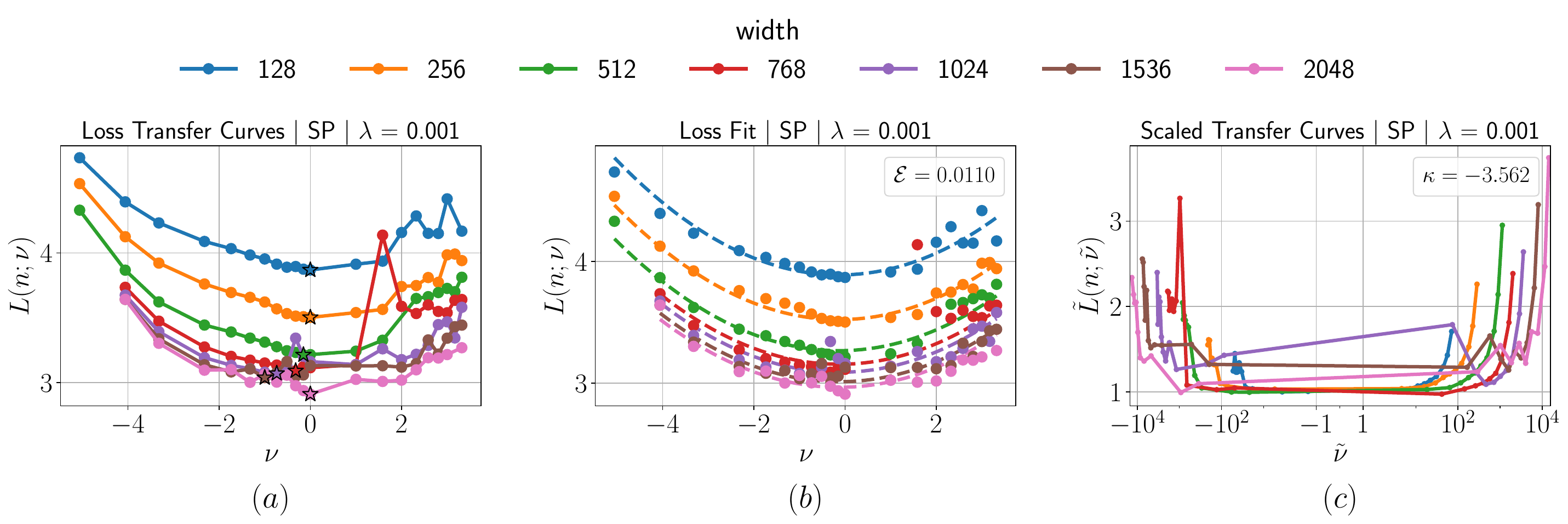}        \includegraphics[width=0.9\textwidth]{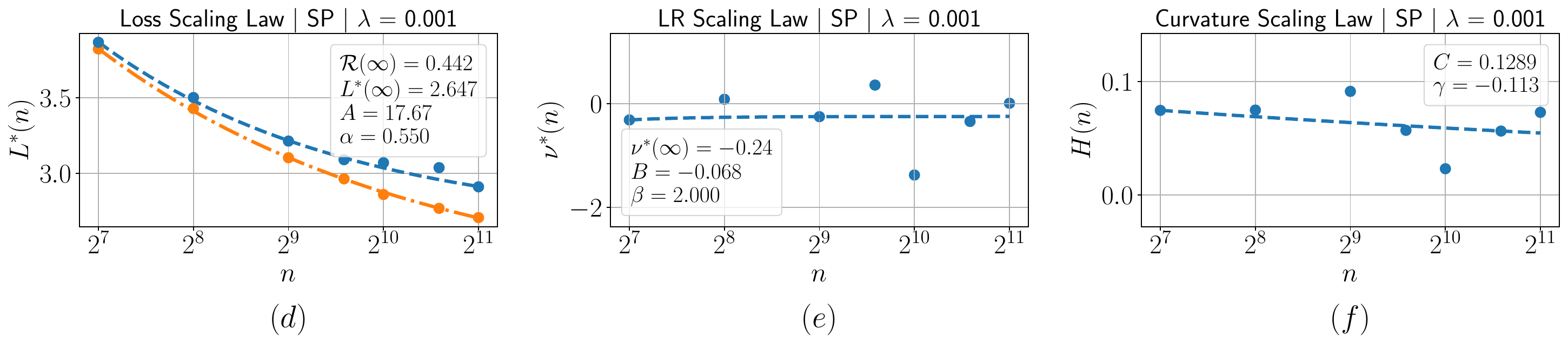}
    \caption{Transfer metrics for SP with weight decay $\lambda = 0.001$. }
    \label{fig:transfer_curves_sp}
\end{figure}

\paragraph{SP vs. $\mu$P.} As described in \Cref{section:mup_sp}, $\mu$P and SP are surprisingly similar across most metrics (\Cref{fig:transfer_curves_mup_appendix,fig:transfer_curves_sp}). Despite being more unstable (a $3\times$ larger loss predictability error), SP exhibits a large negative robustness exponent $\kappa = -3.505$. SP thus has the right ingredients for reliable transfer but is held back by training instability.

\begin{figure}[!htb]   
    \vspace{-0.15in}
    \centering
    \includegraphics[width=0.9\textwidth]{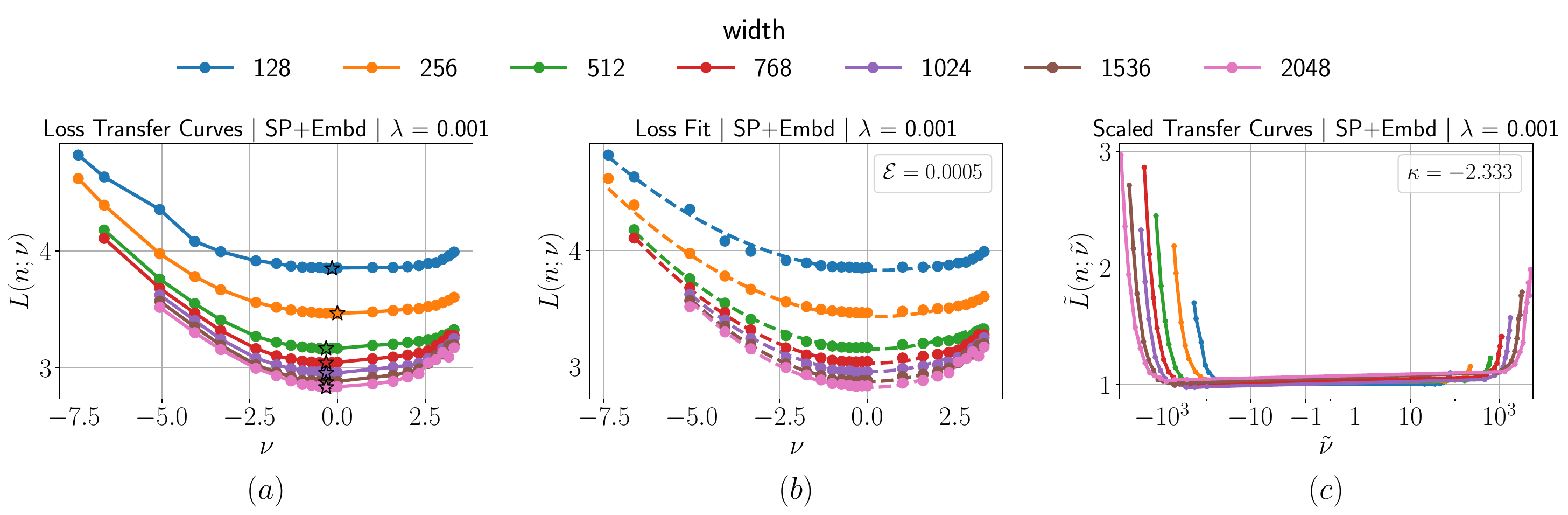}        \includegraphics[width=0.9\textwidth]{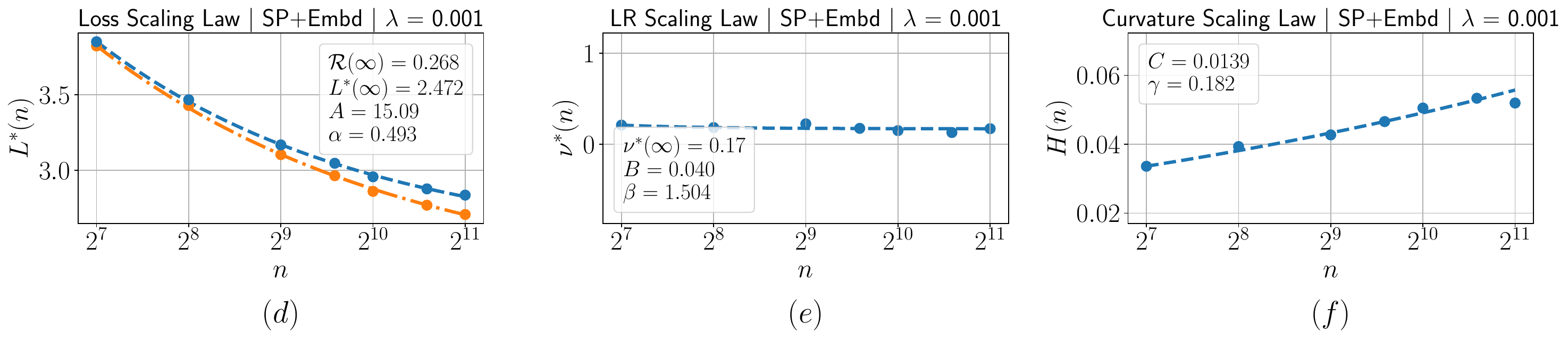}
    \caption{Transfer metrics for SP+Embd with weight decay $\lambda = 0.001$. }
    \label{fig:transfer_curves_sp_embd}
\end{figure}

\begin{figure}[!htb]  
\vspace{-0.15in}
    \centering
    \includegraphics[width=0.9\textwidth]{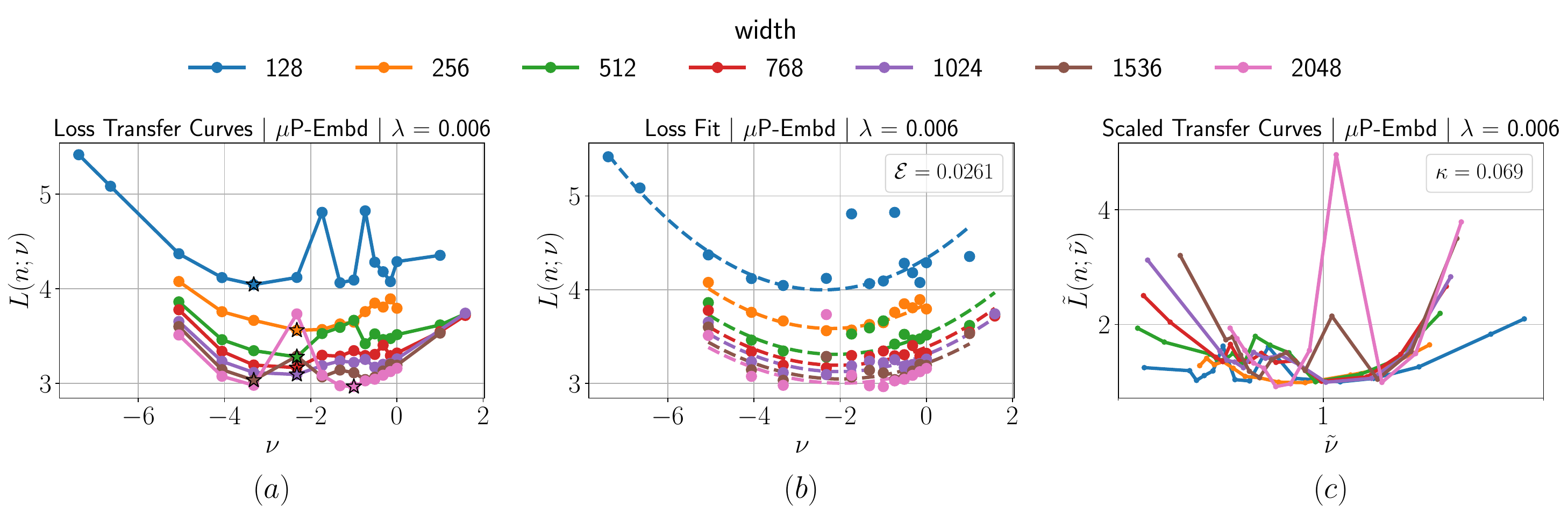}        \includegraphics[width=0.9\textwidth]{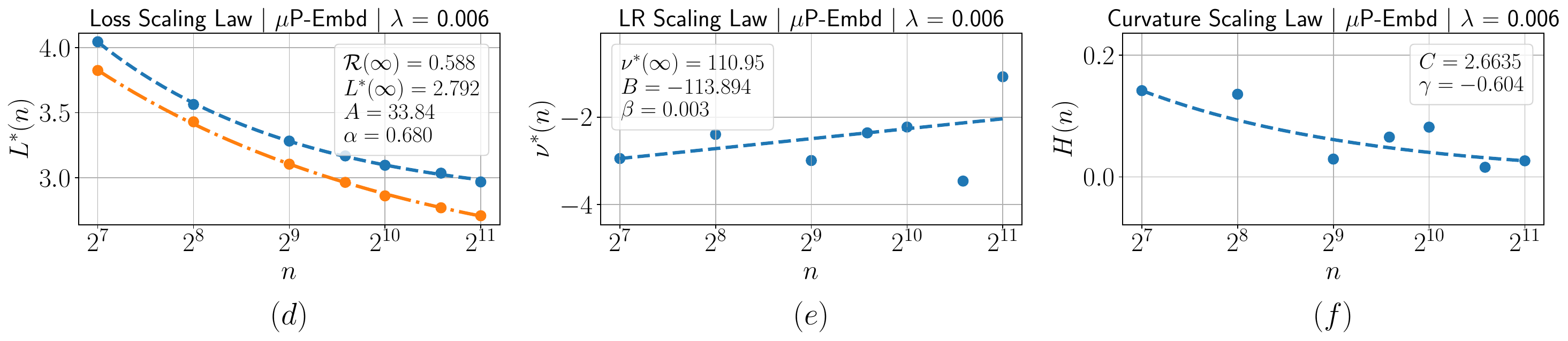}
    \caption{Transfer metrics for $\mu$P-Embd with weight decay $\lambda = 0.006$. }
    \label{fig:transfer_curves_mup_sp_embd}
\end{figure}

\paragraph{Embedding Layer Learning Rate.}
The embedding layer learning rate has the most pronounced effect on stability. Training SP with a $\Theta(1)$ embedding learning rate (SP+Embd) eliminates the instabilities, resulting in smooth loss curves (\Cref{fig:transfer_curves_sp_embd}). Conversely, training $\mu$P with a $\Theta(\nicefrac{1}{n})$ embedding learning rate ($\mu$P-Embd) destabilizes training entirely (\Cref{fig:transfer_curves_mup_sp_embd}).

\begin{figure}[!htb]  
\vspace{-0.15in}
    \centering
    \includegraphics[width=0.9\textwidth]{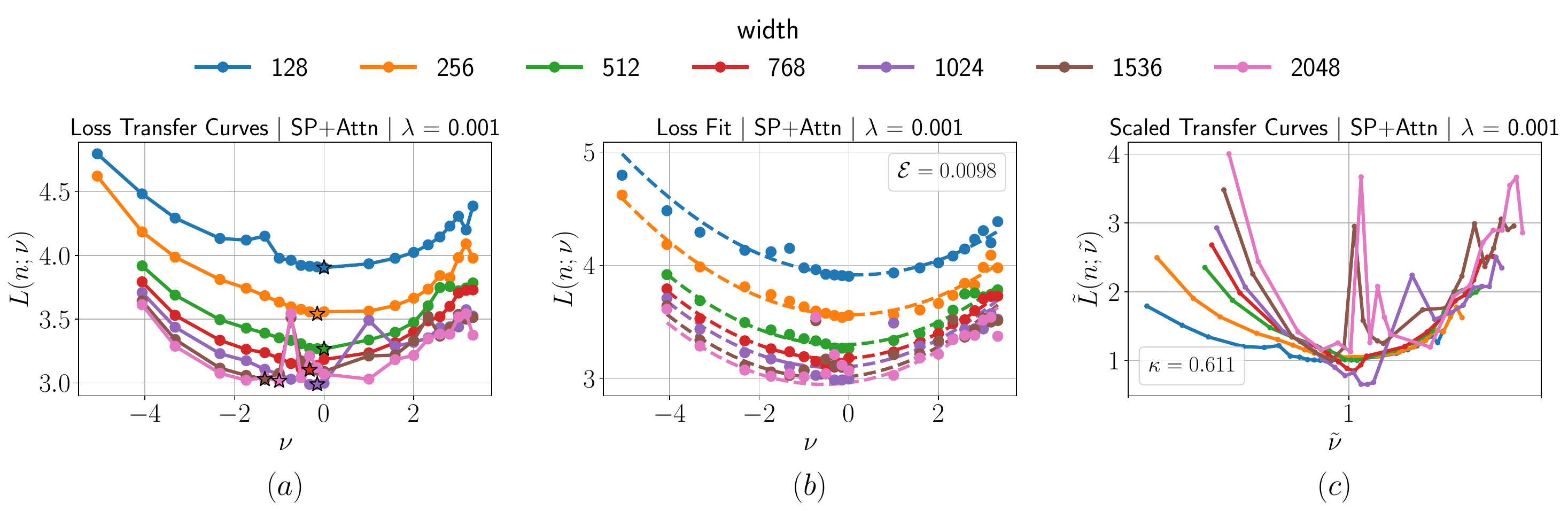}        \includegraphics[width=0.9\textwidth]{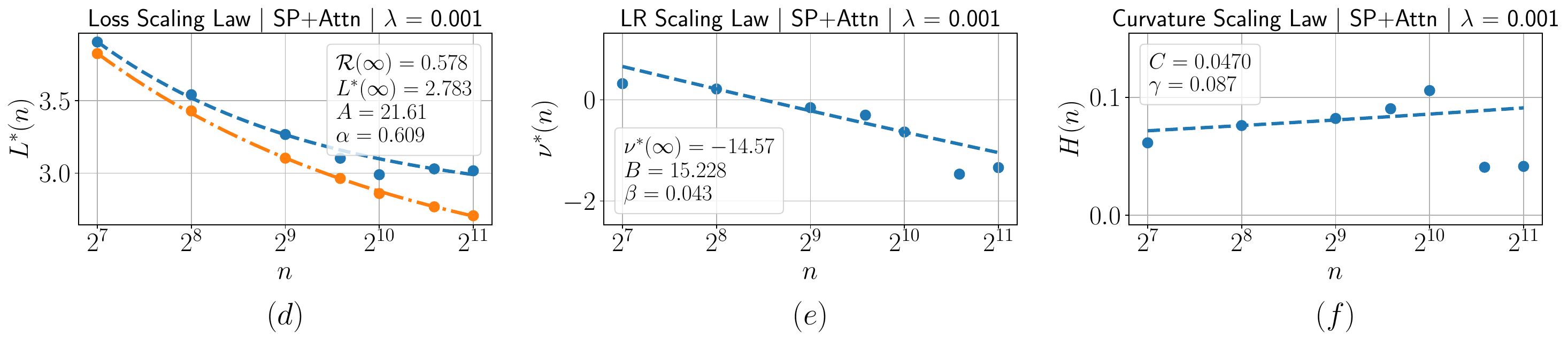}
    \caption{Transfer metrics for SP+Attn with weight decay $\lambda = 0.001$. }
    \label{fig:transfer_curves_sp_attn}
\end{figure}

\begin{figure}[!htb]  
\vspace{-0.15in}
    \centering
    \includegraphics[width=0.9\textwidth]{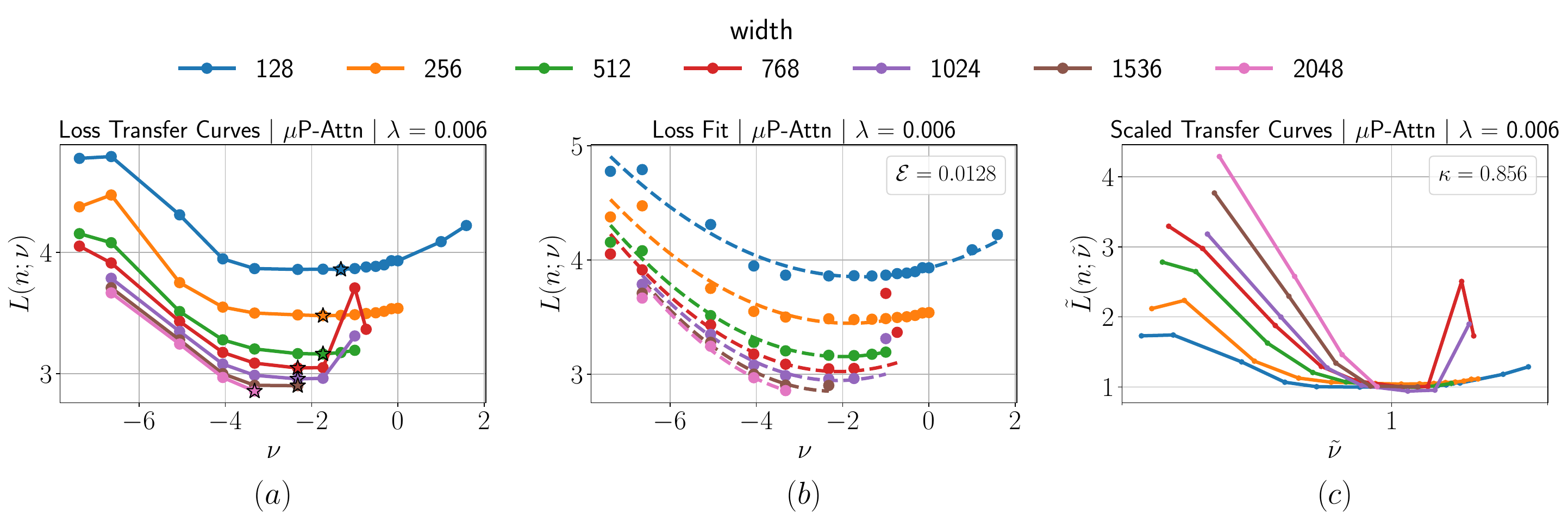}        \includegraphics[width=0.9\textwidth]{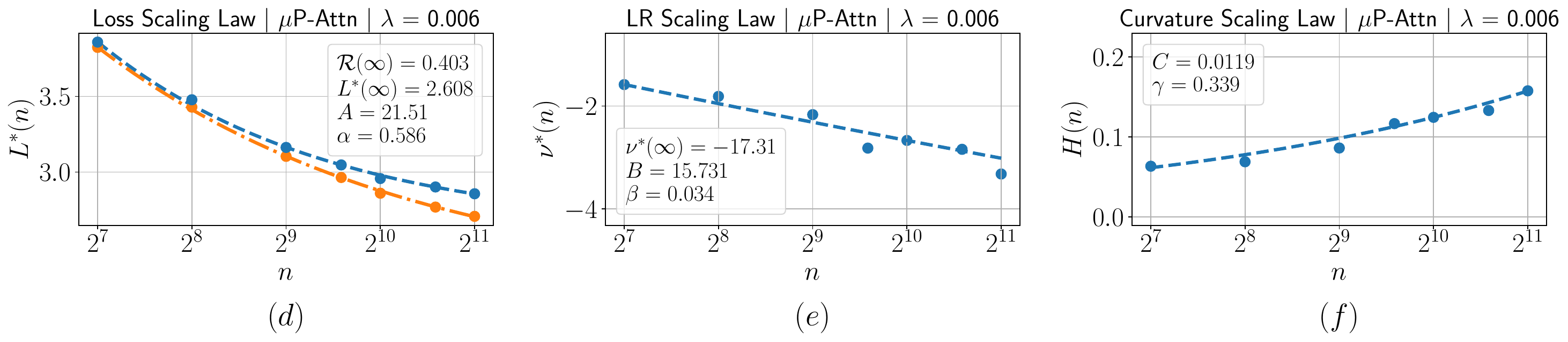}
    \caption{Transfer metrics for $\mu$P-Attn with weight decay $\lambda = 0.006$. }
    \label{fig:transfer_curves_mup_sp_attn}
\end{figure}

\paragraph{Attention Scaling.}
The effect of $\nicefrac{1}{d}$ attention scaling is more subtle. Training SP with $\nicefrac{1}{d}$ attention scaling (SP+Attn) does not improve stability (\Cref{fig:transfer_curves_sp_attn}): the loss predictability error $\mathcal{E}$ remains similar to SP, $\mathcal{R}(\infty)$ is slightly worse, and instabilities worsen at large widths. Most notably, $\kappa$ becomes positive ($\kappa = 0.509$), indicating that transfer becomes brittle. A similar degradation is observed when training $\mu$P with $\nicefrac{1}{\sqrt{d}}$ attention scaling ($\mu$P-Attn) (\Cref{fig:transfer_curves_mup_sp_attn}): $\mathcal{E}$ is similar to SP, the optimal learning rate decreases with width, and $\kappa = 0.881$, making transfer unreliable. 
These results suggest that the attention scaling has a more subtle effect depending on the scaling of other layers.

\begin{figure}[!htb]  
\vspace{-0.15in}
    \centering
    \includegraphics[width=0.9\textwidth]{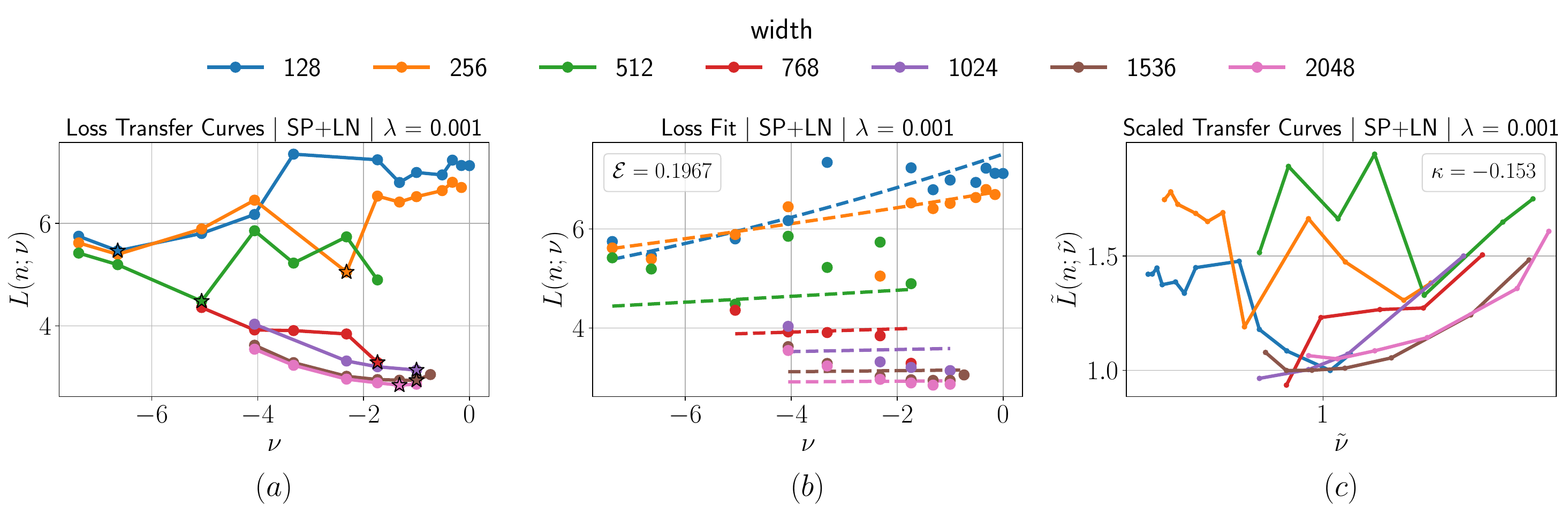}        \includegraphics[width=0.9\textwidth]{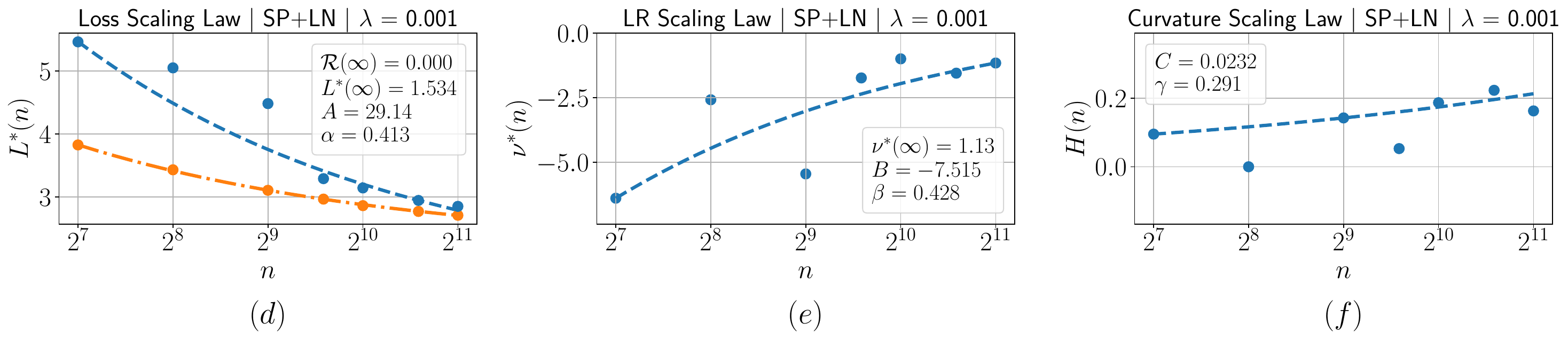}
    \caption{Transfer metrics for SP+LN with weight decay $\lambda = 0.001$. }
    \label{fig:transfer_curves_sp_ln}
\end{figure}

\paragraph{LayerNorm Learning Rate.}
Increasing the LayerNorm learning rate to $\Theta(1)$ in SP (SP+LN) severely destabilizes training (\Cref{fig:transfer_curves_sp_ln}), with $\mathcal{E}$ roughly $10\times$ larger than SP. Interestingly, while the loss at small widths is poor, SP+LN achieves the best asymptotic loss across all parameterizations at large widths. This highlights a sharp tradeoff: SP+LN can in principle reach excellent performance at scale, but its unpredictable loss landscape makes transfer unreliable\textemdash a high-reward but high-risk parameterization. By comparison, decreasing the LayerNorm learning rate to $\Theta(\nicefrac{1}{n})$ in $\mu$P ($\mu$P-LN) has surprisingly almost no effect. The loss predictability error $\mathcal{E}$ is slightly better than $\mu$P, while the asymptotic loss is slightly worse. The consistent improvement in loss predictability from slowing down LayerNorm training observed in both SP+LN and $\mu$P-LN suggests that LayerNorm parameters are best trained slowly. This again reflects a tradeoff: slower LayerNorm training stabilizes the training dynamics at the cost of a small performance penalty. 

\paragraph{Last Layer Initialization.}
Finally, reducing the variance of the last-layer initialization to $\Theta(\nicefrac{1}{n^2})$ in SP has a negligible effect, with all three metrics remaining comparable to SP. By comparison, increasing it to $\nicefrac{1}{n}$ in $\mu$P ($\mu$P-Last) causes training instabilities at small widths, but transfer remains robust at larger widths. The minimal role of last layer initialization is consistent with the observation that a larger initialization variance leads to higher Hessian sharpness, but learning rate warmup gradually reduces this sharpness early in training~\citep{kalra2024why}, effectively mitigating any initialization differences.

\clearpage

\subsection{Transfer Metric Phase Diagrams}

\begin{figure}[!htb]
    \centering
    \includegraphics[width=\linewidth]{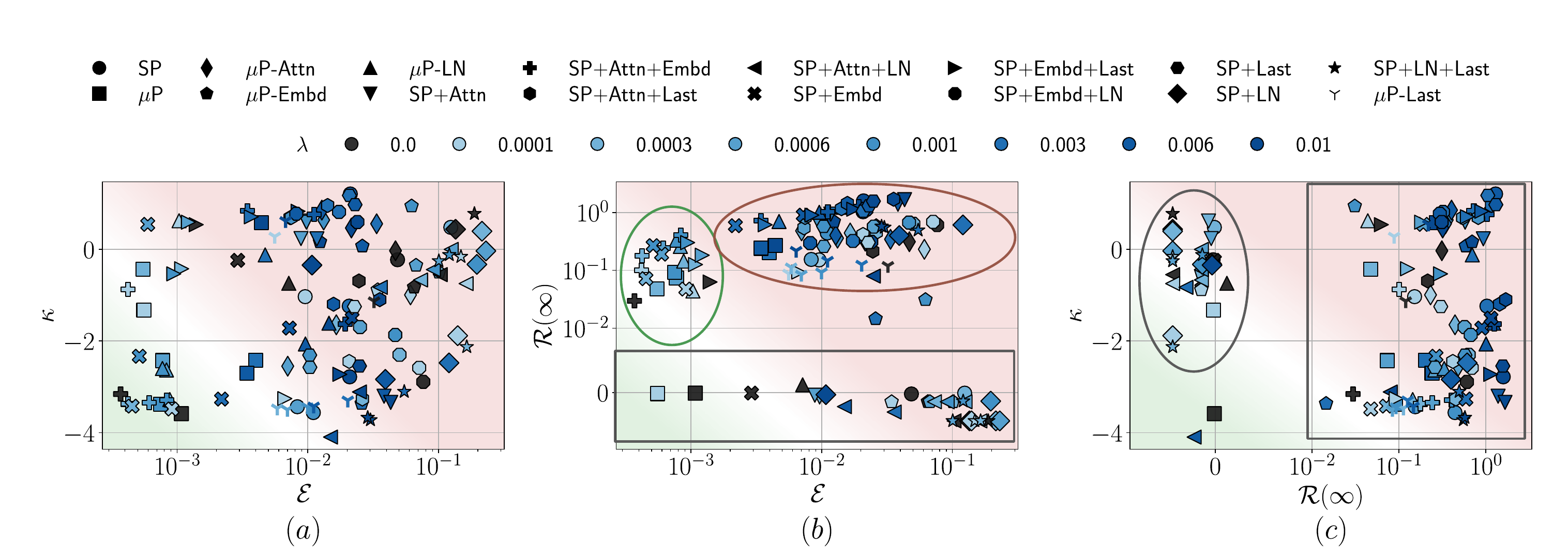}
    \caption{\emph{Transfer metrics for parameterizations interpolating between SP and $\mu$P.}
    Parameterizations with `$+$' denote incremental changes from SP towards $\mu$P, while `$-$' denotes changes from $\mu$P towards SP. Green and red regions indicate desirable and undesirable regimes, respectively.}
    \label{fig:transfer-framework-phase-diagram-full}
\end{figure}

\Cref{fig:transfer-framework-phase-diagram-full} shows the three transfer metrics for all $16$ parameterizations and $8$ weight decay values. 
The metrics naturally separate parameterizations into clusters. 
For example, in panel (b), parameterizations that train the embedding well ($\mu$P, SP+Embd, $\mu$P-LN, SP+Attn+Embd) cluster in the desirable low $\mathcal{E}$, low $\mathcal{R}(\infty)$ region, while unstable parameterizations (SP, SP+LN, $\mu$P-Embd) achieve near-zero $\mathcal{R}(\infty)$ but at the cost of high $\mathcal{E}$. This illustrates the tradeoff between the metrics noted in \Cref{section:mup_sp}: a parameterization can achieve excellent asymptotic performance while remaining unreliable for transfer. Large weight decay values are located in the top right, improving $\mathcal{E}$ relative to the unstable cluster but at an asymptotic performance cost, never reaching the stable cluster. Similar clustering patterns emerge in panels (a) and (c), with stable parameterizations consistently occupying the desirable regions across all three metrics.

\clearpage
\newpage

\section{Importance of the First Layer Learning Rate in CNN Image Classification}
\label{appendix:cnn_cifar_adam}

\begin{figure}[!htb]
    \centering
    \includegraphics[width=0.7\linewidth]{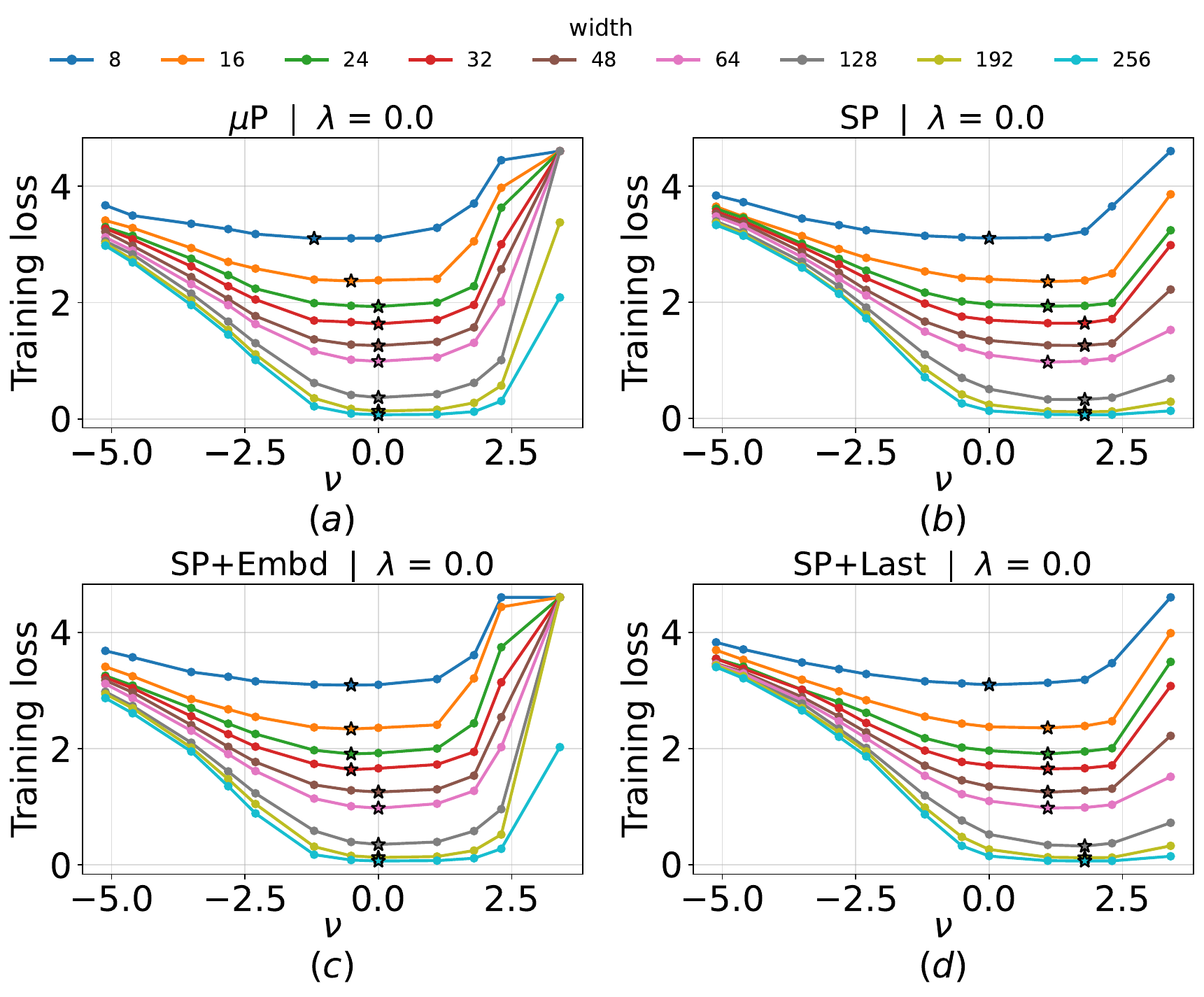}
    \caption{
    \looseness -1
    \emph{CNN experiments on CIFAR with Adam.}
    Training loss as a function of log learning rate $\nu$ for four parameterizations across widths. 
    In SP, the optimal learning rate drifts with width, while $\mu$P shows substantially less drift. 
    Increasing the learning rate of the input-facing layer in SP (SP+Embd) largely removes this drift and places the optimum in a similar region to $\mu$P. 
    By contrast, changing only the last-layer initialization (SP+Last) leaves both the learning-rate drift and the optimum location closer to SP. 
    This suggests that the role of the embedding layer learning rate in Transformers reflects a more general importance of the input-facing layer learning rate.
    }
    \label{fig:cnn_cifar_adam}
\end{figure}

In \Cref{section:embedding_layer_switch}, we demonstrated that the embedding layer learning rate is the primary factor in explaining the difference between SP and $\mu$P in Transformers trained with Adam. 
A natural question is whether this result is specific to Transformers with an embedding layer or generalizes to other architectures and tasks. 
To test this, we consider CNNs trained on CIFAR with Adam, where the first layer is a simple convolutional layer, and there is no LayerNorm or attention.

As there are only two differences in this setting (first and last layer), there are only four variants: $\mu$P, SP, SP+Embd, and SP+Last. 
\Cref{fig:cnn_cifar_adam} shows the training loss as a function of log learning rate $\nu$ across widths. 
We observe that the optimal learning rate increases with width in SP, whereas in $\mu$P it remains fairly constant. 
Training SP with a $\Theta(1)$ first-layer learning rate largely removes this drift: SP+Embd behaves similarly to $\mu$P both in the location of the optimal learning rate and in its reduced drift. 
By contrast, changing only the last-layer initialization has little effect: SP+Last remains closer to SP, both in the location of the optimum and in the observed learning-rate drift.

These results suggest that the role of the embedding layer learning rate is not specific to Transformers, and that training the input layer sufficiently fast is important for learning-rate transfer under Adam.

\section{Weight Decay scaling in Compute-Optimal Regime}
\label{appendix:wd_step_scaling_tpp}

\begin{figure}[!htb]
    \centering
    \includegraphics[width=0.75\linewidth]{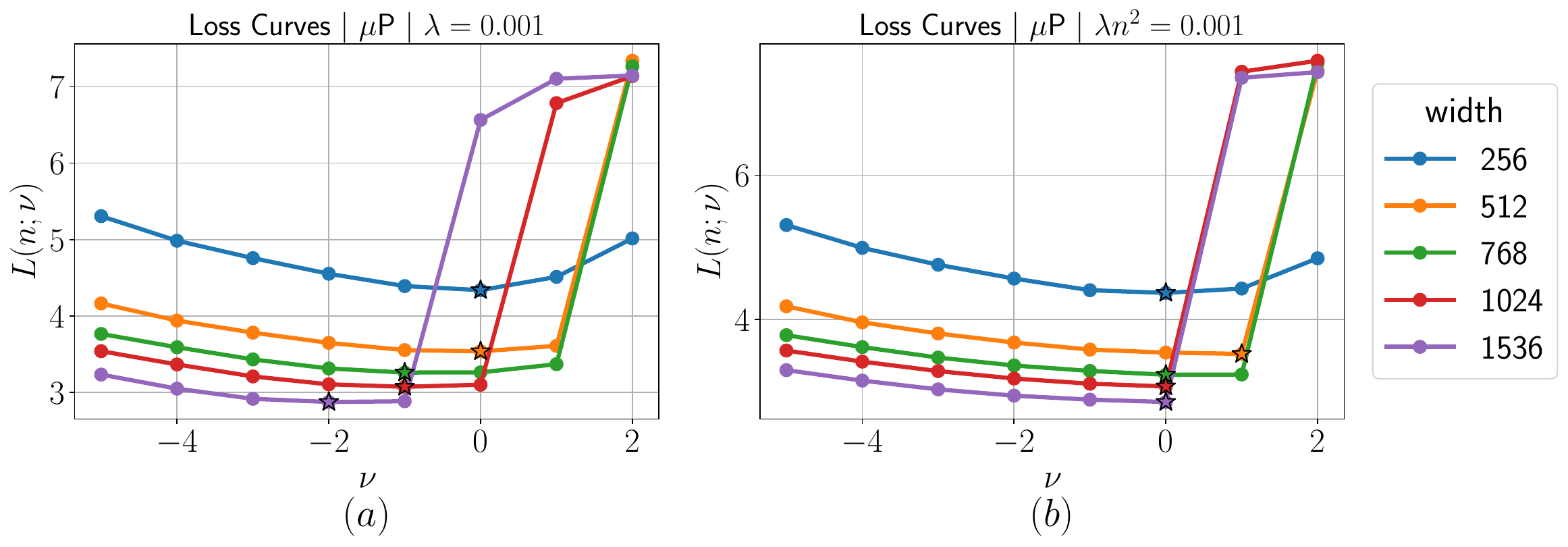}
    \caption{Loss curves for $\mu$P in the compute-optimal setting ($20$ TPP) under two weight decay scalings. (a) Standard $\mu$P convention $\eta \cdot \lambda = \Theta(1)$: $\nu^*(n)$ shifts to the left with increasing width. (b) Corrected convention $\eta \cdot \lambda = \Theta(\nicefrac{1}{n^2})$: $\nu^*(n)$ doesn't vary much, but the shape of the loss curves around the minimum changes across widths.}
    \label{fig:wd_scaling_tpp}
\end{figure}

\looseness -1
In the compute-optimal (fixed TPP) setting, the number of training steps scales as $T \propto n^2$, which violates $\mu$P's assumption of $\Theta(1)$ training steps compared to width. This makes the standard weight decay convention $\eta \cdot \lambda = \Theta(1)$ inadequate, as we observed in \Cref{section:wd_and_tpp} where transfer robustness $\kappa$ degrades with increasing weight decay. A natural scaling, motivated by \cite{bergsma2025power}, is to scale weight decay strength as $\lambda = \Theta(\nicefrac{1}{n^2})$ so that the scale of weight decay's cumulative contribution over $T = \Theta(n^2)$ steps $\eta \lambda T$ remains $\Theta(1)$ across widths.
That being said, we caution that this choice is not well motivated. There is no first-principle reason why the cumulative weight decay must be $\Theta(1)$, as different widths may benefit from different total contributions. 

\Cref{fig:wd_scaling_tpp} compares the loss curves for $\mu$P under two scaling conventions. Under $\eta \cdot \lambda = \Theta(1)$ (a), the optimal log learning rate $\nu^*(n)$ drifts noticeably to the left with increasing width, suggesting a slow convergence (small $\beta$). As a result, $\kappa = \alpha - 2\beta + \gamma$ becomes large, resulting in brittle transfer observed in \Cref{section:wd_and_tpp}. By comparison, under $\eta \cdot \lambda = \Theta(\nicefrac{1}{n^2})$ (panel b), this drift is visually reduced, suggesting an improvement in transfer robustness. However, the shape of the loss curves around the minimum changes across widths. We leave a systematic analysis of the appropriate weight decay scaling in the TPP regime to future work.

\section{The Effect of Learning Rate Warmup on Learning Rate Transfer}
\label{appendix:learning_rate_warmup}

\begin{figure*}[!htb]
    \centering
    \includegraphics[width=\linewidth]{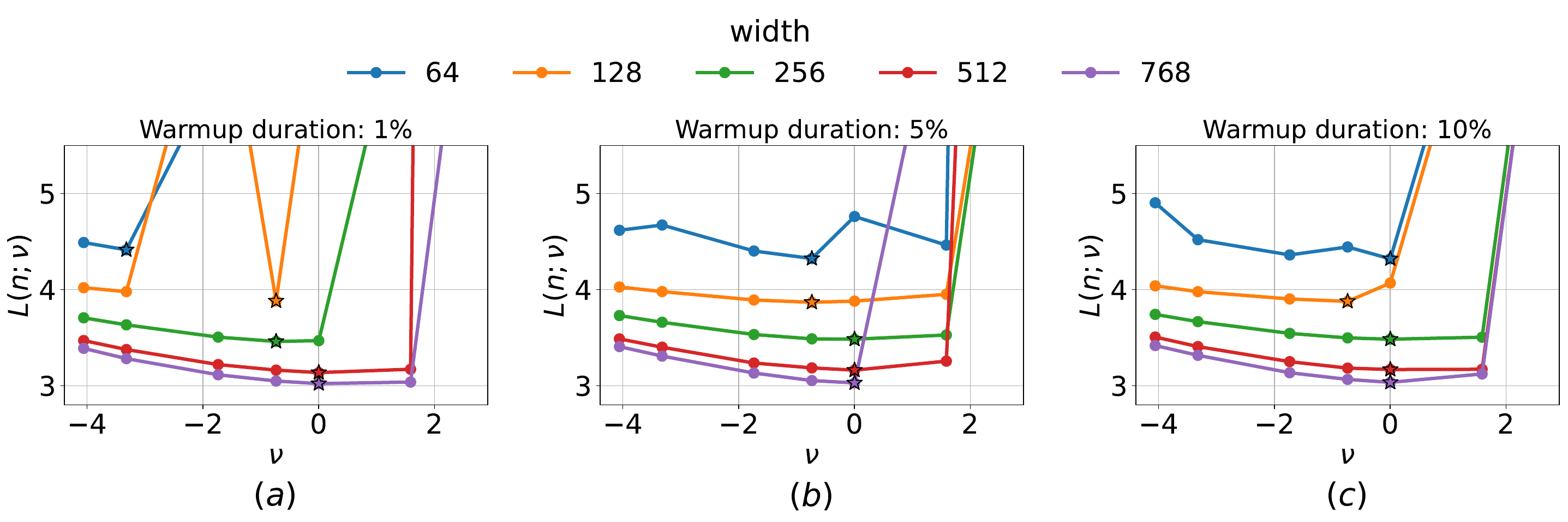}
    \caption{Learning rate warmup is essential for observing reliable learning rate transfer in $\mu$P.}
\label{fig:gpt_pretraining_transfer_warmup}
\end{figure*}

Learning rate warmup is standard practice in large-scale training~\citep{deepseekai2024,olmo20242}. 
Its primary effect is gradually reducing the sharpness of the loss Hessian (or pre-conditioned sharpness for Adam), effectively steering optimization towards well-conditioned regions of the loss landscape where the model can be trained at large learning rates~\citep{gilmer2022a,kalra2024why}.
Despite its widespread use, its interaction with learning rate transfer has not been systematically studied.
\Cref{fig:gpt_pretraining_transfer_warmup} shows that short warmup durations ($1$-$5\%$) result in training instabilities that make transfer unreliable\textemdash the loss curves are noisy across widths, making extrapolation meaningless.
At $10\%$ warmup, the loss curves become noticeably smoother and $\nu^*(n)$ aligns more consistently across widths.
This result confirms that warmup is crucial for observing reliable learning rate transfer, which is not accounted for in $\mu$P's theoretical derivation.
We use a conservative warmup duration of $20\%$ throughout experiments to observe reliable transfer.

\clearpage
\newpage

\section{Effect of Freezing the Embedding Layer}
\label{appendix:frozen_embd}

\begin{figure}[!htb]
    \centering
    \includegraphics[width=0.7\textwidth]{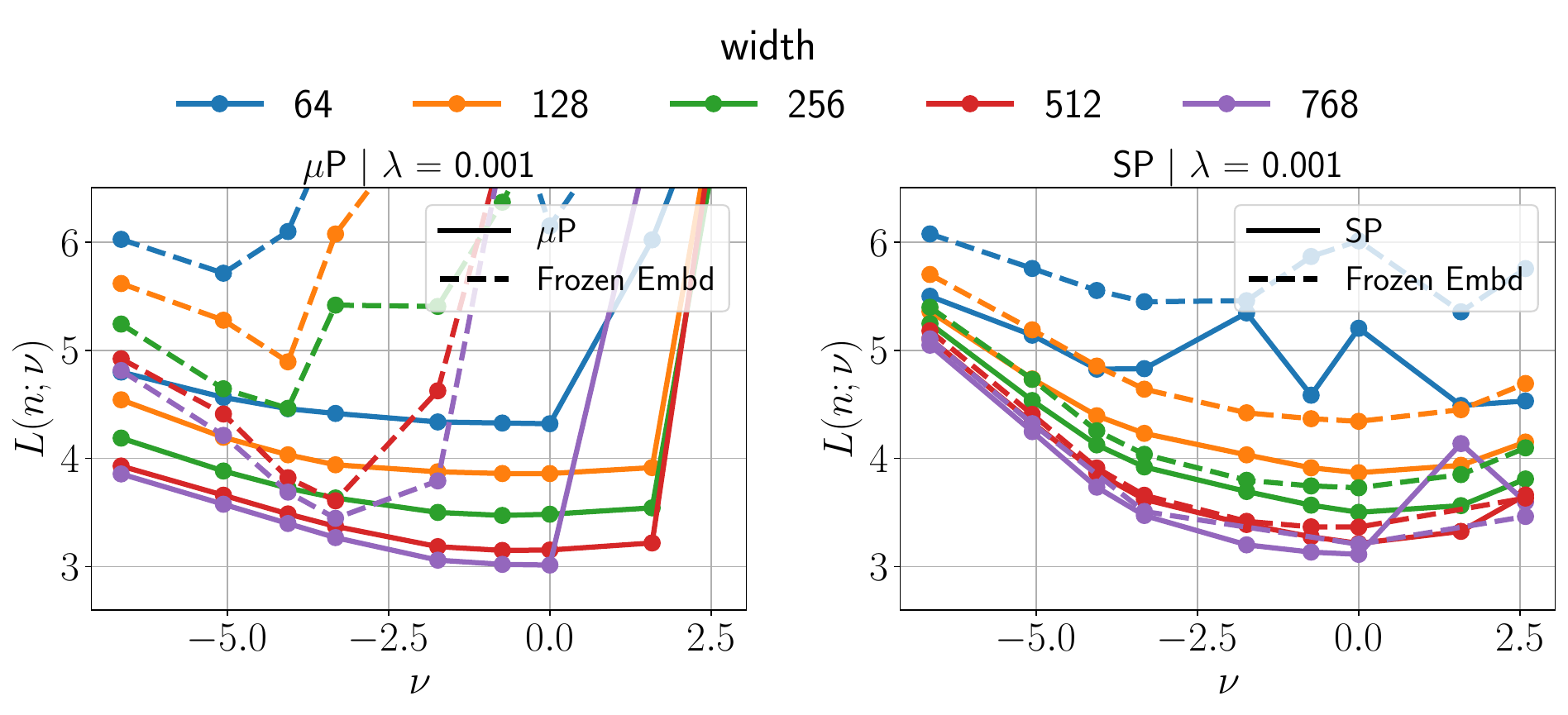}        
    \includegraphics[width=0.7\textwidth]{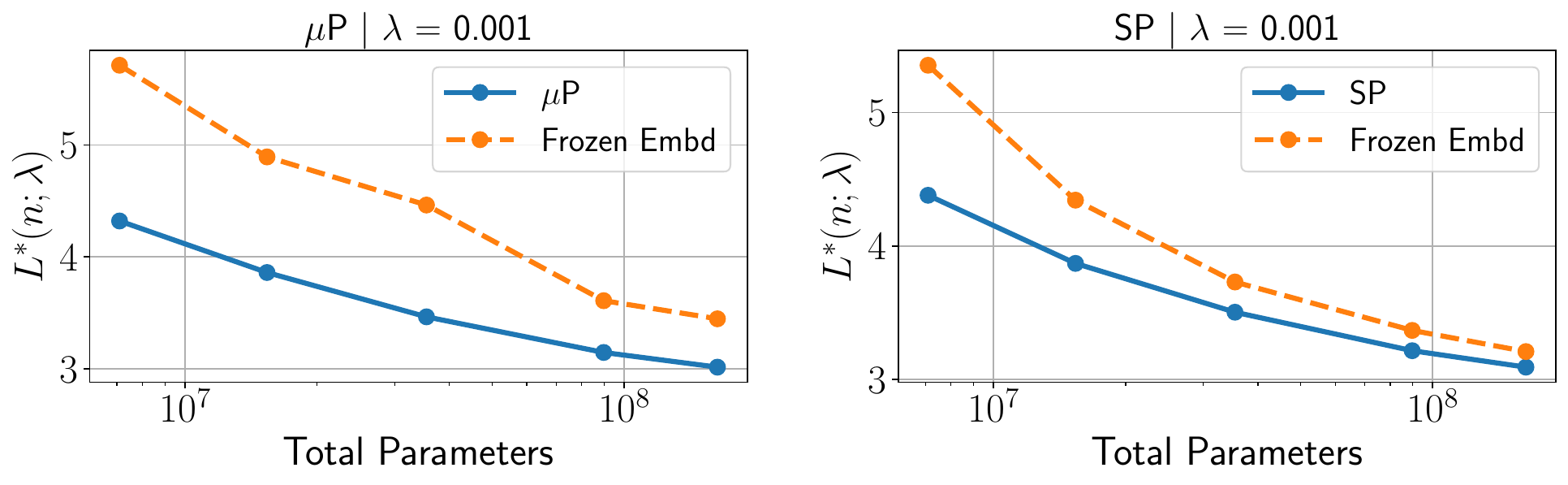}
    \includegraphics[width=0.7\textwidth]{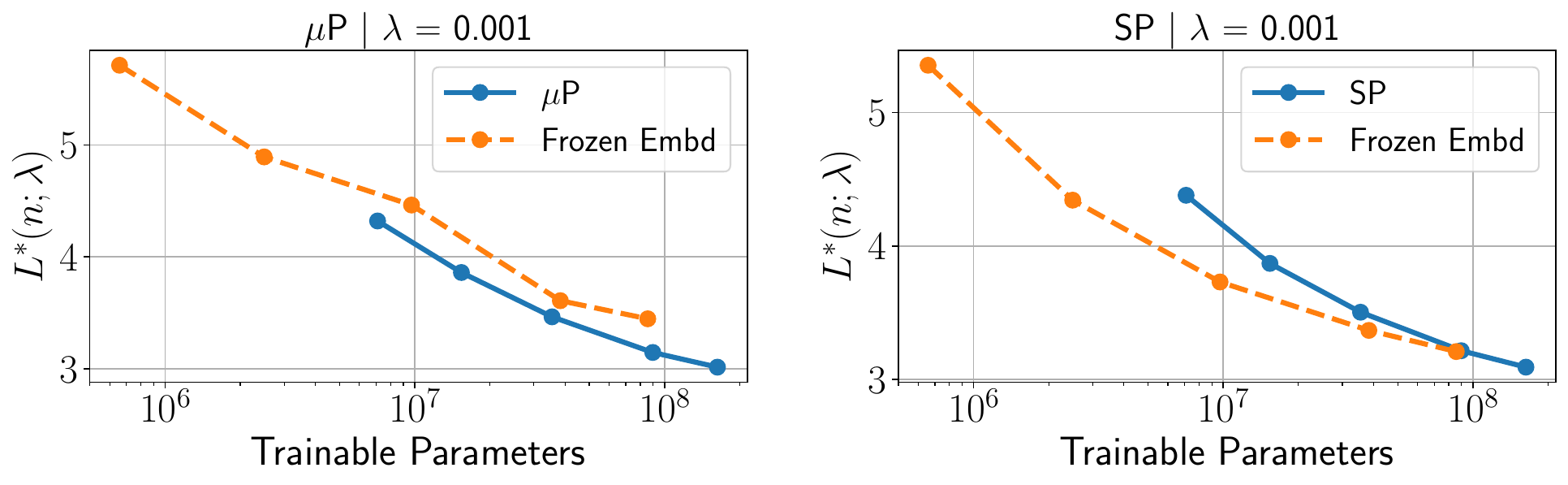}
        \caption{Effect of freezing the embedding layer in $\mu$P and SP. (top) Loss vs. log learning rate curves. (middle) Optimal loss vs. total parameters. (bottom) Optimal loss vs. trainable parameters, where trainable parameters exclude the embedding when frozen.}

    \label{fig:transfer_curves_frozen}
\end{figure}

The switch experiments in \Cref{section:embedding_layer_switch} show that training the embedding layer too slowly can both slow down learning and cause training instabilities. Here, we take this further by studying the effect of completely freezing the embedding layer in both $\mu$P and SP.

\Cref{fig:transfer_curves_frozen} (top) shows the loss vs. log learning rate curves. Freezing the embedding layer significantly affects $\mu$P, causing pronounced training instabilities and a shift in the optimal learning rate. SP is comparatively more robust: freezing causes instabilities at small widths, but the curves match the unfrozen case at large widths.

\Cref{fig:transfer_curves_frozen} (middle) shows the optimal loss scaling laws against total parameters. For $\mu$P, freezing the embedding causes a large performance gap that narrows with width. For SP, the effect is more modest, with frozen and unfrozen performance converging at large widths.

\looseness -1
However, comparing by total parameters is unfair, since at small widths the embedding dominates the parameter count (vocabulary size $\times$ width, with vocabulary size $50{,}304$). We therefore also compare against trainable parameters, which excludes frozen embedding parameters (\Cref{fig:transfer_curves_frozen}, bottom). Even after this correction, freezing $\mu$P's embedding remains worse across all widths, though the gap narrows considerably. For SP, the frozen variant achieves lower loss at small widths, suggesting that non-embedding parameters might be more parameter-efficient at small scales, though the unfrozen case catches up at large widths. We leave a detailed understanding of this phenomenon to future work.

\clearpage
\newpage

\section{Minimal Derivation of $\mu$P}
\label{appendix:mup_derivation}

In this section, we present a self-contained derivation of scaling conditions for neural networks trained at large widths and show that $\mu$P arises as one particular solution to these conditions.
We consider a minimal three-layer linear network.
The choice is minimal for two reasons. 
First, three layers suffice to capture all distinct layer types in a neural network: input-to-hidden, hidden-to-hidden, and hidden-to-output. 
Second, standard activation functions used in practice (ReLU, GeLU, SiLU, tanh) act element-wise, so they contribute only $\Theta_n(1)$ factors to both the forward and backward passes and leave the width-scaling exponents unchanged. 
We build upon the analysis and notation of~\citet{everett2024scaling}, and extend the derivation to cover weight decay, LayerNorm, and attention scaling. 
Our goal here is to provide a simple, transparent derivation of the scaling rules, making every assumption explicit at the step where it is used.

\subsection{Notation and Norms}
\label{appendix:notations}

In this derivation, we examine how every quantity (weights, activations, gradients, and updates) scales with the width $n$, using its RMS norm:
\begin{definition}[RMS norm]
\label{definition:rms_norm}
For a vector $\mathbf{v} \in \mathbb{R}^n$, the RMS norm is defined as:
\begin{align}
\|\mathbf{v}\|_{\mathrm{RMS}} := \sqrt{\frac{1}{n} \sum_{i=1}^n v_i^2}. \nonumber
\end{align}
For a matrix $M \in \mathbb{R}^{m \times n}$, we use the entrywise extension:
\begin{align}
\|M\|_{\mathrm{RMS}} := \sqrt{\frac{1}{mn} \sum_{i=1}^m \sum_{j=1}^n M_{ij}^2}. \nonumber
\end{align}
Unless we state otherwise, we use $\|\cdot\|$ to denote the RMS norm throughout this derivation.
\end{definition}

\begin{definition}[Width-scaling]
Let $\{\mathbf{v}_n\}$ be a family of vectors with $\mathbf{v}_n \in \mathbb{R}^n$. We say $\mathbf{v}_n = \Theta_n(n^\alpha)$ if there exist $C_1, C_2 > 0$ and $N$ such that $C_1 n^\alpha \leq \|\mathbf{v}_n\| \leq C_2 n^\alpha$ for all $n \geq N$. The one-sided versions $\mathcal{O}_n$ and $\Omega_n$ keep only the upper or lower bound. 
Furthermore, for two families $\{\mathbf{u}_n\}$ and $\{\mathbf{v}_n\}$, we write $\mathbf{v}_n \sim \mathbf{u}_n$ to indicate that they share the same width-scaling, i.e., $\|\mathbf{v}_n\| = \Theta_n(\|\mathbf{u}_n\|)$.
The subscript $n$ specifies width as the scaling variable, and we will drop it whenever it's clear from context.
\end{definition}

\subsection{Model Architecture}
\label{appendix:model_architecture}

Consider a three-layer linear network with trainable parameters $\boldsymbol{\theta} := \{U, W, V\}$, width $n$, input dimension $d_{\mathrm{in}}$, and output dimension $d_{\mathrm{out}}$. For a training example $(\mathbf{x}, \mathbf{y})$, the forward pass is:
\begin{align}
    h(\mathbf{x}) &= n^{-a_u} U \mathbf{x}, \nonumber \\
    z(\mathbf{x}) &= n^{-a_w} W h(\mathbf{x}), \nonumber \\
    f(\mathbf{x}) &= n^{-a_v} V z(\mathbf{x}),
    \label{equation:forward-pass-mlp}
\end{align}
with $U \in \mathbb{R}^{n \times d_{\mathrm{in}}}$, $W \in \mathbb{R}^{n \times n}$, and $V \in \mathbb{R}^{d_{\mathrm{out}} \times n}$. Here, the exponents $a:=(a_u, a_w, a_v)$ scale the forward pass in each layer. The initialization variances are controlled by the exponents $b:=(b_u, b_w, b_v)$:
\begin{align}
    U_{ij} \sim \mathcal{N}(0, n^{-2b_u}), \qquad
    W_{ij} \sim \mathcal{N}(0, n^{-2b_w}), \qquad
    V_{ij} \sim \mathcal{N}(0, n^{-2b_v}).
\end{align}
The per-layer learning rates are scaled by the exponents $c:=(c_u, c_w, c_v)$:
\begin{align}
    \eta_u = \eta \, n^{-c_u}, \qquad \eta_w = \eta \, n^{-c_w}, \qquad \eta_v = \eta \, n^{-c_v}.
\end{align}
where $\eta$ is the global learning rate. Finally, the per-layer weight decay strengths are scaled by the exponents $d:=(d_u, d_w, d_v)$:
\begin{align}
    \lambda_u = \lambda \, n^{-d_u}, \qquad \lambda_w = \lambda \, n^{-d_w}, \qquad \lambda_v = \lambda \, n^{-d_v},
\end{align}
where $\lambda$ is the weight decay strength. 
The exponents $\{a, b, c, d\}$ collectively define our $abcd$ parameterization, which extends the $abc$ parameterization of~\cite{yang21feature} with weight decay exponents $d$. 

We use the subscript $t$ to denote the training step. Thus, $U_t$, $W_t$, and $V_t$ denote the parameters after $t$ training steps, and $h_t(\mathbf{x})$, $z_t(\mathbf{x})$, and $f_t(\mathbf{x})$ denote the
corresponding activations and network output. Initialization corresponds to
$t=0$. For any quantity $q_t$, we write $\Delta q_t := q_{t} - q_{t-1}$ to denote its change at step $t$.

\subsection{Design Principles for Scaling Analysis}

For training to remain \emph{stable}, the activations and their updates must not vanish or explode as the width grows.
If the activation updates vanish, the corresponding layer remains frozen at initialization; on the other hand, if they explode, training diverges.
We formalize stability through the following two conditions~\cite{yang21feature}.

\begin{desideratum}[Stable Initialization]
\label{desideratum:stable-init}
The hidden activations and the network output remain stable with width at initialization:
\begin{align}
    \|h_0\| = \Theta_n(1), \qquad \|z_0\| = \Theta_n(1), \qquad \|f_0\| = \mathcal{O}_n(1).
\end{align}
\end{desideratum}

The hidden activations $h_0$ and $z_0$ are required to be $\Theta_n(1)$ for stable signal propagation across layers~\cite{Roberts_Yaida_Hanin_2022}. The network output $f_0$ in contrast has a looser $\mathcal{O}_n(1)$ constraint because initializing the last layer with small weights does not affect signal propagation through the network~\citep{yaida2022metaprincipledfamilyhyperparameterscaling}.

\begin{desideratum}[Stable Updates at the First step]
\label{desideratum:stable-updates}
At the first gradient step $t=1$, the hidden activation and network output updates remain stable with width:
\begin{align}
    \|\Delta h_1\| = \Theta_n(1), \qquad \|\Delta z_1\| = \Theta_n(1), \qquad \|\Delta f_1\| = \Theta_n(1).
\end{align}
\end{desideratum}
\looseness -1
Here, we impose this condition at $t=1$ so that the relevant computations are tractable. The same condition is desirable at every training step; however, extending it to general training step $t$ requires further assumptions on weights, activations, and their alignment, which we introduce in \Cref{appendix:multi_step_scaling}.

It is widely observed that parameterizations satisfying \Cref{desideratum:stable-init,desideratum:stable-updates} (e.g., $\mu$P) exhibit \emph{learning rate transfer}: optimal learning rates at small width remain optimal at large widths. 
The implication, however, is non-trivial: ensuring that activations and their updates remain stable does not by itself imply that the optimal hyperparameters are width-independent, particularly in the limit where number of training steps is much larger than width.
Several works support transfer under $\mu$P, though in restricted settings. 
The simplest setting is of~\citet{kalra2025universal}, who showed that, for a two-layer linear network trained on a single example, the dynamical equations under $\mu$P have no width dependence whatsoever, directly implying hyperparameter transfer without any assumptions. 
A recent work by~\citet{hayou2026a} provided a formal proof of learning-rate transfer under $\mu$P in restricted settings of linear networks with input and output layers fixed at initialization. 
At the first gradient step, they provide an explicit limit and convergence rate; by comparison, at any fixed step $t$, they establish transfer under the additional assumption that the loss has a unique minimizer in $\eta$, but do not provide convergence rates as in the first step case.
Beyond linear networks,~\citet{yaida2022metaprincipledfamilyhyperparameterscaling} showed that the leading-order terms in the function-space update equations are width-independent under $\mu$P, suggesting that the dynamics, and by extension the optimal hyperparameters, may also be width-stable in general neural networks.
The subleading term, however, can accumulate over long training horizons and contribute to the optimal learning rate.
Finally, ~\citet{noci2024super} empirically observed that the sharpness (largest Hessian eigenvalue) does not vary with width under $\mu$P, providing complementary evidence that the dominant loss landscape features remain invariant in $\mu$P.
Proving learning rate transfer in generic settings remains an open question.

\subsection{Conditions for Stable Initialization}

In this section, we derive the conditions on $\{a, b\}$ required to satisfy \Cref{desideratum:stable-init}; the conditions on exponents $\{c, d\}$ will be imposed by the training dynamics in the later sections. 
At initialization, the weights and their inputs are independent, so the scale of each layer's output can be easily computed by averaging over the weight distribution while treating the input as fixed. 
Since all the quantities in this section are at $t=0$, we suppress the time index for brevity.

\paragraph{\textbf{First layer}.} The forward pass of the first layer is:
\begin{align}
    h(\mathbf{x}) = n^{-a_u} U \mathbf{x}. \nonumber
\end{align}
Let $\langle \cdot \rangle$ denote expectation over the weight distribution. The expected squared norm of $h(\mathbf{x})$ is:
\begin{align}
    \left\langle \|h(\mathbf{x})\|^2 \right\rangle 
    = \left\langle \frac{1}{n} \sum_{i=1}^{n} h_i^2(\mathbf{x}) \right\rangle 
    &= \frac{1}{n} \sum_{i=1}^n n^{-2a_u} \sum_{j,k=1}^{d_{\mathrm{in}}} \langle U_{ij} U_{ik} \rangle\, x_j x_k \nonumber \\
    &= n^{-2a_u - 1} \sum_{i=1}^n \sum_{j=1}^{d_{\mathrm{in}}} n^{-2b_u}\, x_j^2  
    \qquad \left(\text{using } \langle U_{ij}U_{ik}\rangle = n^{-2b_u}\delta_{jk}\right) \nonumber \\
    &= n^{-2a_u - 2b_u} \sum_{j=1}^{d_{\mathrm{in}}} x_j^2.
\end{align}
To handle the input term, we assume:
\begin{assumption}[Input scaling]
\label{assumption:input}
The inputs are normalized so that $\|\mathbf{x}\| = \Theta_n(1)$.
\end{assumption}
Under \Cref{assumption:input}, $\sum_{j=1}^{d_{\mathrm{in}}} x_j^2 = \Theta_n(1)$, so
\begin{align}
 \left\langle \|h(\mathbf{x})\|^2 \right\rangle = \Theta_n(n^{-2a_u - 2b_u}).
\end{align}
Requiring $\|h(\mathbf{x})\| = \Theta_n(1)$ yields:
\begin{align}
    \boxed{a_u + b_u = 0.}
\end{align}

\paragraph{\textbf{Middle layer}.} The forward pass of the middle layer is:
\begin{align}
    z(\mathbf{x}) = n^{-a_w} W h(\mathbf{x}). \nonumber
\end{align}
The expected squared norm of $z(\mathbf{x})$, with the expectation taken over $W$ at fixed $h(\mathbf{x})$, is:
\begin{align}
    \left\langle \|z(\mathbf{x})\|^2 \right\rangle 
    = \left\langle \frac{1}{n} \sum_{i=1}^{n} z_i^2(\mathbf{x}) \right\rangle 
    &= \frac{1}{n} \sum_{i=1}^n n^{-2a_w} \sum_{j,k=1}^n \langle W_{ij} W_{ik}\rangle\, h_j(\mathbf{x}) h_k(\mathbf{x}) \nonumber \\
    &= n^{-2a_w - 1} \sum_{i=1}^n \sum_{j=1}^n n^{-2b_w}\, h_j^2(\mathbf{x}) 
    \qquad \left(\text{using } \langle W_{ij}W_{ik}\rangle = n^{-2b_w}\delta_{jk}\right) \nonumber \\
    &= n^{-2a_w - 2b_w + 1}\, \|h(\mathbf{x})\|^2.
\end{align}
Using $\|h(\mathbf{x})\| = \Theta_n(1)$ from the first layer, we have:
\begin{align}
    \boxed{a_w + b_w = \tfrac{1}{2}.}
    \label{equation:stable-init-cond2}
\end{align}

\paragraph{\textbf{Last layer}.} The forward pass of the last layer is:
\begin{align}
    f(\mathbf{x}) = n^{-a_v} V z(\mathbf{x}). \nonumber
\end{align}
The calculation follows as in the middle layer, with $V$ in place of $W$ and the output dimension $d_{\mathrm{out}}$ replacing $n$ in the output index. 
\begin{align}
    \left\langle \|f(\mathbf{x})\|^2 \right\rangle = n^{-2a_v - 2b_v + 1}\, \|z(\mathbf{x})\|^2.
\end{align}
Using $\|z(\mathbf{x})\| = \Theta_n(1)$ and the one-sided requirement $\|f(\mathbf{x})\| = \mathcal{O}_n(1)$:
\begin{align}
    \boxed{a_v + b_v \geq \tfrac{1}{2}.}
\end{align}
Unlike the previous two layers, the last-layer initialization scale is not fixed by stability alone, as any $a_v + b_v \geq 1/2$ satisfies the constraint.

The conditions above fix the weight and activation scales only at initialization, and the weights and activations can evolve in a width-dependent manner during training. We will discuss the implications in the later sections.

\subsection{Gradient Scaling at Initialization}
\label{appendix:gradient_scaling}

Given a dataset $\mathcal{D} = \{(\mathbf{x}^\mu, \mathbf{y}^\mu)\}_{\mu=1}^D$ and a per-example loss function $\ell: \mathbb{R}^{d_{\mathrm{out}}} \times \mathbb{R}^{d_{\mathrm{out}}} \to \mathbb{R}$, we define the training loss as the empirical average:
\begin{align}
    L(\boldsymbol{\theta}) = \frac{1}{D} \sum_{\mu=1}^D \ell\!\left(f(\mathbf{x}^\mu), \mathbf{y}^\mu\right).
\end{align}
The per-layer gradients are:
\begin{align}
    g_v:= \nabla_v L(\boldsymbol{\theta}) &= \frac{1}{D} \sum_{\mu=1}^D n^{-a_v} \nabla_f \ell^\mu \cdot z(\mathbf{x}^\mu)^\top, \nonumber \\
    g_w := \nabla_w L(\boldsymbol{\theta})&= \frac{1}{D} \sum_{\mu=1}^D n^{-a_w - a_v} V^\top \nabla_f \ell^\mu \cdot h(\mathbf{x}^\mu)^\top, \nonumber \\
    g_u := \nabla_u L(\boldsymbol{\theta})&= \frac{1}{D} \sum_{\mu=1}^D n^{-a_u - a_w - a_v} W^\top V^\top \nabla_f \ell^\mu \cdot \mathbf{x}^{\mu\top},
\end{align}
where $\nabla_f \ell^\mu := \nabla_f \ell(f(\mathbf{x}^\mu), \mathbf{y}^\mu)$. Next, we assume that the loss derivative does not scale with width:
\begin{assumption}[Loss derivative scaling]
\label{assumption:loss_derivative}
For every example $(\mathbf{x}^\mu, \mathbf{y}^\mu) \in \mathcal{D}$, the per-example loss derivative satisfies $\| \nabla_f \ell^\mu \| = \Theta_n(1)$.
\end{assumption}

We first analyze the per-example terms. For $g_v$, the per-example term $\nabla_f \ell^\mu \cdot z(\mathbf{x}^\mu)^\top$ is an outer product of two vectors with $\Theta_n(1)$ entries, so: 
\begin{align}
    \|n^{-a_v} \nabla_f \ell^\mu z(\mathbf{x}^\mu)^\top\| = n^{-a_v} \| \nabla_f \ell^\mu \| \| z(\mathbf{x}^\mu) \| = \Theta_n(n^{-a_v}).
\end{align}

For $g_w$, the per-example term is the outer product $(V^\top \nabla_f \ell^\mu) h(\mathbf{x}^\mu)^\top$. The left factor is a matrix-vector product along the output dimension $d_{\mathrm{out}}$ and does not pick up a width scaling factor. Therefore, we can write:
\begin{align}
    \langle \|V^\top \nabla_f \ell^\mu\|^2 \rangle \sim n^{-2b_v}\, \|\nabla_f \ell^\mu\|^2 = \Theta_n(n^{-2b_v}), \nonumber
\end{align}
This gives $\|V^\top \nabla_f \ell^\mu\| = \Theta_n(n^{-b_v})$. The right factor satisfies $\|h(\mathbf{x}^\mu)\| = \Theta_n(1)$, so:
\begin{align}
    \|n^{-a_w - a_v} V^\top \nabla_f \ell^\mu  h(\mathbf{x}^\mu)^\top\| = n^{-a_w - a_v} \|V^\top \nabla_f \ell^\mu\| \|h(\mathbf{x}^\mu)\| = \Theta_n(n^{-a_w - a_v - b_v}).
\end{align}

For $g_u$, the per-example term is the outer product $(W^\top V^\top \nabla_f \ell^\mu) \mathbf{x}^{\mu\top}$. The left factor is a matrix--vector product, whose expected squared RMS norm at initialization is:
\begin{align}
    \langle \|W^\top V^\top \nabla_f \ell^\mu\|^2 \rangle \sim n^{-2b_w + 1}\, \|V^\top \nabla_f \ell^\mu\|^2 = \Theta_n(n^{-2 b_w - 2b_v + 1}), \nonumber
\end{align}
Thus, $\|W^\top V^\top \nabla_f \ell^\mu\| = \Theta_n(n^{-b_w - b_v + \frac{1}{2}})$. Using $a_w + b_w = \frac{1}{2}$, this scaling becomes $\Theta_n(n^{a_w - b_v})$.
The right factor satisfies $\|\mathbf{x}^\mu\| = \Theta_n(1)$, so:
\begin{align}
    \|n^{-a_u - a_w - a_v} W^\top V^\top \nabla_f \ell^\mu  \mathbf{x}^{\mu\top}\| = n^{-a_u - a_w - a_v} \|W^\top V^\top \nabla_f \ell^\mu\| \|\mathbf{x}^\mu\| = \Theta_n(n^{-a_u - a_v - b_v}).
\end{align}

In summary, the per-example contributions scale as:
\begin{align}
    \|g_v^\mu\| = \Theta_n(n^{-a_v}), \qquad
    \|g_w^\mu\| = \Theta_n(n^{-a_w - a_v - b_v}), \qquad
    \|g_u^\mu\| = \Theta_n(n^{-a_u - a_v - b_v}).
\end{align}
The full gradient $g_l$ is obtained by averaging the per-example contributions:
\begin{align}
    g_l = \frac{1}{D} \sum_{\mu=1}^D g_l^\mu.
\end{align}
We make the following assumption on this aggregation:
\begin{assumption}[Aggregation does not induce width-scaling]
\label{assumption:aggregation}
The average $\frac{1}{D} \sum_{\mu=1}^D g_l^\mu$ inherits the width-scaling of the per-example terms.
\end{assumption}
Under \Cref{assumption:aggregation}, the layerwise gradients scale as:
\begin{align}
    \|g_v\| = \Theta_n(n^{-a_v}), \qquad
    \|g_w\| = \Theta_n(n^{-a_w - a_v - b_v}), \qquad
    \|g_u\| = \Theta_n(n^{-a_u - a_v - b_v}).
\end{align}
The assumption rules out the gradient acquiring extra width scaling from aggregation. This is a benign assumption, as we do not expect a width dependence from aggregating. Without it, the triangle inequality provides only $\mathcal{O}_n$ upper bounds.

Similar to the activations, the gradient scales can also evolve in a width-dependent manner during training. We will discuss the implications in the later sections.

\subsection{Alignment and Feature Learning Exponents}

Following~\cite{everett2024scaling}, we define three alignment exponents $\{\rho, \omega, \sigma\} \in [0, 1]$ per layer that quantify the correlations between weight matrices, their updates, and activation vectors at the first gradient step.

\begin{definition}[Alignment exponents]
\label{definition:alignment}
For each layer, we define:
\begin{align}
    \rho_{v,1} &:= \lim_{n \to \infty} \log_n \frac{\|\Delta V_1\, z_0\|}{\|\Delta V_1\| \|z_0\|}, &
    \rho_{w,1} &:= \lim_{n \to \infty} \log_n \frac{\|\Delta W_1\, h_0\|}{\|\Delta W_1\| \|h_0\|}, \nonumber \\
    \omega_{v,1} &:= \lim_{n \to \infty} \log_n \frac{\|V_0\, \Delta z_1\|}{\|V_0\| \|\Delta z_1\|}, &
    \omega_{w,1} &:= \lim_{n \to \infty} \log_n \frac{\|W_0\, \Delta h_1\|}{\|W_0\| \|\Delta h_1\|}, \nonumber \\
    \sigma_{v,1} &:= \lim_{n \to \infty} \log_n \frac{\|\Delta V_1\, \Delta z_1\|}{\|\Delta V_1\| \|\Delta z_1\|}, &
    \sigma_{w,1} &:= \lim_{n \to \infty} \log_n \frac{\|\Delta W_1\, \Delta h_1\|}{\|\Delta W_1\| \|\Delta h_1\|}.
\end{align}
We do not define alignment exponents for the first layer because the matrix-vector product $U\mathbf{x}$ does not involve the width, and thus no alignment exponent is needed.
\end{definition}

\begin{definition}[Activation Update exponents]
\label{definition:feature_learning}
Let $\Delta h_1$, $\Delta z_1$, and $\Delta f_1$ denote the activation updates at the first gradient step. We define:
\begin{align}
    r_{u,1} := \lim_{n \to \infty} \log_n \|\Delta h_1\|, \quad
    r_{w,1} := \lim_{n \to \infty} \log_n \|\Delta z_1\|, \quad
    r_{v,1} := \lim_{n \to \infty} \log_n \|\Delta f_1\|.
\end{align}
\end{definition}

These definitions generalize to any step $t$ by replacing the subscript $1$ with $t$ throughout. We work at $t=1$ in the following sections because the scaling of weights, updates, and activations is tractable at the first step.

\subsection{Stable Update Conditions for SGD}
We now derive the conditions on $\{a, b, c\}$ that satisfy \Cref{desideratum:stable-updates} for SGD, assuming \Cref{desideratum:stable-init} holds.
We analyze the weight decay scaling in \Cref{appendix:weight_decay_scaling} for SGD and Adam together, as it follows the same logic for both optimizers.

\paragraph{SGD update.} For SGD with per-layer learning rate $\eta_l = \eta\, n^{-c_l}$, the parameter update at step $t$ is:
\begin{align}
    U_{t+1} = U_t - \eta_u\, g_u(\boldsymbol{\theta}_t), \qquad
    W_{t+1} = W_t - \eta_w\, g_w(\boldsymbol{\theta}_t), \qquad
    V_{t+1} = V_t - \eta_v\, g_v(\boldsymbol{\theta}_t).
\end{align}
At the first step, $\|\Delta U_1\| = \eta_u \|g_u\|$, $\|\Delta W_1\| = \eta_w \|g_w\|$, and $\|\Delta V_1\| = \eta_v \|g_v\|$. Using the gradient scalings at initialization from \Cref{appendix:gradient_scaling}:
\begin{align}
    \|\Delta U_1\| = \Theta_n(n^{-c_u - a_u - a_v - b_v}), \quad \|\Delta W_1\| = \Theta_n(n^{-c_w - a_w - a_v - b_v}), \quad \|\Delta V_1\| = \Theta_n(n^{-c_v - a_v}).
\end{align}
\Cref{desideratum:stable-updates} requires $\|\Delta h_1\|, \|\Delta z_1\|, \|\Delta f_1\| = \Theta_n(1)$, which we enforce by setting the activation update exponents to zero: $r_{u,1} = r_{w,1} = r_{v,1} = 0$.

\paragraph{\textbf{First layer}.} The activation update is $\Delta h_1(\mathbf{x}) = n^{-a_u} \Delta U_1\, \mathbf{x}$. Since the matrix-vector product $U \mathbf{x}$ does not scale with width, we can write $\|\Delta U_1 \mathbf{x}\| = C \|\Delta U_1\| \|\mathbf{x}\|$ for some width-independent constant $C$. As a result, the activation update scales as:
\begin{align}
    \|\Delta h_1(\mathbf{x})\| = C n^{-a_u} \|\Delta U_1\| \|\mathbf{x}\| = \Theta_n(n^{-2a_u - c_u - a_v - b_v}).
\end{align}
Setting $r_{u,1} = 0$:
\begin{align}
    \boxed{2a_u + c_u + a_v + b_v = 0.}
    \label{eq:sgd-cond-u}
\end{align}

\paragraph{\textbf{Middle layer}.} The activation $z(\mathbf{x})$ at step $t=1$ is:
\begin{align}
    z_{1}(\mathbf{x}) = n^{-a_w} (W_0 + \Delta W_1)(h_0(\mathbf{x}) + \Delta h_1(\mathbf{x})).
\end{align}
After expanding and subtracting $z_0(\mathbf{x}) = n^{-a_w} W_0 h_0(\mathbf{x})$, the activation update decomposes into three terms:
\begin{align}
    \Delta z_1(\mathbf{x}) 
    &= \underbrace{n^{-a_w} \Delta W_1 \, h_0(\mathbf{x})}_{\text{weight-update term}} 
    \;+\; \underbrace{n^{-a_w} W_0 \, \Delta h_1(\mathbf{x})}_{\text{activation-update term}} 
    \;+\; \underbrace{n^{-a_w} \Delta W_1 \, \Delta h_1(\mathbf{x})}_{\text{second-order term}}.
\end{align}
By the triangle inequality:
\begin{align}
    \|\Delta z_1(\mathbf{x})\| \;\leq\; n^{-a_w}\!\left( \|\Delta W_1 h_0(\mathbf{x})\| + \|W_0 \Delta h_1(\mathbf{x})\| + \|\Delta W_1 \Delta h_1(\mathbf{x})\| \right).
\end{align}
We make the following assumption on the superposition of the three terms:
\begin{assumption}[Superposition does not induce width-scaling]
\label{assumption:superposition}
The superposition of the three terms in $\Delta z_1(\mathbf{x})$ does not introduce additional width-scaling: if each term is $\Theta_n(n^\alpha)$, then their sum is $\Theta_n(n^\alpha)$.
\end{assumption}
Under \Cref{assumption:superposition}, it suffices to analyze the width-scaling of each term individually. Throughout this derivation, we impose the condition $r_{u,1} = 0$, i.e., $\|\Delta h_1\| = \Theta_n(1)$.

\emph{Weight-update term.} By the definition of the alignment exponent (\Cref{definition:alignment}), $\|\Delta W_1 \, h_0\| = n^{\rho_{w,1}}\, \|\Delta W_1\|\, \|h_0\|$. Using $\|\Delta W_1\| = \Theta_n(n^{-c_w - a_w - a_v - b_v})$ from the SGD update and $\|h_0\| = \Theta_n(1)$:
\begin{align}
    n^{-a_w} \|\Delta W_1 \, h_0(\mathbf{x})\| = \Theta_n(n^{\rho_{w,1} - 2a_w - c_w - a_v - b_v}).
\end{align}
Requiring this to be $\Theta_n(1)$:
\begin{align}
    \boxed{\rho_{w,1} - 2a_w - c_w - a_v - b_v = 0.}
    \label{eq:sgd-cond-w-rho}
\end{align}

\emph{Activation-update term.} Similarly, $\|W_0 \, \Delta h_1\| = n^{\omega_{w,1}}\, \|W_0\|\, \|\Delta h_1\|$. Using $\|W_0\| = \Theta_n(n^{-b_w})$ at initialization, $\|\Delta h_1\| = \Theta_n(1)$, and $a_w + b_w = 1/2$:
\begin{align}
    n^{-a_w} \|W_0 \, \Delta h_1(\mathbf{x})\| = \Theta_n(n^{\omega_{w,1} - a_w - b_w}) = \Theta_n(n^{\omega_{w,1} - 1/2}).
\end{align}
Requiring this to be $\Theta_n(1)$ fixes the alignment exponent itself:
\begin{align}
    \boxed{\omega_{w,1} - \tfrac{1}{2} = 0.}
    \label{eq:sgd-cond-w-omega}
\end{align}
Unlike $\rho_{w,1}$ (and $\sigma_{w,1}$, as we will see below), $\omega_{w,1}$ is not a free variable if we demand stability and non-vanishing updates $r_{w, 1} = 0$ and its value is pinned to $1/2$. This scaling $\omega = 1/2$ is expected to hold at the first gradient step, but as training progresses, $\omega_{w,t}$ is free to evolve, and the activation-update term contribution may vanish or diverge as width is scaled.

\emph{Second-order term.} By the definition of the alignment exponent (\Cref{definition:alignment}), $\|\Delta W_1 \, \Delta h_1\| = n^{\sigma_{w,1}}\, \|\Delta W_1\|\, \|\Delta h_1\|$. Using $\|\Delta W_1\| = \Theta_n(n^{-c_w - a_w - a_v - b_v})$ and $\|\Delta h_1\| = \Theta_n(1)$:
\begin{align}
    n^{-a_w} \|\Delta W_1 \, \Delta h_1(\mathbf{x})\| = \Theta_n(n^{\sigma_{w,1} - 2a_w - c_w - a_v - b_v}).
\end{align}
Requiring this to be $\Theta_n(1)$:
\begin{align}
    \boxed{\sigma_{w,1} - 2a_w - c_w - a_v - b_v = 0.}
    \label{eq:sgd-cond-w-sigma}
\end{align}

The boxed conditions \Cref{eq:sgd-cond-w-rho,eq:sgd-cond-w-omega,eq:sgd-cond-w-sigma} correspond to requiring \emph{all three} terms in $\Delta z_1$ to be $\Theta_n(1)$. The minimal requirement for $\|\Delta z_1\| = \Theta_n(1)$ is weaker: each term must be $\mathcal{O}_n(1)$ and at least one must be $\Theta_n(1)$, i.e.,
\begin{align}
    \boxed{\max\!\big(\rho_{w,1} - 2a_w - c_w - a_v - b_v,\;\; \omega_{w,1} - \tfrac{1}{2},\;\; \sigma_{w,1} - 2a_w - c_w - a_v - b_v\big) = 0.}
    \label{eq:sgd-cond-w-max}
\end{align}

\paragraph{\textbf{Last layer}.} The output update decomposes similarly:
\begin{align}
    \Delta f_1(\mathbf{x}) 
    = \underbrace{n^{-a_v} \Delta V_1 \, z_0(\mathbf{x})}_{\text{weight-update term}} 
    + \underbrace{n^{-a_v} V_0 \, \Delta z_1(\mathbf{x})}_{\text{activation-update term}} 
    + \underbrace{n^{-a_v} \Delta V_1 \, \Delta z_1(\mathbf{x})}_{\text{second-order term}}.
\end{align}
Under \Cref{assumption:superposition}, it suffices to analyze each term individually. We impose $r_{w,1} = 0$, i.e., $\|\Delta z_1\| = \Theta_n(1)$.

\emph{Weight-update term.} By the definition of the alignment exponent (\Cref{definition:alignment}), $\|\Delta V_1 \, z_0\| = n^{\rho_{v,1}}\, \|\Delta V_1\|\, \|z_0\|$. Using $\|\Delta V_1\| = \Theta_n(n^{-c_v - a_v})$ from the SGD update and $\|z_0\| = \Theta_n(1)$:
\begin{align}
    n^{-a_v} \|\Delta V_1 \, z_0(\mathbf{x})\| = \Theta_n(n^{\rho_{v,1} - 2a_v - c_v}).
\end{align}
Requiring this to be $\Theta_n(1)$:
\begin{align}
    \boxed{\rho_{v,1} - 2a_v - c_v = 0.}
    \label{eq:sgd-cond-v-rho}
\end{align}

\emph{Activation-update term.} Similarly, $\|V_0 \, \Delta z_1\| = n^{\omega_{v,1}}\, \|V_0\|\, \|\Delta z_1\|$. Using $\|V_0\| = \Theta_n(n^{-b_v})$ at initialization and $\|\Delta z_1\| = \Theta_n(1)$:
\begin{align}
    n^{-a_v} \|V_0 \, \Delta z_1(\mathbf{x})\| = \Theta_n(n^{\omega_{v,1} - a_v - b_v}).
\end{align}
Requiring this to be $\Theta_n(1)$:
\begin{align}
    \boxed{\omega_{v,1} - a_v - b_v = 0.}
    \label{eq:sgd-cond-v-omega}
\end{align}
Unlike the middle layer, demanding $r_{v, 1} = 0$ does not pin $\omega_{v,1}$ to a fixed value, and it is constrained only through the relation $\omega_{v,1} = a_v + b_v$, where $a_v + b_v$ is itself underdetermined due to the weaker initialization constraint (\Cref{desideratum:stable-init}).

\emph{Second-order term.} By the definition of the alignment exponent (\Cref{definition:alignment}), $\|\Delta V_1 \, \Delta z_1\| = n^{\sigma_{v,1}}\, \|\Delta V_1\|\, \|\Delta z_1\|$. Using the same scalings:
\begin{align}
    n^{-a_v} \|\Delta V_1 \, \Delta z_1(\mathbf{x})\| = \Theta_n(n^{\sigma_{v,1} - 2a_v - c_v}).
\end{align}
Requiring this to be $\Theta_n(1)$:
\begin{align}
    \boxed{\sigma_{v,1} - 2a_v - c_v = 0.}
    \label{eq:sgd-cond-v-sigma}
\end{align}

As in the middle layer, the boxed conditions \Cref{eq:sgd-cond-v-rho,eq:sgd-cond-v-omega,eq:sgd-cond-v-sigma} correspond to requiring all three terms in $\Delta f_1$ to be $\Theta_n(1)$. The minimal requirement for $\|\Delta f_1\| = \Theta_n(1)$ is weaker:
\begin{align}
    \boxed{\max\!\big(\rho_{v,1} - 2a_v - c_v,\;\; \omega_{v,1} - a_v - b_v,\;\; \sigma_{v,1} - 2a_v - c_v\big) = 0.}
    \label{eq:sgd-cond-v-max}
\end{align}

\paragraph{Gauge symmetry for SGD.} The boxed conditions \Cref{eq:sgd-cond-w-max,eq:sgd-cond-v-max} are invariant under the per-layer transformation:
\begin{align}
    a_l \to a_l + \Delta_l, \qquad b_l \to b_l - \Delta_l, \qquad c_l \to c_l - 2\Delta_l, \nonumber
\end{align}
for any $\Delta_l \in \mathbb{R}$ and each layer $l \in \{u, w, v\}$. 
The forward-pass exponent $a_l + b_l$ is unchanged, and the shift in $c_l$ compensates so that update equations have the same width scalings. As a consequence, different choices of $(a, b, c)$ describe the same training dynamics. We will later use this freedom later to write down several equivalent forms of $\mu$P (\Cref{tab:mup_implementations}).

\subsection{Stable Update Conditions for Adam}
\label{appendix:stable-update-adam}

We now derive the conditions on $\{a, b, c\}$ that satisfy \Cref{desideratum:stable-updates} for Adam, assuming \Cref{desideratum:stable-init} holds. The only change from SGD is in the scaling of the parameter updates; the activation decompositions and alignment-exponent calculations are otherwise identical. Throughout this section, we use \Cref{assumption:superposition} on the superposition of three-term decomposition, and apply the alignment-exponent identities from \Cref{definition:alignment} (e.g., $\|\Delta W h\| = n^{\rho_w}\|\Delta W\|\|h\|$) without further citation. Weight decay is analyzed in \Cref{appendix:weight_decay_scaling} for SGD and Adam together.

\paragraph{Adam update.} Adam computes the update as $\Delta \theta_l = -\eta_l \cdot m_l / (\sqrt{v_l} + \epsilon)$, where $m_l$ and $v_l$ are the first and second moment estimates of $g_l$. We assume:
\begin{assumption}[Adam normalized update]
\label{assumption:adam-update}
The normalized update $m_l / (\sqrt{v_l} + \epsilon)$ has $\Theta_n(1)$ entries. This is a reasonable assumption as $m_l$ scales as $g_l$ and $v_l$ scales as $g_l^2$ entrywise, so $m_l/\sqrt{v_l}$ behaves like a sign-magnitude normalization independent of the gradient width scaling. 
\end{assumption}
Under \Cref{assumption:adam-update}, the parameter update norms at the first step depend only on the layerwise learning rate:
\begin{align}
    \|\Delta U_1\| = \Theta_n(n^{-c_u}), \qquad
    \|\Delta W_1\| = \Theta_n(n^{-c_w}), \qquad
    \|\Delta V_1\| = \Theta_n(n^{-c_v}).
\end{align}
\Cref{desideratum:stable-updates} requires $\|\Delta h_1\|, \|\Delta z_1\|, \|\Delta f_1\| = \Theta_n(1)$, which we enforce by setting the activation update exponents to zero: $r_{u,1} = r_{w,1} = r_{v,1} = 0$.

\paragraph{\textbf{First layer}.} The activation update is $\Delta h_1(\mathbf{x}) = n^{-a_u} \Delta U_1\, \mathbf{x}$. Since the matrix-vector product $U \mathbf{x}$ does not scale with width, we can write $\|\Delta U_1 \mathbf{x}\| = C \|\Delta U_1\| \|\mathbf{x}\|$ for some width-independent constant $C$. As a result:
\begin{align}
    \|\Delta h_1(\mathbf{x})\| = C n^{-a_u} \|\Delta U_1\| \|\mathbf{x}\| = \Theta_n(n^{-a_u - c_u}).
\end{align}
Setting $r_{u,1} = 0$:
\begin{align}
    \boxed{a_u + c_u = 0.}
    \label{eq:adam-cond-u}
\end{align}

\paragraph{\textbf{Middle layer}.} As in the SGD case, the activation update decomposes into three terms:
\begin{align}
    \Delta z_1(\mathbf{x}) 
    &= \underbrace{n^{-a_w} \Delta W_1 \, h_0(\mathbf{x})}_{\text{weight-update term}} 
    \;+\; \underbrace{n^{-a_w} W_0 \, \Delta h_1(\mathbf{x})}_{\text{activation-update term}} 
    \;+\; \underbrace{n^{-a_w} \Delta W_1 \, \Delta h_1(\mathbf{x})}_{\text{second-order term}}.
\end{align}

\emph{Weight-update term.} Using $\|\Delta W_1\| = \Theta_n(n^{-c_w})$ and $\|h_0\| = \Theta_n(1)$:
\begin{align}
    n^{-a_w} \|\Delta W_1 \, h_0(\mathbf{x})\| = n^{-a_w}\, n^{\rho_{w,1}}\, \|\Delta W_1\|\, \|h_0\| = \Theta_n(n^{\rho_{w,1} - a_w - c_w}).
\end{align}
Requiring this to be $\Theta_n(1)$:
\begin{align}
    \boxed{\rho_{w,1} - a_w - c_w = 0.}
    \label{eq:adam-cond-w-rho}
\end{align}

\emph{Activation-update term.} Using $\|W_0\| = \Theta_n(n^{-b_w})$ at initialization, $\|\Delta h_1\| = \Theta_n(1)$, and $a_w + b_w = 1/2$:
\begin{align}
    n^{-a_w} \|W_0 \, \Delta h_1(\mathbf{x})\| = n^{-a_w}\, n^{\omega_{w,1}}\, \|W_0\|\, \|\Delta h_1\| = \Theta_n(n^{\omega_{w,1} - 1/2}).
\end{align}
Requiring this to be $\Theta_n(1)$:
\begin{align}
    \boxed{\omega_{w,1} - \tfrac{1}{2} = 0.}
    \label{eq:adam-cond-w-omega}
\end{align}
As in the SGD case, $\omega_{w,1}$ is not a free variable if we demand stability and non-vanishing updates $r_{w,1} = 0$, and its value gets pinned to $1/2$.

\emph{Second-order term.} Using $\|\Delta W_1\| = \Theta_n(n^{-c_w})$ and $\|\Delta h_1\| = \Theta_n(1)$:
\begin{align}
    n^{-a_w} \|\Delta W_1 \, \Delta h_1(\mathbf{x})\| = n^{-a_w}\, n^{\sigma_{w,1}}\, \|\Delta W_1\|\, \|\Delta h_1\| = \Theta_n(n^{\sigma_{w,1} - a_w - c_w}).
\end{align}
Requiring this to be $\Theta_n(1)$:
\begin{align}
    \boxed{\sigma_{w,1} - a_w - c_w = 0.}
    \label{eq:adam-cond-w-sigma}
\end{align}

The boxed conditions \Cref{eq:adam-cond-w-rho,eq:adam-cond-w-omega,eq:adam-cond-w-sigma} correspond to requiring all three terms to be $\Theta_n(1)$. The minimal requirement for $\|\Delta z_1\| = \Theta_n(1)$ is weaker:
\begin{align}
    \boxed{\max\!\big(\rho_{w,1} - a_w - c_w,\;\; \omega_{w,1} - \tfrac{1}{2},\;\; \sigma_{w,1} - a_w - c_w\big) = 0.}
    \label{eq:adam-cond-w-max}
\end{align}

\paragraph{\textbf{Last layer}.} The output update decomposes similarly:
\begin{align}
    \Delta f_1(\mathbf{x}) 
    = \underbrace{n^{-a_v} \Delta V_1 \, z_0(\mathbf{x})}_{\text{weight-update term}} 
    + \underbrace{n^{-a_v} V_0 \, \Delta z_1(\mathbf{x})}_{\text{activation-update term}} 
    + \underbrace{n^{-a_v} \Delta V_1 \, \Delta z_1(\mathbf{x})}_{\text{second-order term}}.
\end{align}

\emph{Weight-update term.} Using $\|\Delta V_1\| = \Theta_n(n^{-c_v})$ and $\|z_0\| = \Theta_n(1)$:
\begin{align}
    n^{-a_v} \|\Delta V_1 \, z_0(\mathbf{x})\| = n^{-a_v}\, n^{\rho_{v,1}}\, \|\Delta V_1\|\, \|z_0\| = \Theta_n(n^{\rho_{v,1} - a_v - c_v}).
\end{align}
Requiring this to be $\Theta_n(1)$:
\begin{align}
    \boxed{\rho_{v,1} - a_v - c_v = 0.}
    \label{eq:adam-cond-v-rho}
\end{align}

\emph{Activation-update term.} Using $\|V_0\| = \Theta_n(n^{-b_v})$ at initialization and $\|\Delta z_1\| = \Theta_n(1)$:
\begin{align}
    n^{-a_v} \|V_0 \, \Delta z_1(\mathbf{x})\| = n^{-a_v}\, n^{\omega_{v,1}}\, \|V_0\|\, \|\Delta z_1\| = \Theta_n(n^{\omega_{v,1} - a_v - b_v}).
\end{align}
Requiring this to be $\Theta_n(1)$:
\begin{align}
    \boxed{\omega_{v,1} - a_v - b_v = 0.}
    \label{eq:adam-cond-v-omega}
\end{align}
As in the SGD case, demanding $r_{v,1} = 0$ does not pin $\omega_{v,1}$ to a fixed value; it is constrained only through the relation $\omega_{v,1} = a_v + b_v$, where $a_v + b_v$ is itself underdetermined due to the weaker initialization constraint (\Cref{desideratum:stable-init}).

\emph{Second-order term.} Using $\|\Delta V_1\| = \Theta_n(n^{-c_v})$ and $\|\Delta z_1\| = \Theta_n(1)$:
\begin{align}
    n^{-a_v} \|\Delta V_1 \, \Delta z_1(\mathbf{x})\| = n^{-a_v}\, n^{\sigma_{v,1}}\, \|\Delta V_1\|\, \|\Delta z_1\| = \Theta_n(n^{\sigma_{v,1} - a_v - c_v}).
\end{align}
Requiring this to be $\Theta_n(1)$:
\begin{align}
    \boxed{\sigma_{v,1} - a_v - c_v = 0.}
    \label{eq:adam-cond-v-sigma}
\end{align}

As in the middle layer, the boxed conditions \Cref{eq:adam-cond-v-rho,eq:adam-cond-v-omega,eq:adam-cond-v-sigma} correspond to requiring all three terms in $\Delta f_1$ to be $\Theta_n(1)$. The minimal requirement for $\|\Delta f_1\| = \Theta_n(1)$ is weaker:
\begin{align}
    \boxed{\max\!\big(\rho_{v,1} - a_v - c_v,\;\; \omega_{v,1} - a_v - b_v,\;\; \sigma_{v,1} - a_v - c_v\big) = 0.}
    \label{eq:adam-cond-v-max}
\end{align}

\paragraph{Gauge symmetry for Adam.} The boxed conditions \Cref{eq:adam-cond-w-max,eq:adam-cond-v-max} are invariant under the per-layer transformation:
\begin{align}
    a_l \to a_l + \Delta_l, \qquad b_l \to b_l - \Delta_l, \qquad c_l \to c_l - \Delta_l, \nonumber
\end{align}
for any $\Delta_l \in \mathbb{R}$ and each layer $l \in \{u, w, v\}$. The shift in $c_l$ differs from the SGD case ($-2\Delta_l$ there) because Adam's update norm $\Theta_n(n^{-c_l})$ depends on $c_l$ only once, whereas SGD's $\eta_l \|g_l\|$ inherits $c_l$ explicitly and additional $a, b$ dependence implicitly through the gradient.

\subsection{Weight Decay Scaling}
\label{appendix:weight_decay_scaling}

With weight decay, the parameter update at the first step becomes:
\begin{align}
    \Delta \theta_l = -\eta_l\, \phi(g_l) - \eta_l \lambda_l\, \theta_l,
\end{align}
where $\phi(g_l)$ is the optimizer's transformation of the gradient: $\phi(g_l) = g_l$ for SGD and $\phi(g_l) = m_l / (\sqrt{v_l} + \epsilon)$ for Adam. For the weight decay term to contribute at the same scale as the parameters, we require $\eta_l \lambda_l = \Theta_n(1)$. Using $\eta_l = \eta\, n^{-c_l}$ and $\lambda_l = \lambda\, n^{-d_l}$:
\begin{align}
    \boxed{c_l + d_l = 0.}
    \label{eq:wd-cond}
\end{align}
This holds identically for SGD and Adam, since the weight decay term does not depend on $\phi$.

\paragraph{Extended gauge symmetry.} Including weight decay extends the $abc$ gauge symmetry to $abcd$. The condition $c_l + d_l = 0$ is preserved under
\begin{align}
    d_l \to d_l + \zeta \Delta_l,
\end{align}
where $\zeta = 2$ for SGD and $\zeta = 1$ for Adam.

\subsection{Summary of Stability Conditions}
\label{appendix:derivation_summary}

\begin{table}[htbp]
\centering
\caption{Summary of stability conditions for SGD and Adam. Each $r_{l,1} = 0$ condition is a max over the three terms (weight-update, activation-update, second-order) in the activation decomposition.}
\label{tab:stability_conditions_summary}
\vspace{0.5em}
\renewcommand{\arraystretch}{1.15}
\begin{tabular}{|>{\centering\arraybackslash}m{1.26cm}|>{\centering\arraybackslash}m{6.8cm}|>{\centering\arraybackslash}m{5.125cm}|}
\hline
& SGD & Adam \\
\hline
\rotatebox{90}{\parbox{1.8cm}{\centering Stable Init}} 
& 
\multicolumn{2}{c|}{
\begin{tabular}{@{}c@{}}

$a_u + b_u = 0$ \\[0.3em]
$a_w + b_w = 1/2$ \\[0.3em]
$a_v + b_v \geq 1/2$
\end{tabular}
} \\
\hline
\rotatebox{90}{\parbox{3.2cm}{\centering Stable Updates}}\vspace{-1.4cm}
& 
\begin{tabular}{@{}l@{}}
$r_{u,1} = 2a_u + c_u + a_v + b_v = 0$ \\[0.6em]
$r_{w,1} = \max\!\begin{Bmatrix}
\rho_{w,1} - 2a_w - c_w - a_v - b_v \\
\omega_{w,1} - \tfrac{1}{2} \\
\sigma_{w,1} - 2a_w - c_w - a_v - b_v
\end{Bmatrix} \!= 0$ \\[1.2em]
$r_{v,1} = \max\!\begin{Bmatrix}
\rho_{v,1} - 2a_v - c_v \\
\omega_{v,1} - a_v - b_v \\
\sigma_{v,1} - 2a_v - c_v
\end{Bmatrix} \!= 0$
\end{tabular}
&
\begin{tabular}{@{}l@{}}
$r_{u,1} = a_u + c_u = 0$ \\[0.6em]
$r_{w,1} = \max\!\begin{Bmatrix}
\rho_{w,1} - a_w - c_w \\
\omega_{w,1} - \tfrac{1}{2} \\
\sigma_{w,1} - a_w - c_w
\end{Bmatrix} \!= 0$ \\[1.2em]
$r_{v,1} = \max\!\begin{Bmatrix}
\rho_{v,1} - a_v - c_v \\
\omega_{v,1} - a_v - b_v \\
\sigma_{v,1} - a_v - c_v
\end{Bmatrix} \!= 0$
\end{tabular}
\\
\hline
\rotatebox{90}{\parbox{1.2cm}{\centering Weight Decay}}
& 
\multicolumn{2}{c|}{$c_l + d_l = 0$ for all layers $l$} \\
\hline
\end{tabular}
\end{table}

\Cref{tab:stability_conditions_summary} summarizes the conditions on the $\{a, b, c, d\}$ exponents that satisfy~\Cref{desideratum:stable-init,desideratum:stable-updates}. The activation update conditions $r_{l,1} = 0$ at each layer are expressed as a max over the three decomposition terms (weight-update, activation-update, second-order), reflecting the minimal requirement for $\|\Delta h_1\|, \|\Delta z_1\|, \|\Delta f_1\| = \Theta_n(1)$. In the next section, we discuss the conditions on the alignment exponents.

\subsection{Alignment Exponent Values}
\label{appendix:alignment_values}

The stability conditions constrain the alignment exponents, however, the alignment exponents are not free design choices, and are empirical properties that quantify the correlations between weights, activations and their updates. 

At initialization, the weights and activations are all random and their products behave like products of independent Gaussian vectors (same logic as in \Cref{desideratum:stable-init}). As a result, all alignment exponents have the value $1/2$:
\begin{align}
    \rho_{l,1} = \omega_{l,1} = \sigma_{l,1} = \tfrac{1}{2} \qquad \text{at initialization}.
\end{align}

As training progresses, the weights, activations, and their updates become correlated.
As a result, the alignment exponents increase above their random-alignment value of $1/2$ during training, although they typically remain far below the fully aligned value of $1$~\cite{everett2024scaling}.

To derive $\mu$P, ~\citet{yang21feature} first impose $r_{u,1}=r_{w,1}=r_{v,1}=0$, so that all activation updates are $\Theta_n(1)$. 
This constrains the hidden layer exponent $\omega_{w} = 1/2$.
Since weights, activations, and updates become correlated during training, ~\citet{yang21feature} assume that the remaining free exponents attain their maximum value:
\begin{assumption}[$\mu$P Full Alignment Assumption]
\label{assumption:mup_full_alignment}
After imposing $r_{u,1}=r_{w,1}=r_{v,1}=0$, all remaining free alignment exponents are set to $1$:
\begin{align}
    \rho_{w,1} = \rho_{v,1} = \sigma_{w,1} = \sigma_{v,1} = \omega_{v,1} = 1.
\end{align}
\end{assumption}
\Cref{tab:derived_stability} shows the conditions on $abcd$ after imposing the above assumption.
Note that the assumption $\omega_v = 1$ tightens the condition from $a_v + b_v \geq 1/2$ to $a_v + b_v = 1$.
\citet{everett2024scaling} relax this assumption to $\omega_{v,1} = 1/2$, and observe that hyperparameter transfer still holds empirically, which suggests that $\mu$P's full alignment assumption may be excessive.

\begin{table}[htbp]
\centering
\caption{Conditions on exponents $(a, b, c, d)$ for SGD and Adam obtained after imposing~\Cref{assumption:mup_full_alignment}.}
\label{tab:derived_stability}
\renewcommand{\arraystretch}{1.5}
\setlength{\tabcolsep}{8pt}
\begin{tabular}{l l l}
\toprule
\textbf{Layer} & \textbf{SGD} & \textbf{Adam} \\
\midrule
\textbf{First}
& $\begin{aligned}
& a_u + b_u = 0 \\
& 2a_u + c_u + a_v + b_v = 0 \\
& c_u + d_u = 0
\end{aligned}$
& $\begin{aligned}
& a_u + b_u = 0 \\
& a_u + c_u = 0 \\
& c_u + d_u = 0
\end{aligned}$ \\
\midrule
\textbf{Middle}
& $\begin{aligned}
& a_w + b_w = \nicefrac{1}{2} \\
& 2a_w + c_w + a_v + b_v = 1 \\
& c_w + d_w = 0
\end{aligned}$
& $\begin{aligned}
& a_w + b_w = \nicefrac{1}{2} \\
& a_w + c_w = 1 \\
& c_w + d_w = 0
\end{aligned}$ \\
\midrule
\textbf{Last}
& $\begin{aligned}
& a_v + b_v = 1 \\
& 2a_v + c_v = 1 \\
& c_v + d_v = 0
\end{aligned}$
& $\begin{aligned}
& a_v + b_v = 1 \\
& a_v + c_v = 1 \\
& c_v + d_v = 0
\end{aligned}$ \\
\bottomrule
\end{tabular}
\end{table}

\subsection{Gauge Symmetry and Parameterization Choice}
Even with these constraints, the system remains underdetermined due to the gauge symmetry of the $abcd$ parameterization. One exponent per layer must be fixed to obtain a specific parameterization. These choices are typically made based on implementation convenience. For instance, setting $a = 0$ removes the explicit width factors in the forward pass, while setting all learning rate exponents $c = 0$ allows using a single global learning rate. The most common choice is the canonical $\mu$P proposed in~\citet{yang21feature}. \Cref{tab:mup_implementations} summarizes several such implementations, all of which are gauge-equivalent and yield identical training dynamics under SGD and Adam.
This symmetry, however, can be broken by operations that modify the gradients, such as gradient clipping or Adam's $\epsilon$.

\subsection{Multi-Step Scaling}
\label{appendix:multi_step_scaling}

So far we have analyzed the scaling of activation updates at the first training step. 
At later steps, the weights, activations, gradients, and alignment exponents can evolve in a width-dependent way. 
To extend the first-step scaling calculation to multiple steps, we assume that the weight and activation scalings persist throughout training.
\begin{assumption}[Persistent weight and activation scaling]
\label{assumption:persistent_weight_activation_scaling}
For any training step $t$, we assume:
\begin{align}
    \|U_t\| = \Theta_n(n^{-b_u}), \qquad
    \|W_t\| = \Theta_n(n^{-b_w}), \qquad
    \|V_t\| = \Theta_n(n^{-b_v}),
\end{align}
and
\begin{align}
    \|h_t\| = \Theta_n(1), \qquad
    \|z_t\| = \Theta_n(1), \qquad
    \|f_t\| = \mathcal{O}_n(1).
\end{align}
\end{assumption}

For SGD, we additionally assume that the gradient scalings remain the same as at initialization.

\begin{assumption}[Persistent gradient scaling for SGD]
\label{assumption:persistent_gradient_scaling_sgd}
For all training steps $t \leq T$,
\begin{align}
    \|g_{v,t}\| = \Theta_n(n^{-a_v}), \qquad
    \|g_{w,t}\| = \Theta_n(n^{-a_w-a_v-b_v}), \qquad
    \|g_{u,t}\| = \Theta_n(n^{-a_u-a_v-b_v}).
\end{align}
\end{assumption}

Adam does not require this gradient-scaling assumption, because under \Cref{assumption:adam-update} the normalized update has $\Theta_n(1)$ entries.

Under these assumptions, the first-step scaling analysis can be repeated at any fixed step $t$. 
However, this still does not control accumulation over a number of steps that grows with width. 
We therefore also assume a width-independent training horizon.
\begin{assumption}[Width-independent training horizon]
\label{assumption:width_independent_training_horizon}
The number of training steps does not scale with width:
\begin{align}
    T = \Theta_n(1).
\end{align}
\end{assumption}

This assumption is quite restrictive and does not hold in realistic training regimes where the number of optimization steps is much larger than the width, or where the number of steps itself scales with width.
This occurs, for example, in compute-optimal training. 
For a width-$n$ model with $\Theta(n^2)$ parameters, the token budget scales with the parameter count, and as a result the number of optimization steps scales as $T(n) = \Theta_n(n^2)$.
In such regimes, different-width models are trained for different numbers of steps, and therefore accumulate different total update magnitudes. 
Therefore, fixed-step scaling arguments do not directly guarantee that training would remain consistent across widths.

\begin{table}[htbp]
\centering
\caption{Comparison of different $\mu$P implementations for SGD and Adam. All implementations are gauge-equivalent and yield identical training dynamics. Weight decay exponents satisfy $c_l + d_l = 0$.}
\label{tab:mup_implementations}
\renewcommand{\arraystretch}{1.5}
\setlength{\tabcolsep}{4pt}
\begin{tabular}{|l|p{1.2cm}|cccc|cccc|}
\hline
\multirow{3}{*}{Implementation} & \multirow{3}{*}{Layer} & \multicolumn{4}{c|}{SGD} & \multicolumn{4}{c|}{Adam} \\
\cline{3-10}
 & & Mult. & Var. & LR & WD & Mult. & Var. & LR & WD \\
 & & $(n^{-a})$ & $(n^{-2b})$ & $(n^{-c})$ & $(n^{-d})$ & $(n^{-a})$ & $(n^{-2b})$ & $(n^{-c})$ & $(n^{-d})$ \\
\hline
\multirow{3}{*}{\shortstack{No multipliers \\ \citep{hayou2025optimalembeddinglearningrate}}}
& First  & $1$ & $1$ & $n$ & $\nicefrac{1}{n}$ & $1$ & $1$ & $1$ & $1$ \\
& Middle & $1$ & $\nicefrac{1}{n}$ & $1$ & $1$ & $1$ & $\nicefrac{1}{n}$ & $\nicefrac{1}{n}$ & $n$ \\
& Last   & $1$ & $\nicefrac{1}{n^2}$ & $\nicefrac{1}{n}$ & $n$ & $1$ & $\nicefrac{1}{n^2}$ & $\nicefrac{1}{n}$ & $n$ \\
\hline
\multirow{3}{*}{No LR scaling}
& First  & $\sqrt{n}$ & $\nicefrac{1}{n}$ & $1$ & $1$ & $1$ & $1$ & $1$ & $1$ \\
& Middle & $1$ & $\nicefrac{1}{n}$ & $1$ & $1$ & $\nicefrac{1}{n}$ & $n$ & $1$ & $1$ \\
& Last   & $\nicefrac{1}{\sqrt{n}}$ & $\nicefrac{1}{n}$ & $1$ & $1$ & $\nicefrac{1}{n}$ & $1$ & $1$ & $1$ \\
\hline
\multirow{3}{*}{\shortstack{Canonical \\ \citep{yang21feature}}}
& First  & $\sqrt{n}$ & $\nicefrac{1}{n}$ & $1$ & $1$ & $\sqrt{n}$ & $\nicefrac{1}{n}$ & $\nicefrac{1}{\sqrt{n}}$ & $\sqrt{n}$ \\
& Middle & $1$ & $\nicefrac{1}{n}$ & $1$ & $1$ & $1$ & $\nicefrac{1}{n}$ & $\nicefrac{1}{n}$ & $n$ \\
& Last   & $\nicefrac{1}{\sqrt{n}}$ & $\nicefrac{1}{n}$ & $1$ & $1$ & $\nicefrac{1}{\sqrt{n}}$ & $\nicefrac{1}{n}$ & $\nicefrac{1}{\sqrt{n}}$ & $\sqrt{n}$ \\
\hline
\multirow{3}{*}{\shortstack{CompleteP \\ \citep{dey2025dontlazycompletepenables}}}
& First  & $1$ & $1$ & $n$ & $\nicefrac{1}{n}$ & $1$ & $1$ & $1$ & $1$ \\
& Middle & $1$ & $\nicefrac{1}{n}$ & $1$ & $1$ & $1$ & $\nicefrac{1}{n}$ & $\nicefrac{1}{n}$ & $n$ \\
& Last   & $\nicefrac{1}{n}$ & $1$ & $n$ & $\nicefrac{1}{n}$ & $\nicefrac{1}{n}$ & $1$ & $1$ & $1$ \\
\hline
\end{tabular}
\end{table}

\subsection{Attention and LayerNorm scaling} 

In standard transformers, two additional layer types are commonly used beyond the linear layers analyzed so far: attention and LayerNorm. These two have been analyzed in \citet{yang21feature} and \citet{dey2025dontlazycompletepenables}.
We treat them in turn, starting with attention.

\paragraph{Attention Scaling.} For queries, keys, and values $Q, K, V \in \mathbb{R}^{T \times d}$, where $T$ is the context length and $d$ is the head dimension, the attention mechanism computes:
\begin{align}
    A = \mathrm{softmax}\!\left(\frac{Q K^\top}{d^{\alpha}}\right) V,
\end{align}
where $d^{\alpha}$ is the attention scaling factor. If the logits are too large, the softmax concentrates all mass on a single entry. To avoid it, we require the logits norm to be $\| QK^\top / d^{\alpha} \| = \Theta_d(1)$.
Assuming $\|Q\|, \|K\| = \Theta_d(1)$, the dot product norm scales according to the alignment exponent:
\begin{align}
    \rho_{a} := \lim_{d \to \infty} \log_d \frac{\| QK^\top\|}{\|Q\| \|K\|},
\end{align}
giving $\| Q K^\top \| = \Theta_d(d^{\rho_{a}})$. 
At initialization, $W_Q$ and $W_K$ are independent Gaussian matrices, so $Q$ and $K$ behave as independent random matrices and $\|Q K^\top\| = \Theta_d(\sqrt{d})$, giving $\alpha = 1/2$, which is the scaling used in the original Transformer paper~\cite{NIPS2017_3f5ee243}. 
During training, $Q$ and $K$ should develop correlations between them. Under the full alignment assumption (analogous to \Cref{assumption:mup_full_alignment}), $\| Q K^\top\| = \Theta_d(d)$, giving $\alpha = 1$, which is the $\mu$P prescription~\cite{yang21feature}.

The alignment exponent $\rho_a$ itself can depend on width $n$ and evolve during training, so neither prescription is \emph{correct} in general.
That said, $\alpha = 1$ acts as an upper bound, since $\rho_a \leq 1$.

\paragraph{LayerNorm.} For $\mathbf{x} \in \mathbb{R}^n$ with trainable gain $\boldsymbol{\gamma} \in \mathbb{R}^n$ and bias $\beta \in \mathbb{R}^n$, the output of LayerNorm is given by:
\begin{align}
    \mathrm{LN}(\mathbf{x}) = \hat{\mathbf{x}} \odot \boldsymbol{\gamma} + \boldsymbol{\beta}, \qquad \hat{\mathbf{x}} := \frac{\mathbf{x} - \mu_x \mathbf{1}}{\sqrt{\sigma_x^2 + \epsilon}},
\end{align}
where $\|\hat{\mathbf{x}}\| = \Theta_n(1)$. 

The LayerNorm parameters are deterministically initialized as $\gamma_i = 1, \beta_i = 0$, which gives $\|\mathrm{LN}(\mathbf{x})\|_0 = \Theta_n(1)$ at initialization. There is no forward-pass multiplier $a_{\mathrm{LN}}$, and weight decay is typically not applied to LayerNorm parameters, so only the learning rate exponent $c_{\mathrm{LN}}$ remains to be determined.

The gradients with respect to $\boldsymbol{\gamma}$ and $\boldsymbol{\beta}$ are:
\begin{align}
    (g_{\boldsymbol{\gamma}})_i = \tfrac{1}{D}\sum_\mu (\nabla_{\mathrm{LN}} \ell^\mu)_i \, \hat{x}^\mu_i, \qquad
    (g_{\boldsymbol{\beta}})_i = \tfrac{1}{D}\sum_\mu (\nabla_{\mathrm{LN}} \ell^\mu)_i,
\end{align}
where $\nabla_{\mathrm{LN}} \ell^\mu := \nabla_{\mathrm{LN}(\mathbf{x}^\mu)} \ell(f(\mathbf{x}^\mu), \mathbf{y}^\mu)$ denotes the gradient of the per-example loss with respect to the LayerNorm output. Assuming $\|\nabla_{\mathrm{LN}} \ell^\mu\|, \|\hat{\mathbf{x}}^\mu\| = \Theta_n(1)$ entrywise, we get $\|g_{\boldsymbol{\gamma}}\|, \|g_{\boldsymbol{\beta}}\| = \Theta_n(1)$.

Since the gradients are $\Theta_n(1)$, for both SGD and Adam, the parameter updates at the first step satisfy $\|\Delta \boldsymbol{\gamma}_1\| =  \|\Delta \boldsymbol{\beta}_1\| = \Theta_n(n^{-c_{\mathrm{LN}}})$. The resulting output update $\Delta \mathrm{LN}_1(\mathbf{x}) = \hat{\mathbf{x}} \odot \Delta \boldsymbol{\gamma}_1 + \Delta \boldsymbol{\beta}_1$ inherits this scaling. Requiring $\|\Delta \mathrm{LN}_1(\mathbf{x})\| = \Theta_n(1)$ yields $c_{\mathrm{LN}} = 0.$

\subsection{Weight Tying}
In language models, the input embedding and output unembedding matrices are often tied, sharing parameters to reduce the parameter count. In our three-layer setup, this corresponds to setting $V = U^\top$. The forward pass in this case becomes:
\begin{align}
    h(\mathbf{x}) = n^{-a_u} U \mathbf{x}, \qquad
    z(\mathbf{x}) = n^{-a_w} W h(\mathbf{x}), \qquad
    f(\mathbf{x}) = n^{-a_v} U^\top z(\mathbf{x}),
\end{align}
with $U \in \mathbb{R}^{n \times d_{\mathrm{in}}}$ and $W \in \mathbb{R}^{n \times n}$. The tied parameter $U$ appears with two different forward-pass exponents $a_u$ and $a_v$. The initialization variance is controlled by the exponents $b_u, b_w$:
\begin{align}
    U_{ij} \sim \mathcal{N}(0, n^{-2b_u}), \qquad W_{ij} \sim \mathcal{N}(0, n^{-2b_w}).
\end{align}
The per-layer learning rates are scaled by $c_u$ and $c_w$:
\begin{align}
    \eta_u = \eta \, n^{-c_u}, \qquad \eta_w = \eta \, n^{-c_w}.
\end{align}

\looseness -1
In contrast to the untied case, the last layer no longer has its own $b_v$ and $c_v$ exponents, since these are inherited from $U$ as $b_u$ and $c_u$. Only the forward-pass exponent $a_v$ remains as a free parameter for the last layer. Performing a similar analysis to the untied case gives the stability conditions summarized in~\Cref{tab:stability_conditions_tied}.

\begin{table}[t]
\centering
\caption{Stability conditions under weight tying ($U = V^\top$).}
\label{tab:stability_conditions_tied}
\renewcommand{\arraystretch}{1.15}
\begin{tabular}{|>{\centering\arraybackslash}m{1.26cm}|>{\centering\arraybackslash}m{6.8cm}|>{\centering\arraybackslash}m{5.125cm}|}
\hline
& SGD & Adam \\
\hline
\rotatebox{90}{\parbox{1.8cm}{\centering Stable Init}} 
& 
\multicolumn{2}{c|}{
\begin{tabular}{@{}c@{}}
$a_u + b_u = 0$ \\[0.3em]
$a_w + b_w = 1/2$ \\[0.3em]
$a_v + b_u \geq 1/2$
\end{tabular}
} \\
\hline
\rotatebox{90}{\parbox{3.2cm}{\centering Stable Updates}}\vspace{-1.4cm}
& 
\begin{tabular}{@{}l@{}}
$r_{u,1} = a_u + c_u + a_v = 0$ \\[0.6em]
$r_{w,1} = \max\!\begin{Bmatrix}
\rho_{w,1} - 2a_w - c_w - a_v - b_u \\
\omega_{w,1} - \tfrac{1}{2} \\
\sigma_{w,1} - 2a_w - c_w - a_v - b_u
\end{Bmatrix} \!= 0$ \\[1.2em]
$r_{v,1} = \max\!\begin{Bmatrix}
\rho_{v,1} - 2a_v - c_u \\
\omega_{v,1} - a_v - b_u \\
\sigma_{v,1} - 2a_v - c_u
\end{Bmatrix} \!= 0$
\end{tabular}
&
\begin{tabular}{@{}l@{}}
$r_{u,1} = a_u + c_u = 0$ \\[0.6em]
$r_{w,1} = \max\!\begin{Bmatrix}
\rho_{w,1} - a_w - c_w \\
\omega_{w,1} - \tfrac{1}{2} \\
\sigma_{w,1} - a_w - c_w
\end{Bmatrix} \!= 0$ \\[1.2em]
$r_{v,1} = \max\!\begin{Bmatrix}
\rho_{v,1} - a_v - c_u \\
\omega_{v,1} - a_v - b_u \\
\sigma_{v,1} - a_v - c_u
\end{Bmatrix} \!= 0$
\end{tabular}
\\
\hline
\end{tabular}
\end{table}

Next, applying~\Cref{assumption:mup_full_alignment} yields~\Cref{tab:derived_stability_tied}, which compares the untied and tied exponents side by side for both SGD and Adam. For Adam, the first-layer and middle-layer conditions are identical in the tied and untied cases, since Adam's normalized update is independent of the gradient magnitude.

\begin{table}[htbp]
\centering
\caption{Conditions on exponents $(a, b, c)$ under $\mu$P, with and without weight tying.}
\label{tab:derived_stability_tied}
\renewcommand{\arraystretch}{1.5}
\setlength{\tabcolsep}{6pt}
\begin{tabular}{l l l l l}
\toprule
{Layer} & {SGD (untied)} & {SGD (tied)} & {Adam (untied)} & {Adam (tied)} \\
\midrule
\textbf{First}
& $\begin{aligned}
& a_u + b_u = 0 \\
& 2a_u + c_u + a_v + b_v = 0
\end{aligned}$
& $\begin{aligned}
& a_u + b_u = 0 \\
& a_u + c_u + a_v = 0
\end{aligned}$
& $\begin{aligned}
& a_u + b_u = 0 \\
& a_u + c_u = 0
\end{aligned}$
& $\begin{aligned}
& a_u + b_u = 0 \\
& a_u + c_u = 0
\end{aligned}$ \\
\midrule
\textbf{Middle}
& $\begin{aligned}
& a_w + b_w = \nicefrac{1}{2} \\
& 2a_w + c_w + a_v + b_v = 1
\end{aligned}$
& $\begin{aligned}
& a_w + b_w = \nicefrac{1}{2} \\
& 2a_w + c_w + a_v + b_u = 1
\end{aligned}$
& $\begin{aligned}
& a_w + b_w = \nicefrac{1}{2} \\
& a_w + c_w = 1
\end{aligned}$
& $\begin{aligned}
& a_w + b_w = \nicefrac{1}{2} \\
& a_w + c_w = 1
\end{aligned}$ \\
\midrule
\textbf{Last}
& $\begin{aligned}
& a_v + b_v = 1 \\
& 2a_v + c_v = 1
\end{aligned}$
& $\begin{aligned}
& a_v + b_u = 1 \\
& 2a_v + c_u = 1
\end{aligned}$
& $\begin{aligned}
& a_v + b_v = 1 \\
& a_v + c_v = 1
\end{aligned}$
& $\begin{aligned}
& a_v + b_u = 1 \\
& a_v + c_u = 1
\end{aligned}$ \\
\bottomrule
\end{tabular}
\end{table}

\paragraph{Why $a_u$ and $a_v$ must remain separate.} Collapsing them to a single exponent ($a_v = a_u$) would force $a_u + b_u = 0$ from the first-layer init and $a_u + b_u = 1$ from the last-layer condition, resulting in a contradiction. The two forward-pass exponents thus provide the only remaining degree of freedom to satisfy both layer constraints simultaneously.

\looseness -1
\paragraph{A weight-tied $\mu$P example.} Unlike the untied case, we cannot set $a_u = a_v = 0$ simultaneously: substituting into the conditions yields $b_u = 0$ from the first layer and $b_u = 1$ from the last layer. We must therefore fix one of $a_u, a_v$ to zero and let the other absorb the layer mismatch. Choosing $a_u = 0$ for Adam yields the table in~\Cref{table:weight_tied_mup}.
In this case, a naive SP implementation with $\nicefrac{1}{\text{fan-in}}$ initialization and a global learning rate of $\nicefrac{1}{n}$ would not only train the embedding layer slowly, but also have $\Theta(\sqrt{n})$ logits at initialization.

\begin{table}[]
\caption{A weight-tied $\mu$P example.}
\label{table:weight_tied_mup}
    \begin{center}
    \begin{tabular}{|l|cccc|}
    \hline
    Layer & Mult. $(n^{-a})$ & Var. $(n^{-2b})$ & LR $(n^{-c})$ & WD $(n^{-d})$ \\
    \hline
    First ($U$)        & $1$            & $1$            & $1$ & $1$ \\
    Middle ($W$)       & $1$            & $\nicefrac{1}{n}$ & $\nicefrac{1}{n}$ & $n$ \\
    Last ($U^\top$)    & $\nicefrac{1}{n}$ & $1$            & $1$ & $1$ \\
    \hline
    \end{tabular}
    \end{center}
    
\end{table}


\end{document}